\documentclass[10pt,twocolumn,letterpaper]{article}

\usepackage[accsupp]{axessibility}  

\usepackage{iccv}
\usepackage{times}
\usepackage{epsfig}
\usepackage{graphicx}
\usepackage{amsmath}
\usepackage{amssymb}
\usepackage{subcaption}
\usepackage{multirow}
\usepackage{booktabs}
\usepackage{array}
\usepackage{float}
\usepackage{enumitem}
\usepackage[accsupp]{axessibility}  

\usepackage{mathtools}
\include{xcolor}

\usepackage{array}
\newcolumntype{L}[1]{>{\raggedright\let\newline\\\arraybackslash\hspace{0pt}}m{#1}}
\newcolumntype{C}[1]{>{\centering\let\newline\\\arraybackslash\hspace{0pt}}m{#1}}
\newcolumntype{R}[1]{>{\raggedleft\let\newline\\\arraybackslash\hspace{0pt}}m{#1}}

\usepackage[breaklinks=true,bookmarks=false]{hyperref}

\iccvfinalcopy 

\def\iccvPaperID{****} 

\ificcvfinal\fi

\begin{document}

\title{DisUnknown: Distilling Unknown Factors for Disentanglement Learning}

\author{Sitao Xiang\textsuperscript{1,2}, Yuming Gu\textsuperscript{1,2}, Pengda Xiang\textsuperscript{1,2},
Menglei Chai\textsuperscript{3}, Hao Li\textsuperscript{1,2}, Yajie Zhao\textsuperscript{2}, {Mingming He\thanks{Corresponding author.} \textsuperscript{~2}}\\
\textsuperscript{1}University of Southern California, \textsuperscript{2}USC Institute for Creative Technologies, \textsuperscript{3}Snap Inc. \\
{\tt\small sitaoxia@usc.edu, \{ygu, pxiang\}@ict.usc.edu, mchai@snap.com, hao@hao-li.com, }\\
{\tt\small \{zhao, he\}@ict.usc.edu}
}

\maketitle
\ificcvfinal\thispagestyle{empty}\fi 

\begin{abstract}
\label{sec:abstract}






Disentangling data into interpretable and independent factors is critical for controllable generation tasks. With the availability of labeled data, supervision can help enforce the separation of specific factors as expected. However, it is often expensive or even impossible to label every single factor to achieve fully-supervised disentanglement. In this paper, we adopt a general setting where all factors that are hard to label or identify are encapsulated as a single unknown factor. Under this setting, we propose a flexible weakly-supervised multi-factor disentanglement framework \textbf{DisUnknown}, which \textbf{Dis}tills \textbf{Unknown} factors for enabling multi-conditional generation regarding both labeled and unknown factors. Specifically, a two-stage training approach is adopted to first disentangle the unknown factor with an effective and robust training method, and then train the final generator with the proper disentanglement of all labeled factors utilizing the unknown distillation. To demonstrate the generalization capacity and scalability of our method, we evaluate it on multiple benchmark datasets qualitatively and quantitatively and further apply it to various real-world applications on complicated datasets {\let\thefootnote\relax\footnote{Project website: \href{https://stormraiser.github.io/disunknown/}{https://stormraiser.github.io/disunknown/}}}

\end{abstract}
\vspace{-10pt}
\section{Introduction}

Disentanglement learning is the task of breaking down the tangled high-dimensional data variation into interpretable factors. In the desired disentangled representation, each dimension corresponds to a distinct factor of variables, such that when one factor changes, the others remain unaffected \cite{DBLP:journals/pami/BengioCV13}. Disentanglement learning thus enables various downstream tasks such as transfer learning and few-shot learning, as well as challenging controllable image synthesis applications (\eg~\cite{DBLP:journals/corr/abs-2005-09635,DBLP:conf/cvpr/DengYCWT20}).

With the availability of fully-labeled data, \textit{supervised disentanglement} has seen much progress~\cite{DBLP:conf/nips/KulkarniWKT15,DBLP:journals/corr/MathieuZSRL16,DBLP:conf/nips/FengWKZTS18,DBLP:journals/tog/AbermanWLCC19,DBLP:conf/cvpr/DengYCWT20}. However, ground-truth labels are not always accessible, while even human labeling could be prohibitively expensive or inconsistent. Thus, fully-supervised approaches often have a hard time generalizing to common scenarios where labels are only partially available or even entirely missing. In light of this, \textit{unsupervised disentanglement} approaches~\cite{DBLP:conf/nips/ChenCDHSSA16,DBLP:conf/iclr/HigginsMPBGBML17,DBLP:conf/icml/KimM18,DBLP:conf/cvpr/SinghOL19,DBLP:conf/iclr/PressGBW19} have been proposed to address these challenges. However, most of them rely on the strong assumption that the target data is well-structured enough to be cleanly decoupled into explanatory and recoverable factors. And more importantly, there is no guarantee that these factors could be explicitly controlled with respect to the true intended semantics in specific manipulation scenarios.
Therefore, \textit{weakly-supervised disentanglement}, a nice mix of the best of both worlds, has recently become popular for more flexible learning~\cite{DBLP:conf/nips/KulkarniWKT15,DBLP:conf/nips/ReedZZL15,DBLP:conf/aaai/ChenB20,DBLP:conf/iclr/GabbayH20}.
Unfortunately, although state-of-the-art performance is achieved on certain two-factor class-content disentanglement tasks~\cite{DBLP:conf/aaai/ChenB20,DBLP:conf/iclr/GabbayH20}, most existing methods in this category are still unable to extract factor-aware latent representation, which is essential for manipulating individual factors especially when multiple ones are presented.
In conclusion, no solution seems completely satisfactory yet on multi-factor disentanglement, due to the limited generalizability and insufficient performance.

In this paper, we propose a weakly-supervised multi-factor disentanglement learning framework, which handles arbitrary numbers of factors through explicit and near-orthogonal latent representation. Given that challenging factors that are hard to label or interpret exist in most tasks, the \textit{key idea} to our approach is a general setting of $N$-factor disentanglement with $N-1$ factors labeled and a single factor unknown, where all the remaining task-irrelevant or difficult-to-label factors are flexibly encapsulated as one unknown factor. We find such a setting highly effective and practical in real scenarios. Take face motion retargeting as an example, facial expression could be a good candidate for the unknown factor since it is much more difficult to precisely label than others such as the identity and the pose. Thanks to its flexibility, our method naturally adapts to various tasks with varying domains (\eg cartoon and real photos), data types (\eg images, skeletons, and landmarks), integrity (well-structured or in-the-wild), and label continuity (discrete or continuous).

To this end, our framework consists of two major stages: 1) \textit{Unknown Factor Distillation} and 2) \textit{Multi-Conditional Generation}. Specifically, we extract the unknown factor using an adversarial training method in the first stage, and then embed all labeled factors to the latent space as the second stage, which are used to condition the final generation. The core of our method lies in the joint adversarial training of factor encoders and discriminative classifiers, which explicitly disentangles unknown and known factors without introducing leakage between their disentangled representations. 

The performance of our approach is extensively evaluated on several benchmark datasets, both qualitatively and quantitatively. Furthermore, we demonstrate the generalization capacity and practical robustness of the framework on multiple challenging tasks using complicated real-world datasets without any additional manual labeling effort.

Our contributions are:
1) A flexible weakly-supervised disentanglement learning framework that models data as a combination of labeled/unlabeled factors, which scales well to different datasets and benefits various challenging tasks;
2) A two-stage training architecture that explicitly learns disentangled representations for both labeled and unknown semantic factors, enabling mutual exclusive manipulation in the dimension of each factor;
3) A set of learning strategies to improve the effectiveness and robustness of adversarial training throughout our pipeline, which could potentially inspire future research;
4) State-of-the-art performance and wide range of practical uses on multiple challenging tasks including controllable image generation.

\section{Related Work}











\noindent\textbf{Unsupervised Disentanglement}
has become the research focus because it does not require the access to the factors of variation. The pioneering work of InfoGAN~\cite{DBLP:conf/nips/ChenCDHSSA16}, an information-theoretic extension to the Generative Adversarial Network framework~\cite{DBLP:journals/corr/GoodfellowPMXWOCB14}, learns disentangled representations by maximizing the mutual information between the observations and a subset of latents. Considering its training instability and reduced diversity, the Variational Autoencoder (VAE)-based methods~\cite{DBLP:conf/iclr/HigginsMPBGBML17,DBLP:conf/nips/ChenLGD18,DBLP:conf/iclr/0001SB18,DBLP:conf/nips/LopezRJY18,DBLP:conf/icml/KimM18} are proposed for better performance and reconstruction quality by enforcing a factorized aggregated posterior on the latent space. However, these models are built on the assumption that the observations are independent and identically distributed in the datasets, thus successfully disentangled models may not be identified without any supervision~\cite{DBLP:conf/icml/LocatelloBLRGSB19}. Some task-specific unsupervised approaches disentangle two or more factors and achieve impressive results, such as image-to-image translation~\cite{DBLP:conf/eccv/HuangLBK18,DBLP:journals/ijcv/LeeTMHLSY20,DBLP:journals/corr/abs-2001-05017} and motion retargeting~\cite{DBLP:conf/nips/SiarohinLT0S19,DBLP:conf/cvpr/YangZW00ZL20}. These methods do learn disentangled representations, relying on specific categories~\cite{DBLP:conf/cvpr/Tran0017,DBLP:conf/eccv/ShuSGSPK18,DBLP:conf/cvpr/LorenzBMO19,DBLP:conf/cvpr/YangZW00ZL20}, clearly defined domains~\cite{DBLP:conf/eccv/HuangLBK18,DBLP:journals/ijcv/LeeTMHLSY20,DBLP:journals/corr/abs-2001-05017}, or well-structured datasets with certain categories~\cite{DBLP:conf/cvpr/SinghOL19,DBLP:conf/cvpr/LiSOL20}. In contrast, our method proposes a general framework, adapting to various tasks, domains, modalities and factor numbers.

\noindent\textbf{Supervised Disentanglement}
requires strong supervision on specific factors of the data. These methods train a subset of the representations to match the known labels using supervised learning \cite{DBLP:conf/icml/ReedSZL14,DBLP:conf/iclr/XiaoHM18}. With observed class labels only available for partial data, \cite{DBLP:journals/corr/KaraletsosBR15} and \cite{DBLP:conf/nips/NarayanaswamyPM17} propose semi-supervised VAE methods that learn disentangled representation. These supervised methods require large amounts of supervised data that would be expensive to acquire in practice. Although some methods can use synthetic data or data priors to provide full supervision \cite{DBLP:journals/tog/AbermanWLCC19,DBLP:conf/cvpr/DengYCWT20,DBLP:journals/tog/TanCC0CYTY20}, they are limited to processing domain-specific data such as human faces/bodies/hairstyles. Comparing to most supervised methods that only apply to specific tasks, what we propose is a general approach that applies to various applications.

    
    
    
    
    
    
    
    
    
 
\noindent\textbf{Weakly-Supervised Disentanglement}
has been recently studied to build robust disentangled representations without requiring large amounts of data. Such weak supervision is provided as either known relations between the factors in different samples or ground truth labels of a subset of factors. To avoid explicitly labeling, some methods consider guiding disentanglement by matching pairs of data that share the same underlying factor~\cite{DBLP:conf/nips/ReedZZL15,DBLP:conf/nips/KulkarniWKT15,DBLP:journals/corr/KaraletsosBR15,DBLP:conf/aaai/BouchacourtTN18,DBLP:conf/aaai/ChenB20}. By observing a subset of the ground truth factors, some methods perform distribution matching over data and observed factors and supervision is leveraged in style-content disentanglement with available labels for style only~\cite{DBLP:conf/nips/KingmaMRW14,DBLP:conf/nips/YangRYL15,DBLP:journals/corr/abs-2003-06581,DBLP:conf/iclr/GabbayH20}. Some of these methods may achieve state-of-the-art performance on certain class-content disentanglement tasks~\cite{DBLP:conf/aaai/ChenB20,DBLP:conf/iclr/GabbayH20}, but they cannot ensure factor-aware latent representations for manipulating individual factors.
The similar idea of a unified representation of labeled/unlabeled factors has emerged~\cite{DBLP:journals/tomccap/FengYJWSYJ19}. But we present a general disentanglement learning framework, which benefits various tasks.

\section{Method}

\begin{figure}
\centering
    \includegraphics[width=0.7\linewidth]{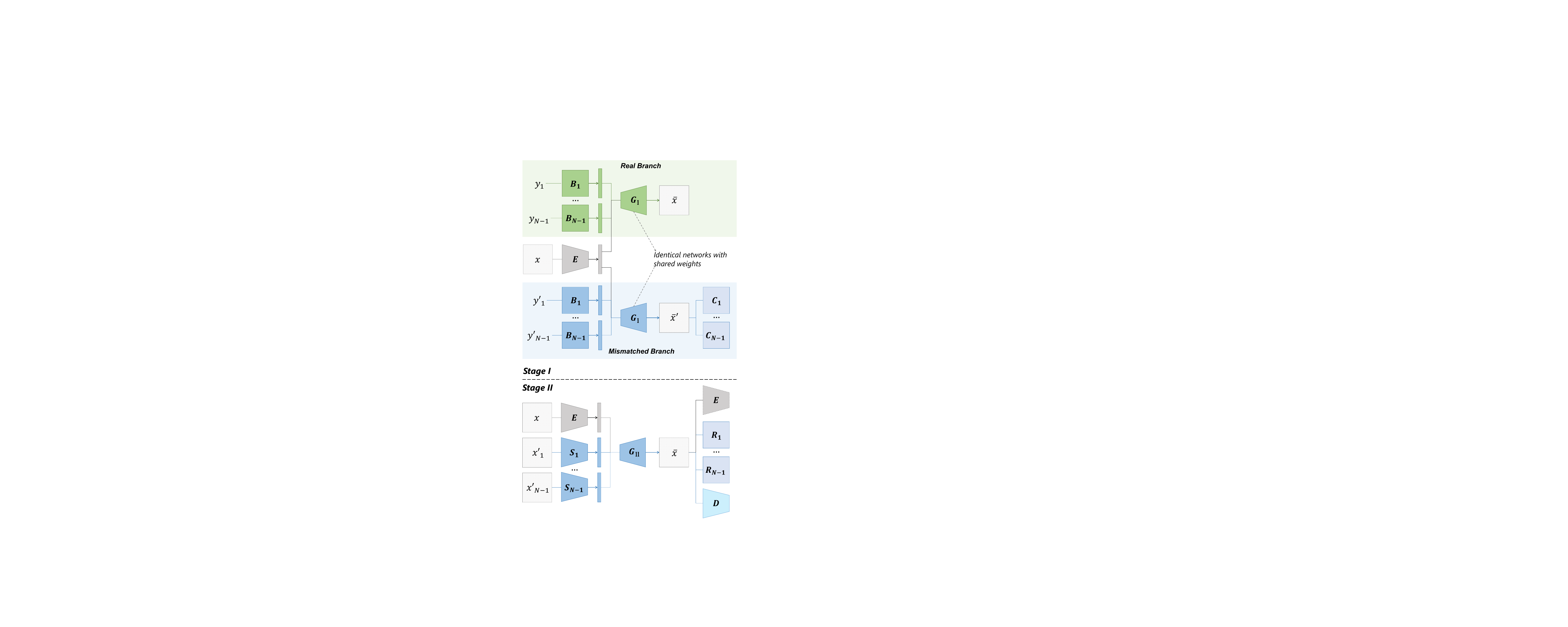}
    \caption{Illustration of our two-stage training architecture.}
    \label{fig:method_stage1}
    \vspace{-0.4cm}
\end{figure}

We propose a generic framework for weakly-supervised disentanglement learning and conditional generation. Instead of jointly training the whole system altogether, we take a two-stage approach. In the first stage, excluding all labeled factors, an encoder is trained to extract disentangled representation of the unknown factor from the input data. And in the second stage, with the unknown factor distilled, a conditional generative adversarial network is trained to embed the labeled data into the latent space, which allows independent control over each factor. By isolating the unknown factor from the labeled ones first, this two-stage training helps reduce the overall complexity of the task and improve the effectiveness of labeled factor disentanglement, as will be elaborated in the Training Strategy part in Stage II.

We note that Stage II is fully-supervised, in which missing labels for the unknown factor is provided by Stage I. Thus our method trivially covers the case where all factors are labeled, by dropping Stage I and using only Stage II.

\subsection{Stage I: Unknown Factor Distillation}
This stage trains an \textit{unknown encoder} $E$ that encodes the unknown factor completely and exclusively. It has two parallel branches in Figure~\ref{fig:method_stage1} (\textit{Stage I}), taking ground truth labels (in the \textit{real branch}) and random labels (in the \textit{mismatched branch}) of all known factors as input, respectively.

Specifically, let there be $N$ factors, with the first $N-1$ ones labeled and the last one unlabeled. $x$ is the training sample, $y=\{y_1,\ldots,y_{N-1}\}$ are the associated \textit{ground truth labels} and $y'=\{y'_1,\ldots,y'_{N-1}\}$ are \textit{random labels} chosen independently of $x$. $E$ is the aforementioned \textit{unknown encoder}, $B=\{B_1,\ldots,B_{N-1}\}$ is a set of \textit{label embedders}, both output normal distributions as in a VAE. $G_\text{I}$ is the \textit{Stage-I generator} that generates a sample $\overline{x}$ or $\overline{x}'$ for the real or mismatched branch, respectively, conditioned on $E$ and $B$. $C=\{C_1,\ldots,C_{N-1}\}$ is a set of \textit{classifiers} that predicts the probability distribution of each factor from a generated sample. Both branches share network structures and weights. The loss functions of the two branches are summed. For now, we assume discrete labels, and discuss continuous-valued factors in the supplementary material.

\noindent\textit{Real branch:}
$B$ map the ground truth labels $y$ to normal distributions. We sample codes from these distributions and feed them to $G_\text{I}$, together with the distilled unknown factor from $E$, to generate the reconstructed sample $\overline{x}$.

\noindent\textit{Mismatched branch:}
By replacing the ground truth labels with random ones $y'_i$, $G_\text{I}$ is asked to generate a mixed sample $\overline{x}'$. $C_i$ predicts the ground truth label from the mixed sample, which indicates if any label information is leaked through $E$, since only $E$ has the access to the ground truth factors in $x$. $C$ are implemented as a single multi-class classifier that only branches at the last layer, and are trained with $E$ in an adversarial manner.

\noindent\textbf{Motivation.}
1) In the real branch, by enforcing a reconstruction loss between the generated sample $\overline{x}$ and the original one $x$, $E$ should include all information not covered by any labeled factor; 2) In the mismatched branch, by minimizing the accuracy of the classifiers $C$ that are trying to predict the ground truth labels from the generated mixed sample $\overline{x}'$, $E$ should exclude any information associated with the labeled factors to avoid label leaking.

\noindent\textbf{Training Strategy.}
As a common problem of adversarial methods, jointly training the adversarial pair of $E$ and $C$ could be unstable. To improve the training robustness, we operate $C$ on samples generated by $G_\text{I}$ instead of codes sampled from the distributions produced by $E$ (similar to~\cite{DBLP:conf/interspeech/ChouYLL18}). This is because, without proper constraints, the distributions in the code space can fluctuate a lot in attempting to prevent the code from being classified. In contrast, with the reconstruction loss in the sample space, the distributions of the generated samples are close to the real ones, which avoids this kind of fluctuation.

As usual, the classifier $C$ minimizes the \textit{negative log-likelihood} (NLL). Let $p$ be a vector representing the probability distribution for a particular factor and $k$ be a class label whose probability is $p_{(k)}$, NLL is defined as:
\begin{equation}
    \mathsf{NLL}(p,k)=-\ln p_{(k)}.
\end{equation}

As the adversarial counterpart, the most obvious choice for the adversarial loss of $E$ is to maximize the NLL loss. However, since NLL is not bounded when the probability $p_{(k)}$ is close to zero, $E$ may prefer to focus on scoring very large NLL values on only a few samples rather than to make every output code equally unclassifiable. Therefore, instead of maximizing the NLL loss, we propose to minimize the \textit{weighted negative log-unlikelihood loss} (NLU):
\begin{equation}
    \mathsf{NLU}_q(p,k)=-\frac{1-q_{(k)}}{q_{(k)}}\ln (1-p_{(k)}),
\end{equation}
where $q$ are the reference distributions, which are always taken to be the actual class distributions in the training set for our purpose.
In the supplementary material, we show how this definition of NLU loss is derived from the desired properties that it should be bounded, yield larger gradients on samples farther from equilibrium, and have the same equilibrium point as maximizing the NLL loss.

\noindent\textbf{Full Objective.}
The full training objective on a single sample for Stage I is formulated as:
\begin{subequations}
\allowdisplaybreaks[4]
\label{eqn:s1}
\begin{align}
    (\mu,\sigma^2)&=E(x),\quad e\sim\mathcal{N}(\mu,\mathrm{diag}(\sigma^2)),\\
    (\alpha_i,\beta^2_i)&=B_i(y_i),\quad b_i\sim\mathcal{N}(\alpha_i,\mathrm{diag}(\beta^2_i)),\\
    (\alpha'_i,(\beta'_i)^2)&=B_i(y'_i),\quad b'_i\sim\mathcal{N}(\alpha'_i,\mathrm{diag}((\beta'_i)^2)),\\
    \overline{x}&=G_\text{I}(e,b_1,\ldots,b_{N-1}),\\
    \overline{x}'&=G_\text{I}(e,b'_1,\ldots,b'_{N-1}),\; p_i=C_i(e,\overline{x}'),\\
    \mathcal{L}_C&=\begin{matrix}\sum_i\mathsf{NLL}(p_i,y_i)\end{matrix},\\
    \mathcal{L}_{GEB}&=\mathsf{Rec}(x,\overline{x})+\lambda_{\textrm{adv1}}\begin{matrix}\sum_i\mathsf{NLU}_q(p_i,y_i)\end{matrix}\nonumber\\
    &+\lambda_{\textrm{KL}}D_{\textrm{KL}}(\mathcal{N}(\mu,\mathrm{diag}(\sigma))||\mathcal{N}(\mathbf{0},I))\\
    &+\lambda_{\textrm{KL}}\begin{matrix}\sum_iD_{\textrm{KL}}(\mathcal{N}(\alpha_i,\mathrm{diag}(\beta^2_i))||\mathcal{N}(\mathbf{0},I)))\end{matrix}.\nonumber
\end{align}
\end{subequations}
The square on the variance vectors $\sigma^2$, $\beta^2_i$ and $(\beta'_i)^2$ are per-element. $\mathsf{Rec}(x,\overline{x})$ is the reconstruction loss function, which is the mean squared error $||x-\overline{x}||^2$ in our experiments. $D_{KL}$ is the KL-divergence. $C$ are trained in the mismatched branch to minimize $\mathcal{L}_C$, averaged over all samples. $E$, $B$, and $G_\text{I}$ jointly minimize $\mathcal{L}_{GEB}$.

\subsection{Stage II: Multi-Conditional Generation}
With the unknown factor distilled in Stage I, this second stage trains encoders $S$ for labeled factors to extract the disentangled representations from the input samples. The final multi-conditional generator $G_\Pi$ accepts conditions for both labeled and unknown factors, and ensures that varying one factor would not affect others in the generated output.

In this stage, as shown in Figure~\ref{fig:method_stage1} (Stage II), the conditions of the unknown and labeled factors come from training samples $x$ and $\{x'_1,\ldots,x'_{N-1}\}$ respectively, all chosen independently. Each $S_i$ of the \textit{labeled-factor encoders} $S=\{S_1,\ldots,S_{N-1}\}$ computes the code for labeled factor $i$ from $x'_i$, while the \textit{unknown encoder} $E$, pre-trained in Stage I, computes the unknown factor code from $x$. The \textit{Stage-II generator} $G_\Pi$ generates a sample $\overline{x}$ conditioned on all the codes (Eq.~\ref{eqn:s2-g}). On $\overline{x}$, a set of \textit{discriminative classifiers} $R=\{R_1,\ldots,R_{N-1}\}$ are trained to enforce the independent controllability of the labeled factor codes, and the pre-trained $E$ is adopted to ensure the consistency of the unknown factor. In addition, a \textit{discriminator} $D$ is applied to ensure the realism of generated samples, as in GAN.

\noindent\textbf{Motivation.}
Trained on random combinations of input samples, the generator $G_\Pi$ is asked to synthesis a new sample with each factor conditioned by encodings from independent sources. Each classifiers $R_i$ enforces that factor $i$ of $\overline{x}$ is completely and solely controlled by $x'_i$, and by choosing each $x'_i$ randomly and independently we ensure that $S_i$ is the only encoder that can consistently compute factor $i$ of $x'_i$. The discriminator $D$ makes the distribution of generated samples and real data indistinguishable globally.

\noindent\textbf{Training Strategy.}
Most previous class-conditional GANs differ on how the generated sample is treated by the classifiers. 
Their classifiers are trained to correctly label the generated sample~\cite{odena2017conditional} or to be uncertain about the task~\cite{springenberg2015unsupervised}. But we go the opposite way: in addition to the NLL loss (Eq.~\ref{eqn:s2-r}) for classifying the training sample $x$ to the correct labels, our discriminative classifiers $R$ are specifically trained to \textit{not} classify the generated sample $\overline{x}$ correctly, by adding the \textit{unweighted} NLU loss:
\begin{equation}
    \mathsf{NLU}(p,k)=-\ln(1-p_{(k)}).
\end{equation}
Its rationale is that a conventional classifier oblivious to the generated samples tends to only learn just enough to distinguish one class from the others, which is insufficient to define the full characteristics of that class. However, if we ask the classifier to identify a generated sample as being in the wrong class, in order to tell real and generated samples apart it would be encouraged to gain a more complete understanding of each class.

$G_\Pi$ and $S$ are jointly trained to ensure that the generated sample $\overline{x}$ is classified to the same labels as the inputs $\{x'_1,\ldots,x'_{N-1}\}$ (the NLL term in Eq.~\ref{eqn:s2-gs}).

Meanwhile, to enforce that the unlabeled factor is consistently controlled by the code from $E$, we minimize the distance between the encodings of the generated sample $\overline{x}$ and the input $x$, using the fixed $E$ (square error term in Eq.~\ref{eqn:s2-gs}). This further explains why $E$ must be trained in a separate stage from the rest of the system: $E$ is used both for providing the input to the generator and for re-encoding the output to compare against the input. If $E$ is allowed to be updated while this distance is being minimized, it could collapse to a state where it encodes everything to a zero vector.

As for the discriminator $D$, we use LSGAN loss functions~\cite{mao2017least} (Eq.~\ref{eqn:s2-d} and the D term in Eq.~\ref{eqn:s2-gs}).

\noindent\textbf{Full Objective.} Similar to Stage I, the full training objective on a single sample for Stage II is formulated as:
\begin{subequations}
\allowdisplaybreaks[4]
\label{eqn:s2}
\begin{align}
    (\mu,\sigma^2)&=E(x),\quad e\sim\mathcal{N}(\mu,\mathrm{diag}(\sigma^2)),\\
    (\alpha'_i,(\beta'_i)^2)&=S_i(x'_i),\quad s'_i\sim\mathcal{N}(\alpha'_i,\mathrm{diag}((\beta'_i)^2)),\label{eqn:s2-s}\\
    \overline{x}&=G_\Pi(e,s'_1,\ldots,s'_{N-1}),\quad(\overline{\mu},\overline{\sigma}^2)=E(\overline{x}),\label{eqn:s2-g}\\
    p_i&=R_i(x),\quad p'_i=R_i(\overline{x}),\\
    \mathcal{L}_R&=\begin{matrix}\sum_i(\mathsf{NLL}(p_i,y_i)+\mathsf{NLU}(p'_i,y'_i))\end{matrix},\label{eqn:s2-r}\\
    \mathcal{L}_D&=(D(x)-1)^2+(D(\overline{x})+1)^2,\label{eqn:s2-d}\\
    \mathcal{L}_{GS}&=||\overline{\mu}-\mu||^2\nonumber\\
    &+\lambda_{\textrm{adv2}}(D(\overline{x})^2+\begin{matrix}\sum_i\mathsf{NLL}(p'_i,y'_i)\end{matrix})\label{eqn:s2-gs}\\
    &+\lambda_{\textrm{KL}}\begin{matrix}\sum_iD_{\textrm{KL}}(\mathcal{N}(\alpha'_i,\mathrm{diag}((\beta'_i)^2))||\mathcal{N}(\mathbf{0},I))\end{matrix}.\nonumber
\end{align}
\end{subequations}
Note that while a total of $N$ input samples are required to generate one sample, in practice this can be efficiently done by computing all factor codes for a whole batch and combining them randomly for generation. Classification labels are permuted accordingly. The classifiers $R$ minimize $\mathcal{L}_R$, the discriminator $D$ minimizes $\mathcal{L}_D$, and the generator $G$ and encoders $S$ jointly minimize $\mathcal{L}_{GS}$.

\subsection{Implementation Details}

For maximum generality we do not favor any specific network architecture. In all our experiments, encoders and generators consist of 3, 4, or 5 stride-2 convolutions for datasets with image sizes of 28, 64, or 128, respectively, followed by 3 fully-connected layers. Discriminators and classifiers have the same convolutional layers but only one fully-connected layer. The convolution feature map depth starts from 32 and doubles after each convolution but does not exceed 256. Fully-connected layers have 512 features.

\section{Experiments}

\subsection{Datasets and Metrics}

\noindent\textbf{Datasets.}
We conduct evaluation experiments on four benchmark datasets: \textit{MNIST}~\cite{lecun1998gradient}, \textit{Fashion-MNIST} (\textit{F-MNIST})~\cite{xiao2017fashion}, \textit{3D Chairs}~\cite{Aubry14}, and \textit{3D Shapes}~\cite{3dshapes18}. For \textit{MNIST} and \textit{F-MNIST}, we use the standard training/testing split. For \textit{3D Chairs} and \textit{3D Shapes}, we randomly hold out $10\%$ of all images for testing and use the rest for training. In \textit{MNIST} and \textit{F-MNIST}, we take \textit{class} as the labeled factor since only it has labels available. In \textit{3D Chairs} which contains three factors, i.e. \textit{model}, \textit{elevation}, and \textit{azimuth}, we combine \textit{elevation} and \textit{azimuth} in to a single unknown factor of \textit{rotation}. In \textit{3D Shapes} which is fully defined by six labeled factors, i.e. \textit{floor hue}, \textit{wall hue}, \textit{object hue}, \textit{scale}, \textit{shape}, and \textit{orientation}, we select one or more factors as labeled and merge the remaining ones into the unknown factor to train various models for our empirical study.

\noindent\textbf{Metrics.}
We evaluate the disentanglement performance by computing the Mutual Information Gap (MIG)~\cite{DBLP:conf/nips/ChenLGD18} of the encoders. Since factors may contain more than one dimension, the mutual information of each factor is defined as the largest one over all dimensions. Then the MIG is computed as the gap of mutual information between the top two factors. Higher MIGs indicate better disentanglement quality.



\begin{figure*}[t]
    \centering
    \begin{subfigure}{0.245\linewidth}
        \centering
        \includegraphics[width=0.85\linewidth]{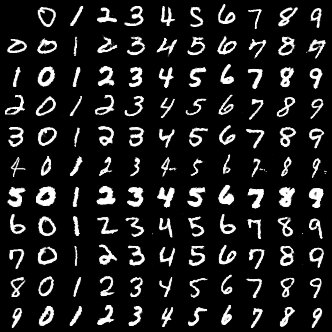}
        \caption{MNIST/ class/ style}
    \end{subfigure}
    \begin{subfigure}{0.245\linewidth}
        \centering
        \includegraphics[width=0.85\linewidth]{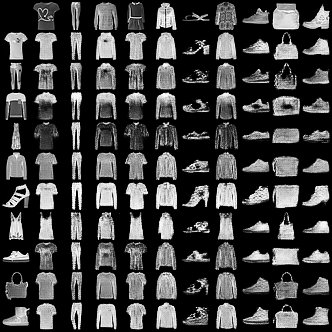}
        \caption{F-MNIST/ class/ style}
    \end{subfigure}
    \begin{subfigure}{0.245\linewidth}
        \centering
        \includegraphics[width=0.85\linewidth]{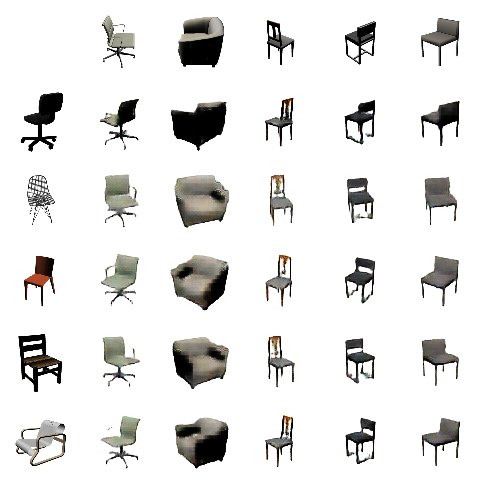}
        \caption{3D Chairs/ model/ rotation}
    \end{subfigure}
    \begin{subfigure}{0.225\linewidth}
        \centering
        \includegraphics[width=0.85\linewidth]{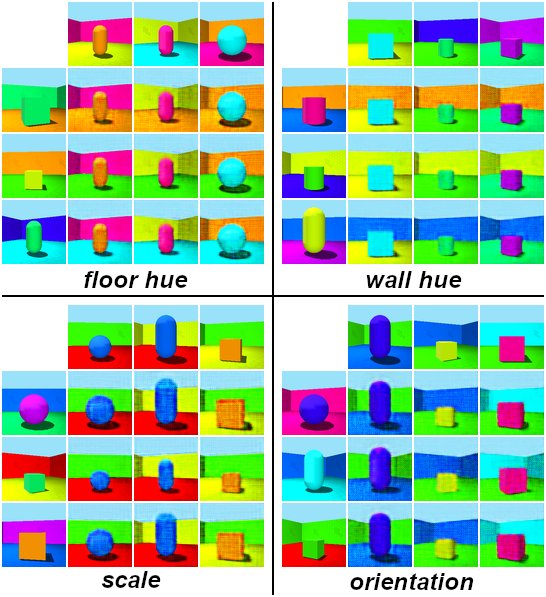}
        \caption{3D Shapes/ noted/ others}
    \end{subfigure}
    \caption{Generated samples on different datasets. The top row and the leftmost column are the input conditions for the labeled and the unknown factors, respectively, annotated as \textit{dataset / labeled / unknown} in the sub-captions.}
    \label{fig:samples}
    \vspace{-0.4cm}
\end{figure*}

\begin{figure}
    \centering
    \begin{subfigure}{0.326\linewidth}
        \centering
        \includegraphics[width=0.8\linewidth,trim=1 1 1 1]{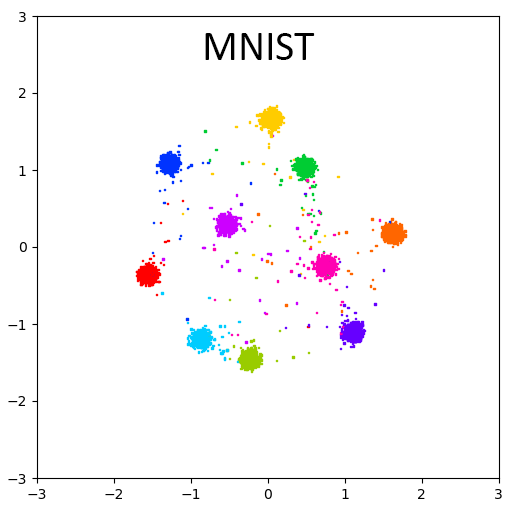}
        \caption{style\hspace{1pt}/\hspace{1pt}class\hspace{1pt}/\hspace{1pt}class}
    \end{subfigure}
    \begin{subfigure}{0.326\linewidth}
        \centering
        \includegraphics[width=0.8\linewidth,trim=1 1 1 1]{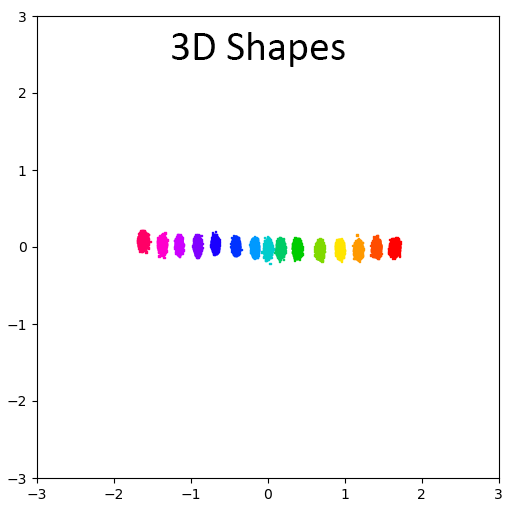}
        \caption{ori.\hspace{1pt}/\hspace{1pt}ori.\hspace{1pt}/\hspace{1pt}ori.}
    \end{subfigure}
    \begin{subfigure}{0.326\linewidth}
        \centering
        \includegraphics[width=0.8\linewidth,trim=1 1 1 1]{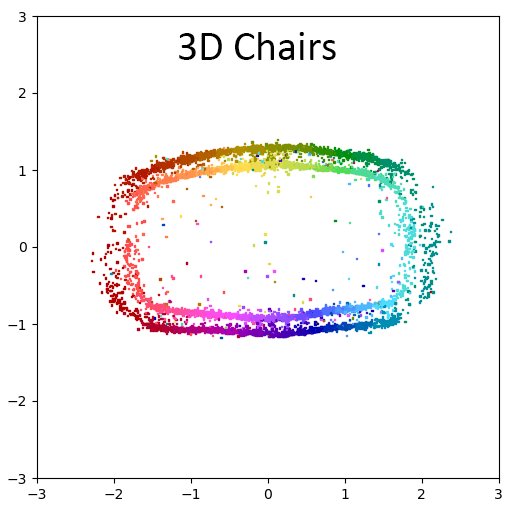}
        \caption{rot.\hspace{1pt}/\hspace{1pt}rot.\hspace{1pt}/\hspace{1pt}rot.}
    \end{subfigure}\\
    \begin{subfigure}{0.326\linewidth}
        \centering
        \includegraphics[width=0.8\linewidth,trim=1 1 1 1]{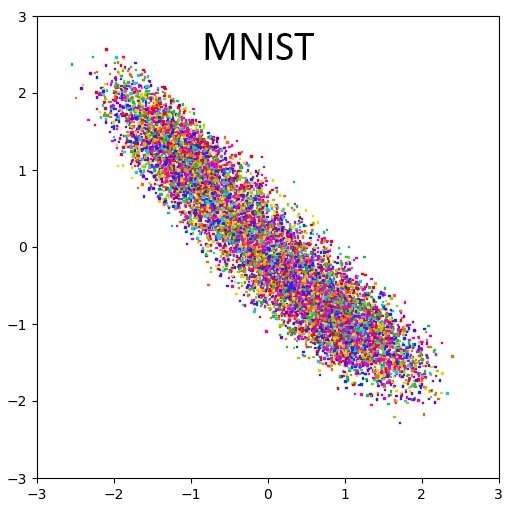}
        \caption{style\hspace{1pt}/\hspace{1pt}style\hspace{1pt}/\hspace{1pt}class}
    \end{subfigure}
    \begin{subfigure}{0.326\linewidth}
        \centering
        \includegraphics[width=0.8\linewidth,trim=1 1 1 1]{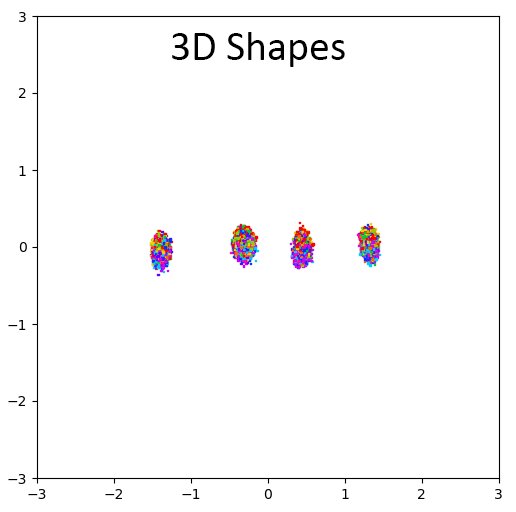}
        \caption{shape\hspace{0.6pt}/\hspace{0.6pt}shape\hspace{0.6pt}/\hspace{0.6pt}floor}
    \end{subfigure}
    \begin{subfigure}{0.326\linewidth}
        \centering
        \includegraphics[width=0.8\linewidth,trim=1 1 1 1]{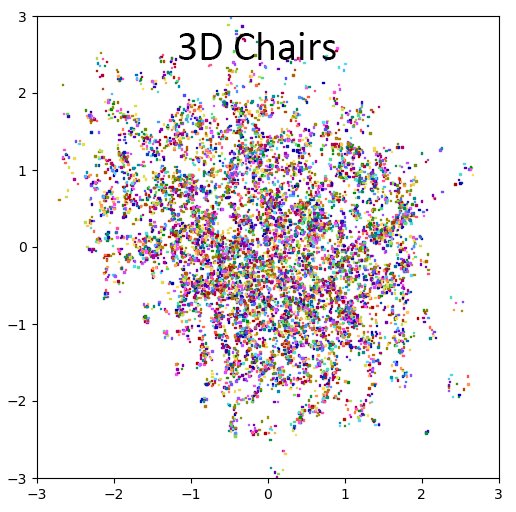}
        \caption{rot.\hspace{1pt}/\hspace{1pt}model\hspace{1pt}/\hspace{1pt}rot.}
    \end{subfigure}
    \caption{Visualizing the disentanglement with test sample distributions. The sub-caption of each figure represents: \textit{unknown factor/ encoding factor/ coloring factor}.}
    \label{fig:plots}
    \vspace{-0.4cm}
\end{figure}

\subsection{Empirical Study}

We empirically study how unknown distillation contributes to the disentanglement of labeled factors and enables control over the unknown factor.

\begin{table}[t]
\caption{Unknown consistency ratios on \textit{3D Shapes} with different unknown factors, w/ and w/o distillation.}
\centering
\setlength{\tabcolsep}{8pt}
 \begin{tabular}{ccc}
  \toprule
  Unknown Factor  & w/ Distillation & w/o Distillation \\
  \hline
  \textit{Floor hue}   & 100.00\% & 63.42\% \\
  \textit{Wall hue}    & 100.00\% & 55.63\% \\
  \textit{Object hue}  & 100.00\% & 68.76\% \\
  \bottomrule
 \end{tabular} 
 \label{tab:w_and_wo_distill}
 \vspace{-0.4cm}
\end{table}

\begin{table}[t]
\caption{Labeled consistency ratios and MIG scores on \textit{3D Shapes} with the unknown factor merged from varying numbers of factors. Zero unknown means fully-supervised.}
\centering
\setlength{\tabcolsep}{12pt}
 \begin{tabular}{ccc}
  \toprule
  \# Unknown & Ratio & MIG $\uparrow$ \\
  \hline
  0  & 100.00\% & 0.9501 \\
  1  & 100.00\% & 0.9555 \\
  2  & 100.00\% & 0.9733 \\
  3  & 100.00\% & 0.9718 \\
  4  & 100.00\% & 0.9393 \\
  5  & 100.00\% & 0.9868 \\
  \bottomrule
 \end{tabular} 
 \label{tab:controll_labeled}
 \vspace{-0.3cm}
\end{table}

\begin{table}[t]
\caption{Mean squared error (MSE) and MIG scores on \textit{3D Shapes} with different unknown factor.}
\centering
\setlength{\tabcolsep}{12pt}
 \begin{tabular}{ccc}
  \toprule
  Unknown Factor  & MSE $\downarrow$ & MIG $\uparrow$ \\
  \hline
  \textit{Floor hue}   & 0.00049 & 0.9607 \\
  \textit{Wall hue}    & 0.00063 & 0.9825 \\
  \textit{Object hue}  & 0.00074 & 0.9766 \\
  \textit{Scale}       & 0.00062 & 0.9411 \\
  \textit{Shape}       & 0.00064 & 0.9637 \\
  \textit{Orientation} & 0.00064 & 0.9537 \\
  \bottomrule
 \end{tabular} 
 \label{tab:diff_factors}
 \vspace{-0.4cm}
\end{table}

\noindent\textbf{Necessity of the Unknown Factor.}
Without the unknown distillation, there is no guarantee that the features represented by the unknown factor remain fixed when altering any labeled ones. To compare, we modify Stage II by replacing the unknown factor code encoded by $E$ with Gaussian noise and removing the feature matching loss $||\overline{\mu}-\mu||^2$ (Eq.~\ref{eqn:s2-gs}), and train three models on \textit{3D Shapes}, with each selecting \textit{floor hue}, \textit{wall hue}, and \textit{object hue} as the unknown factor, respectively. We generate images using the same random code for the unknown factor and independently-sampled random codes for all labeled factors, and then calculate the ratio of results sharing the same unknown feature, namely \textit{consistency ratio}. Due to the simplicity of \textit{3D Shapes}, these three features can be reliably computed by taking the colors at fixed pixel coordinates. Two colors are considered the same if their $L2$ RGB distance is less than half of the mean distance between two adjacent hue samples in the dataset. We generate 10,000 images for each network, and show the results in Table~\ref{tab:w_and_wo_distill}. As can be seen, all ratios reach $100\%$ with distillation, meaning the unknown factor remains unchanged for all test samples. Note that MIGs are not measured here because the disentanglement performance among labeled factors is generally not affected.

\noindent\textbf{Scope of the Unknown Factor.}
In our setting, if there is more than one unknown factor, all these factors will be treated as a whole without individual controllability. However, we can still ensure that the unknown factors are isolated from the labeled ones, and the disentanglement performance of the labeled factors will not be influenced. To verify this, we train six models on \textit{3D Shapes}: starting all factors labeled, we successively merge \textit{floor hue}, \textit{orientation}, \textit{wall hue}, \textit{scale}, and \textit{shape} into the unknown factor, with \textit{object hue} being the last labeled factor at the end. We measure the consistency ratios as introduced in \textit{Necessity of the Unknown Factor} and MIG scores on \textit{object hue} only in Table~\ref{tab:controll_labeled}. Note that all MIG scores are quite close to the upper bound of 1, suggesting good disentanglement quality.

\noindent\textbf{Choice of the Unknown Factor.}
We also study the robustness of our method by choosing different factors as the unknown one on \textit{3D Shapes}. The MSE and MIG results, reflecting the consistent performance of reconstruction and disentanglement, respectively, are shown in Table~\ref{tab:diff_factors}.

\subsection{Results and Visualizations}

To demonstrate the quality of our multi-conditional generator, we plot the generated samples with factors controlled by random references on the benchmark datasets. As shown in Figure~\ref{fig:samples}, our method accurately encodes both known (the top row) and unknown (the leftmost column) factors and uses them to independently control the generation.

We also illustrate the disentanglement quality by visualizing the test sample distributions in the code spaces in Figure~\ref{fig:plots}. For each figure, we pick one encoding factor and one coloring factor from all factors, where both factors may or may not be the same. To draw each test sample on the 2D visualization, we generate the 2D position with the encoding factor and the color with the coloring factor. Specifically, we get its factor code using the encoder corresponding to the encoding factor and project it to 2D by selecting two dimensions with the largest variance. Then we draw a point on that 2D projection using the color mapped to its label of the coloring factor. The indication of good disentanglement is that colors should be clearly separated when the encoding and coloring factors are identical, but entirely mixed with no color pattern or bias when they are different.

\subsection{Comparisons}
We compare our approach against the state-of-the-art, including unsupervised~\cite{DBLP:conf/iclr/HigginsMPBGBML17,DBLP:conf/icml/KimM18,DBLP:conf/nips/ChenLGD18} and weakly-supervised methods~\cite{DBLP:conf/aaai/ChenB20,DBLP:conf/iclr/GabbayH20}. The weakly-supervised methods are run under the same setting as ours where only one factor is labeled for \textit{MNIST}, \textit{F-MNIST}, and \textit{3D Chairs}. Suggested hyperparameters are used to train these models: $\beta=4$ for~\cite{DBLP:conf/iclr/HigginsMPBGBML17}; $\gamma=10$ on \textit{MNIST} and \textit{F-MNIST}, and $\gamma=3.2$ on \textit{3D Chairs} for~\cite{DBLP:conf/icml/KimM18}; $\beta=6$ for \cite{DBLP:conf/nips/ChenLGD18}; and $\beta=10$ for \cite{DBLP:conf/aaai/ChenB20}.


From the results in Table \ref{tab:cmp}, our method achieves substantially higher MIG scores than other methods on all datasets. Since the unsupervised methods~\cite{DBLP:conf/iclr/HigginsMPBGBML17,DBLP:conf/icml/KimM18,DBLP:conf/nips/ChenLGD18} are trained without any supervision, comparing with them is somewhat unfair. Nevertheless, this emphasizes the importance of supervision in the disentanglement tasks, which is also reflected by the observation that the weakly-supervised methods consistently outperform the unsupervised ones.

We show a qualitative comparison in Figure~\ref{fig:traversal} which rotates the \textit{3D Chairs} images via traversing the latent code that depicts the azimuth rotation. The unsupervised methods~\cite{DBLP:conf/iclr/HigginsMPBGBML17,DBLP:conf/icml/KimM18,DBLP:conf/nips/ChenLGD18} can smoothly change the orientation but fail to preserve the original style (\textit{e.g.} shape, color, etc.). Among the weakly-supervised methods, \cite{DBLP:conf/aaai/ChenB20} suffers from over-blurriness, while \cite{DBLP:conf/iclr/GabbayH20} cannot consistently control the orientation. Instead, our method is capable of handling various chair styles and orientations, and achieves better generation quality with the original styles well preserved. Moreover, both weakly-supervised methods are limited to two-factor class-content disentanglement, but our approach is a more flexible multi-factor framework that supports factor-aware latent representation for each individual factor.

\begin{table}
\caption{The MIG scores of different disentanglement methods computed on the benchmark datasets.}

\begin{subtable}[h]{1.0\linewidth}
\centering
\setlength{\tabcolsep}{4.5pt}
 \begin{tabular}{c|ccc|ccc}
  \toprule
  \multirow{2}{*}{Dataset}  & \multicolumn{3}{c|}{Unsupervised} & \multicolumn{3}{c}{Weakly-Supervised} \\
  \cline{2-7}
   & \cite{DBLP:conf/iclr/HigginsMPBGBML17} & \cite{DBLP:conf/icml/KimM18} & \cite{DBLP:conf/nips/ChenLGD18} & \cite{DBLP:conf/aaai/ChenB20} & \cite{DBLP:conf/iclr/GabbayH20} & Ours\\
  \hline
  MNIST & 0.279 & 0.071 & 0.568 & 0.760 & 0.582 & \textbf{0.978}\\
  F-MNIST & 0.105 & 0.043 & 0.111 & 0.630 & 0.539 & \textbf{0.874}\\
  3D Chairs & 0.031 & 0.098 & 0.115 & 0.212 & 0.284 & \textbf{0.404}\\
  \bottomrule
   \end{tabular}
\end{subtable}
 \label{tab:cmp}
 \vspace{-0.3cm}
\end{table}

\begin{figure}[t]
    \centering
    \setlength{\tabcolsep}{0\linewidth}
    \begin{subfigure}{\linewidth}
        \centering
        \includegraphics[width=0.85\linewidth]{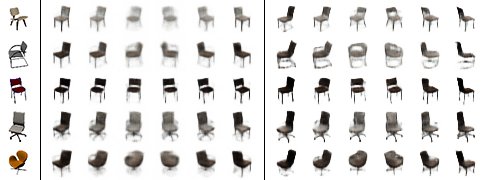}
        \begin{tabular}{C{0.077\linewidth}C{0.461\linewidth}C{0.461\linewidth}}
        &\cite{DBLP:conf/iclr/HigginsMPBGBML17}&\cite{DBLP:conf/icml/KimM18} 
        \end{tabular}
    \end{subfigure}
    \begin{subfigure}{\linewidth}
        \centering
        \includegraphics[width=0.85\linewidth]{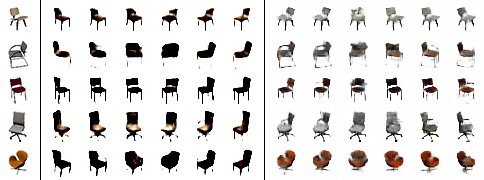}
        \begin{tabular}{C{0.077\linewidth}C{0.461\linewidth}C{0.461\linewidth}}
        &\cite{DBLP:conf/nips/ChenLGD18}&\cite{DBLP:conf/aaai/ChenB20}
        \end{tabular}
    \end{subfigure}
    \begin{subfigure}{\linewidth}
        \centering
        \includegraphics[width=0.85\linewidth]{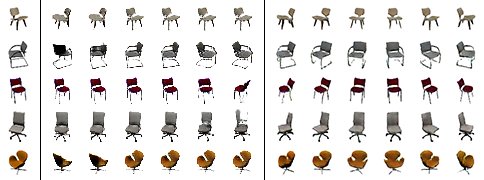}
        \begin{tabular}{C{0.077\linewidth}C{0.461\linewidth}C{0.461\linewidth}}
        &\cite{DBLP:conf/iclr/GabbayH20}&Ours
        \end{tabular}
    \end{subfigure}
    \caption{The rotation manipulation comparison on \textit{3D Chairs} by uniformly sampling the latent codes depicting the azimuth rotation. The leftmost column shows the inputs.}
    \label{fig:traversal}
    \vspace{-0.4cm}
\end{figure}

\section{Downstream Tasks}
\label{sec:app}

\begin{figure}[t]
\centering
    \includegraphics[width=0.85\linewidth]{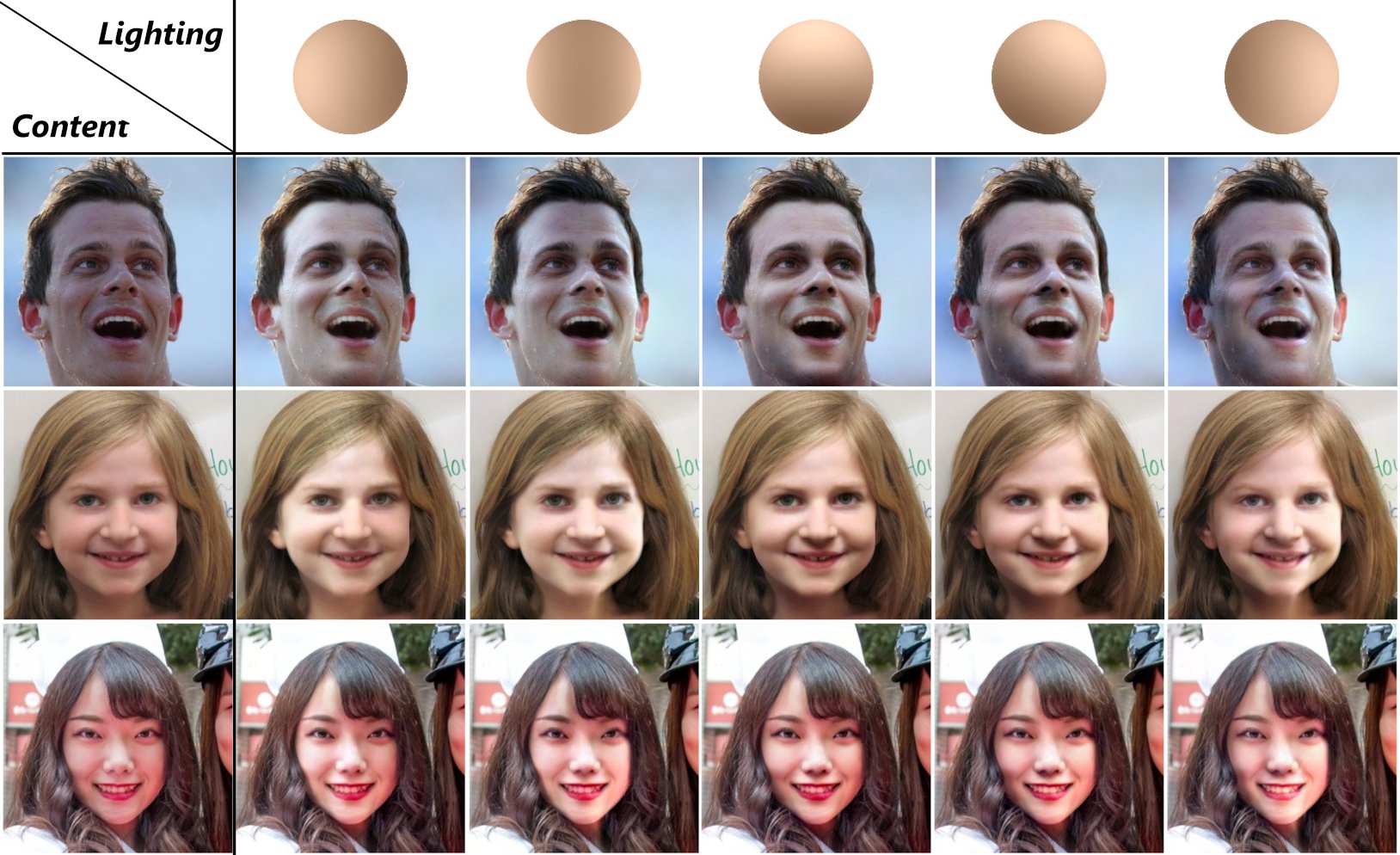}
    \caption{\textbf{Portrait relighting.} The top row shows various environment lightings mapped on a sphere. The leftmost column shows input images, and to the right are the re-lit results conditioned by the lightings in the same column.}
    \label{fig:app_relight}
    \vspace{-0.4cm}
\end{figure}

\begin{figure}
\centering
    \includegraphics[width=0.85\linewidth]{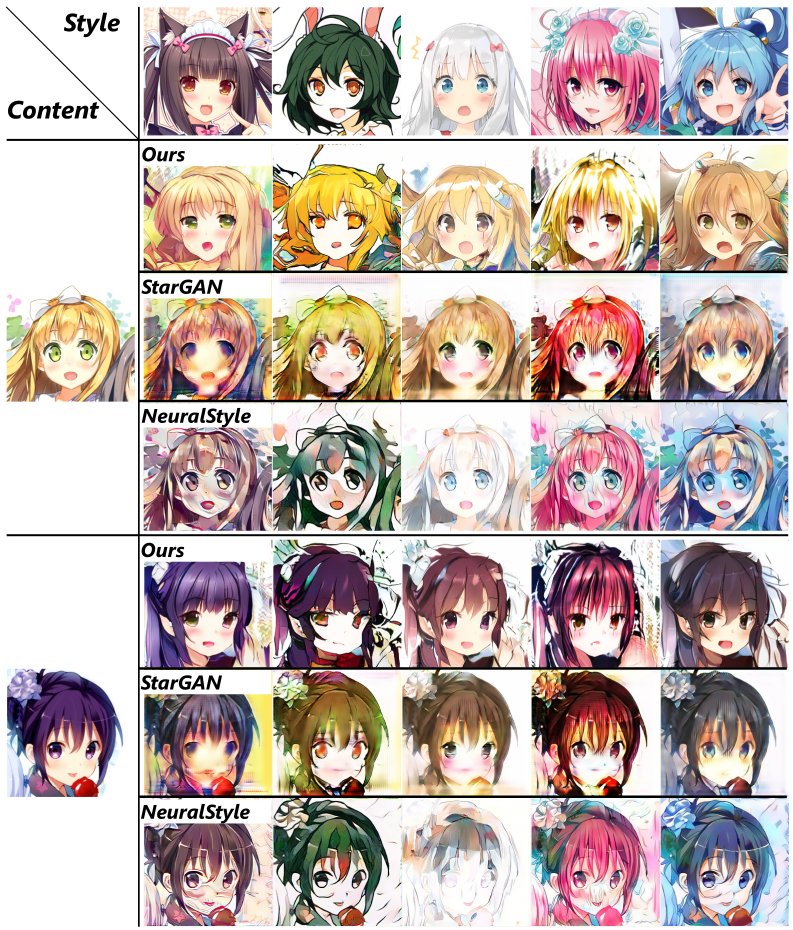}
    \caption{\textbf{Anime style transfer.} Each column is conditioned by the example style at the top row. In each group with three rows, the leftmost image is the content and the results are shown to the right. From top to bottom: our method, StarGAN~\cite{choi2018stargan}, and Neural Style Transfer~\cite{gatys2016image}.}
    \label{fig:app_anime_transfer}
    \vspace{-0.4cm}
\end{figure}

\noindent\textbf{Portrait Relighting.}
We train the network on the dataset combining celebA-HQ~\cite{DBLP:conf/iclr/KarrasALL18} and FFHQ~\cite{DBLP:conf/cvpr/KarrasLA19} by treating the lighting as the labeled factor and the remaining content as unknown. Here, lighting is represented by second-order spherical harmonics coefficients for RGB and estimated with~\cite{kemelmacher20103d,DBLP:journals/tog/ChaiLSCHZ15}.
Figure~\ref{fig:app_relight} shows our portrait relighting results.


\noindent\textbf{Anime Style Transfer.}
We train the network on a custom dataset of $106,814$ anime portrait images drawn by $1,139$ artists collected online. The labeled factor is the artists' identity, which is used as the proxy for style. The unlabeled factor is interpreted as the content of the subject.
Figure~\ref{fig:app_anime_transfer} shows our results on transferring style between different anime portrait illustrations, with comparisons to StarGAN~\cite{choi2018stargan} in multi-domain translation and the original Neural Style Transfer~\cite{gatys2016image}. Our method achieves better results with styles more faithful to the examples. 


\begin{figure}[t]
\centering
    \begin{subfigure}{0.85\linewidth}
        \includegraphics[width=\linewidth]{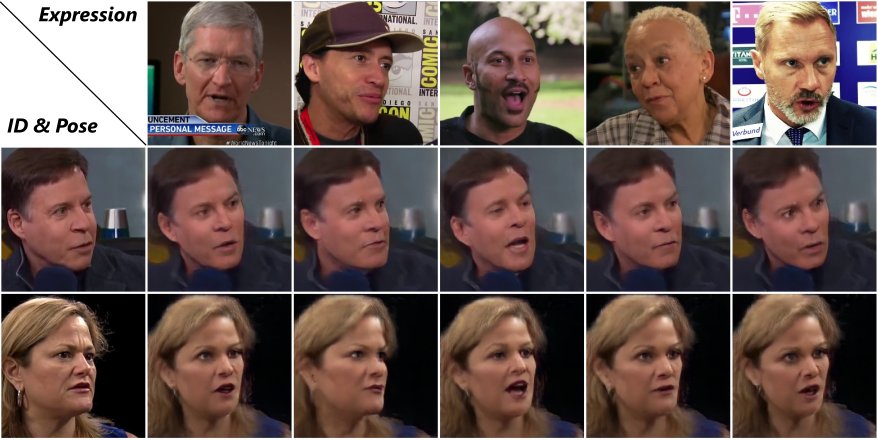}
        \caption{Fix identity and pose, change facial expression.}
    \end{subfigure}
    \begin{subfigure}{0.85\linewidth}
        \includegraphics[width=\linewidth]{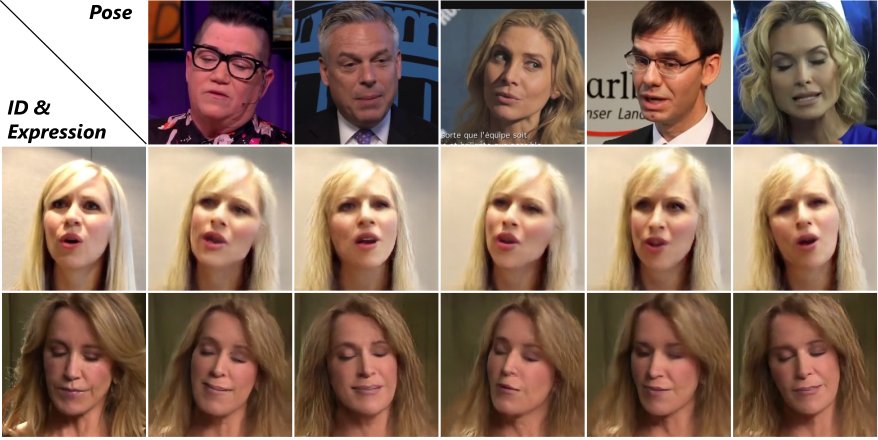}
        \caption{Fix identity and facial expression, change pose.}
    \end{subfigure}
    \caption{\textbf{Face reenactment with expression/pose control}. In each sub-figure, the leftmost column provides the identity and the pose/expression, and the top row provides the expression/pose. The reenactment results are generated with factors conditioned by these inputs.}
    \label{fig:app_face_disentangle}
    \vspace{-0.4cm}
\end{figure}

\begin{figure}[t]
\centering
    \includegraphics[width=0.85\linewidth]{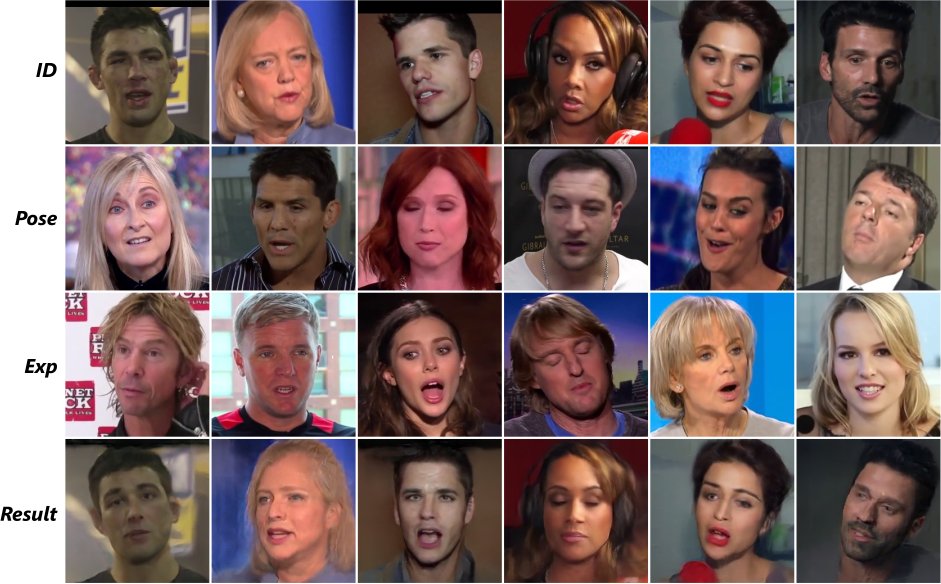}
    \caption{\textbf{Face reenactment with factors from different sources}. The first three rows provide the identity, pose, and expression, respectively. The fourth row shows the results.}
    \label{fig:app_face_transfer}
    \vspace{-0.4cm}
\end{figure}

\noindent\textbf{Landmark-Based Face Reenactment.}
We train our disentanglement network on facial landmark coordinates. After the new landmarks are synthesized with our generator, the output face images are translated from the rasterized landmarks using the image translation network (\eg~\cite{wang2018pix2pixHD},~\cite{DBLP:journals/corr/abs-2004-12452}). We use FD-GAN in \cite{DBLP:journals/corr/abs-2004-12452} for one-shot image translation. The labeled factors are the identity and the head pose, where the pose is represented by Euler angles, estimated from the landmarks. The unlabeled factor is the facial expression. We train the network on VoxCeleb2~\cite{chung2018voxceleb2}.
Figure~\ref{fig:app_face_disentangle}-\ref{fig:app_face_transfer} show our face reenactment results with various controls, including editing a single factor (expression/pose) (Figure~\ref{fig:app_face_disentangle}) and mixing all three factors from different sources (Figure~\ref{fig:app_face_transfer}).


\begin{figure}
\centering
    \includegraphics[width=0.14\linewidth]{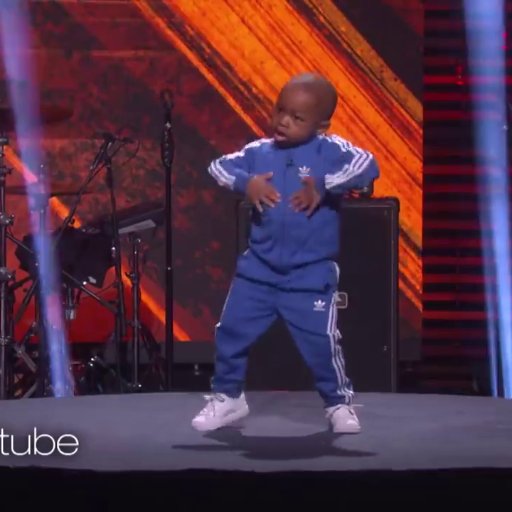}
    \includegraphics[width=0.14\linewidth]{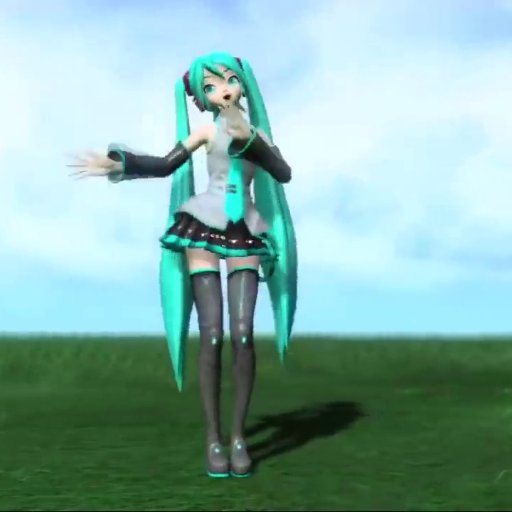}
    \includegraphics[width=0.14\linewidth]{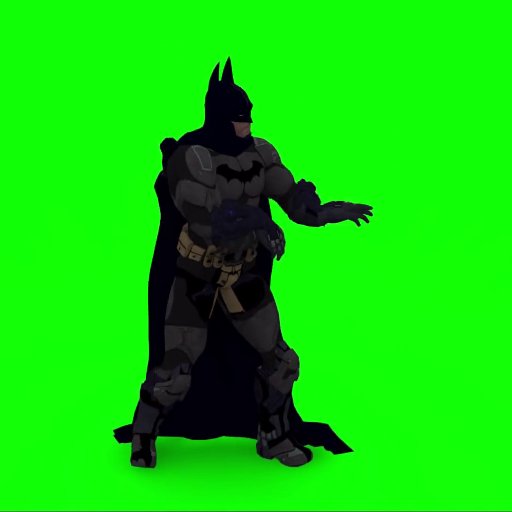}
    \includegraphics[width=0.14\linewidth]{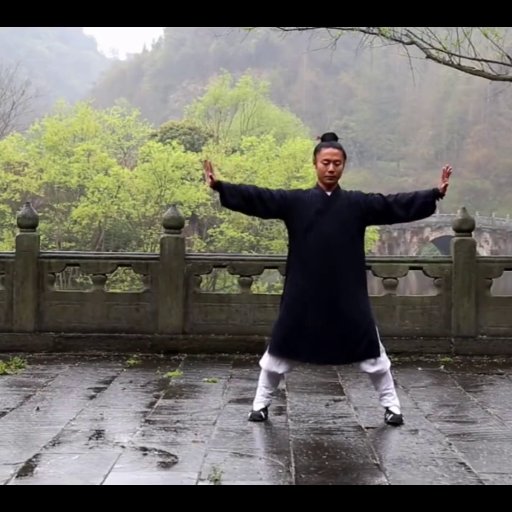}\\
    \includegraphics[width=0.14\linewidth]{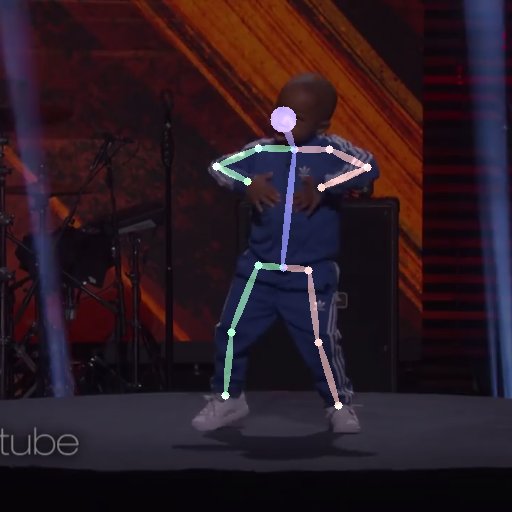}
    \includegraphics[width=0.14\linewidth]{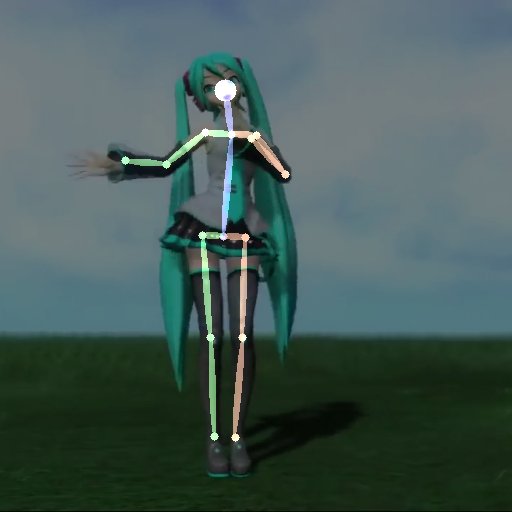}
    \includegraphics[width=0.14\linewidth]{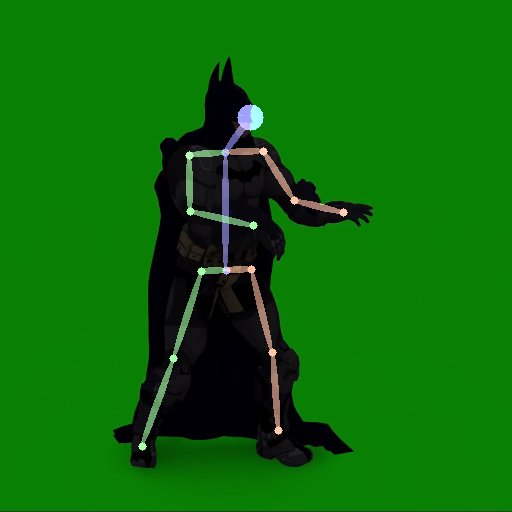}
    \includegraphics[width=0.14\linewidth]{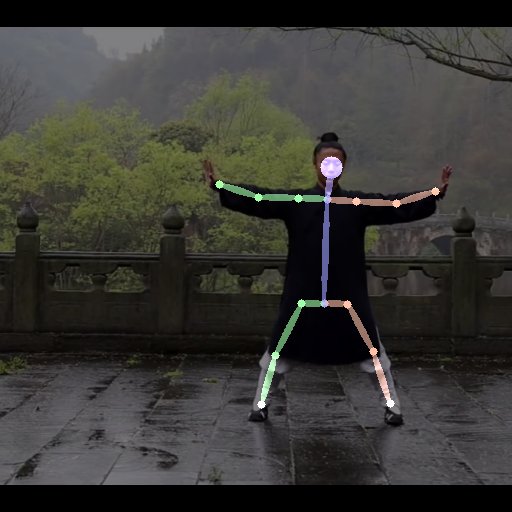}\\
    \includegraphics[width=0.14\linewidth]{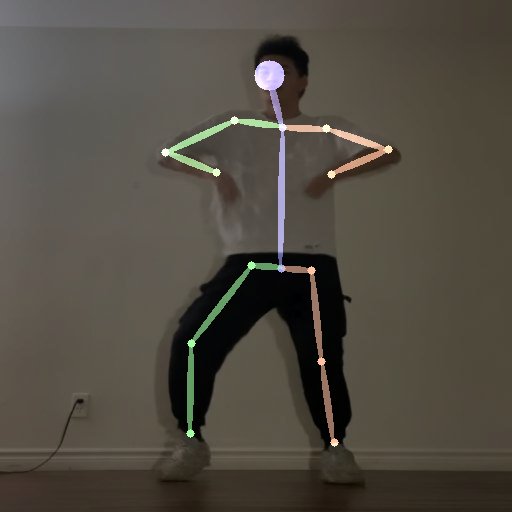}
    \includegraphics[width=0.14\linewidth]{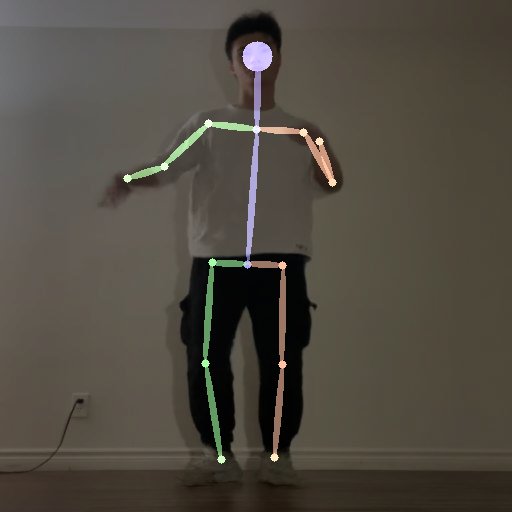}
    \includegraphics[width=0.14\linewidth]{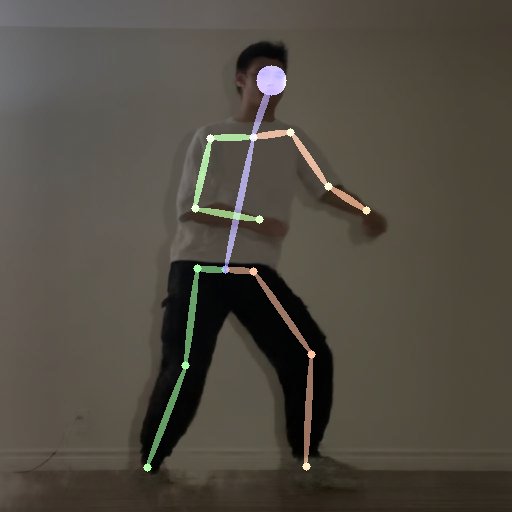}
    \includegraphics[width=0.14\linewidth]{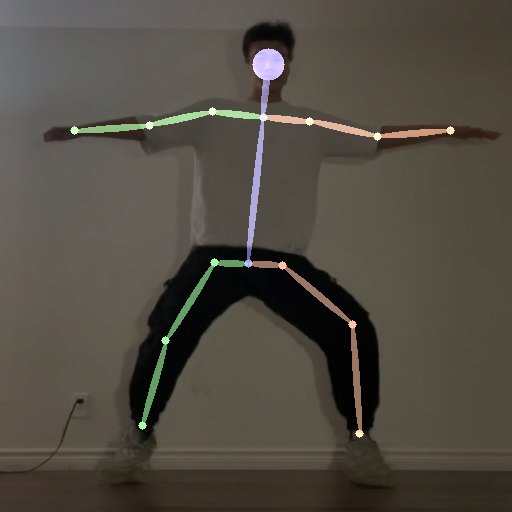}\\
    \includegraphics[width=0.14\linewidth]{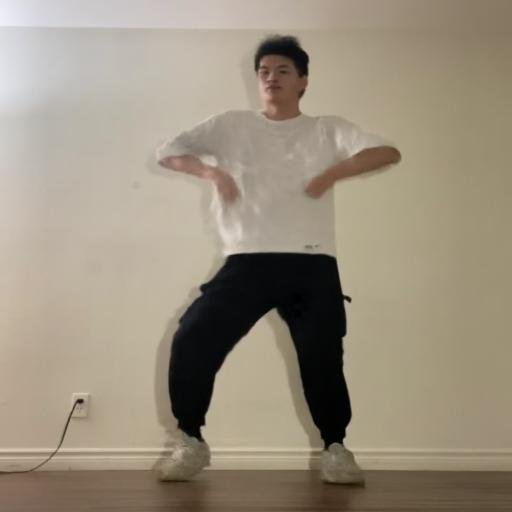}
    \includegraphics[width=0.14\linewidth]{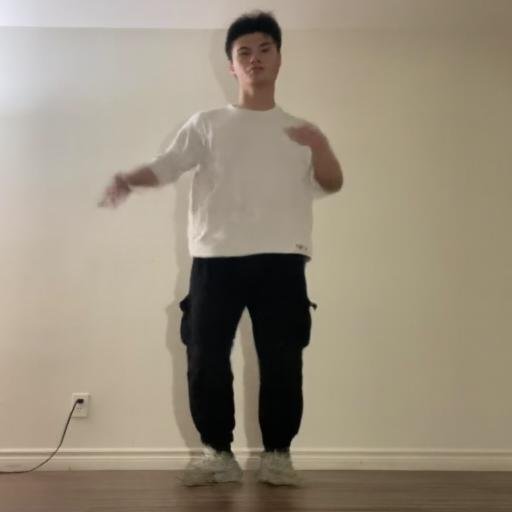}
    \includegraphics[width=0.14\linewidth]{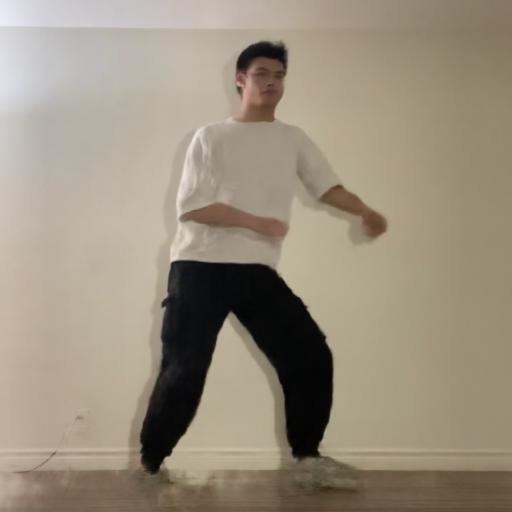}
    \includegraphics[width=0.14\linewidth]{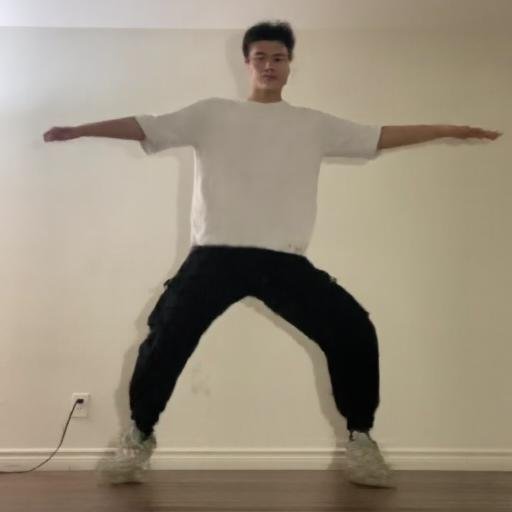}
    \caption{\textbf{Body motion retargeting.} From top to bottom in each column: input source frame, extracted source skeleton, transformed skeleton, and generated frame using~\cite{DBLP:conf/eccv/RenCTFSY20}.}
    \label{fig:app_motion_retarget}
    \vspace{-0.4cm}
\end{figure}

\noindent\textbf{Skeleton-Based Body Motion Retargeting.}
We extract 2D joint coordinates from the driving videos and the actor images. The motion of the driving skeleton and the identity of the actor skeleton are combined to synthesize the target skeleton, with motion as the unknown factor. The images are generated using skeleton-guided synthesis (\eg~\cite{DBLP:conf/eccv/RenCTFSY20}, \cite{DBLP:conf/iccv/ChanGZE19}).
Figure~\ref{fig:app_motion_retarget} shows the motion retargeting results on real images trained on Mixamo~\cite{mixamo}, which demonstrate promising disentanglement between identity and motion. 




\section{Conclusion}
We propose \textit{DisUnknown}, a weakly-supervised multi-factor disentanglement learning framework. By distilling unknown factors, it enables independent control over each factor for multi-conditional generation. Our approach achieves state-of-the-art performance compared to existing unsupervised and weakly-supervised methods on multiple benchmark datasets. We further demonstrate its generalization capacity through various downstream tasks. Moreover, as a general framework, it can easily carry over to other modalities (\textit{e.g.} text, audio) and help improve the stability of other tasks with our adversarial training strategies.

\section*{Acknowledgements}
This research was sponsored by the Army Research Office and was accomplished under Cooperative Agreement Number W911NF-20-2-0053, and sponsored by the CONIX Research Center, one of six centers in JUMP, a Semiconductor Research Corporation (SRC) program sponsored by DARPA, the U.S. Army Research Laboratory (ARL) under contract number W911NF-14-D-0005, and in part by the ONR YIP grant N00014-17-S-FO14. The views and conclusions contained in this document are those of the authors and should not be interpreted as representing the official policies, either expressed or implied, of the Army Research Office or the U.S. Government. The U.S. Government is authorized to reproduce and distribute reprints for Government purposes notwithstanding any copyright notation.
Sitao Xiang wishes to dedicate this work to Sayori, his favorite illustrator, who has always been his inspiration.

{\small
\bibliographystyle{ieee_fullname}
\bibliography{egbib}
}

\cleardoublepage
\appendix

\section{Method Details}
\label{sec:method}

\subsection{Derivation of the Negative Log Unlikelihood}
Here we show how the form of the adversarial loss of the classifiers are chosen.

First, we consider the assumed equilibrium of the adversarial training. In Stage I, the goal is that the encoder's output should not contain any information about the labeled factors. So, each classifier $C_i$ can at best make a guess, and since there is no way to distinguish between inputs from different classes, it should give the same output class distribution for every sample.

Assume that, as commonly done, the output distribution of the classifier is computed by taking the softmax of a vector $t=(t_{(1)},t_{(2)},\ldots,t_{(m)})$, where $m$ is the number of classes for the factor of concern. Let $\mathrm{Sm}$ denote softmax, we have:
\begin{align}
    \label{eqn:nll_d}
    &\frac{\partial}{\partial t_{(i)}}\mathsf{NLL}(\mathrm{Sm}(t),k)\\
    \nonumber =&\frac{\partial}{\partial t_{(i)}}-\ln\frac{e^{t_{(k)}}}{\sum_j e^{t_{(j)}}}\\
    \nonumber =&\frac{\partial}{\partial t_{(i)}} (-t_{(k)}+\ln\sum_j e^{t_{(j)}})\\
    \nonumber =&-\delta_{ik}+\frac{e^{t_{(i)}}}{\sum_j e^{t_{(j)}}}\\
    \nonumber =&-\delta_{ik}+\mathrm{Sm}(t,i),
\end{align}
where $\delta_{ik}=1$ when $i=k$ and $\delta_{ik}=0$ otherwise. At equilibrium, the expectation of gradient over the whole dataset should be zero. Let $q=(q_{(1)},q_{(2)},\ldots,q_{(m)})$ be the class frequency in the dataset. Then we must have
\begin{align}
    &\frac{\partial}{\partial t_{(i)}}\sum_k q_{(k)}\mathsf{NLL}(\mathrm{Sm}(t),k)\\
    \nonumber =&\sum_k q_{(k)}(-\delta_{ik}+\mathrm{Sm}(t,i))\\
    \nonumber =&\mathrm{Sm}(t,i)-q_{(i)}\\
    \nonumber =&0.
\end{align}
That is, $\mathrm{Sm}(t,i)=q_{i}$. So, at the assumed equilibrium, the classifier should give the class distribution in the dataset as the output for any input.

If the adversarial objective is to maximize the NLL loss of the classifier, then naturally at this assumed equilibrium the expected gradient of the adversarial loss function is also zero. But, consider the second-order derivatives:
\begin{align}
    &\frac{\partial^2}{\partial t_{(i)}^2}\mathsf{NLL}(\mathrm{Sm}(t),k)\\
    \nonumber =&\frac{\partial}{\partial t_{(i)}}(-\delta_{ik}+\mathrm{Sm}(t,i))\\
    \nonumber =&\mathrm{Sm}(t,i)(1-\mathrm{Sm}(t,i))\\
    \nonumber >&0,
\end{align}
the adversarial loss function has a local minimum with respect to $t$ at the assumed equilibrium, while the objective is to maximize this function. So in the proximity of the assumed equilibrium, the adversarial objective actually pushes the networks away from the equilibrium. Furthermore, consider the L1 norm of the gradient:
\begin{align}
    \label{eqn:nll_grad}
    &||\frac{\partial}{\partial t}\mathsf{NLL}(\mathrm{Sm}(t),k)||_1\\
    \nonumber =&\sum_i |-\delta_{ik}+\mathrm{Sm}(t,i)|\\
    \nonumber =&\sum_{i\neq k}\mathrm{Sm}(t,i)+(1-\mathrm{Sm}(t,k))\\
    \nonumber =&2-2\cdot\mathrm{Sm}(t,k),
\end{align}
the gradient is larger when the NLL loss is larger, and NLL is not bounded above. If the adversarial objective is to maximize the NLL, it can accelerate towards infinity, which causes strong instability.

Remember that a basic trick in vanilla GAN is that instead of letting the generator maximize
\begin{equation}
    -\ln(1-D(G(z))),
\end{equation}
we let it minimize
\begin{equation}
    -\ln(D(G(z))).
\end{equation}
In a similar vein, instead of maximizing
\begin{equation}
    \mathsf{NLL}(p,k)=-\ln p_{(k)},
\end{equation}
we can minimize what we call ``negative log unlikelihood''
\begin{equation}
    \mathsf{NLU}(p,k)=-\ln (1-p_{(k)}).
\end{equation}

The derivatives are computed as:
\begin{align}
    &\frac{\partial}{\partial t_{(k)}}\mathsf{NLU}(\mathrm{Sm}(t),k)\\
    \nonumber =&\frac{\partial}{\partial t_{(k)}}-\ln(1-\frac{e^{t_{(k)}}}{\sum_j e^{t_{(j)}}})\\
    \nonumber =&-\frac{\sum_j e^{t_{(j)}}}{\sum_j e^{t_{(j)}}-e^{t_{(k)}}}\cdot-\frac{e^{t_{(k)}}(\sum_j e^{t_{(j)}}-e^{t_{(k)}})}{(\sum_j e^{t_{(j)}})^2}\\
    \nonumber =&\mathrm{Sm(t,k)},
\end{align}
\begin{align}
    &\frac{\partial}{\partial t_{(i)}}\mathsf{NLU}(\mathrm{Sm}(t),k)\quad(i\neq k)\\
    \nonumber =&\frac{\partial}{\partial t_{(i)}}-\ln(1-\frac{e^{t_{(k)}}}{\sum_j e^{t_{(j)}}})\\
    \nonumber =&-\frac{\sum_j e^{t_{(j)}}}{\sum_j e^{t_{(j)}}-e^{t_{(k)}}}\cdot-\frac{-e^{t_{(i)}}e^{t_{(k)}}}{(\sum_j e^{t_{(j)}})^2}\\
    \nonumber =&-\frac{\mathrm{Sm}(t,k)\cdot\mathrm{Sm}(t,i)}{1-\mathrm{Sm}(t,k)}.
\end{align}

At the assumed equilibrium $\mathrm{Sm}(t)=q$, where $q$ is the class frequency in the dataset, these evaluate to
\begin{align}
    \label{eqn:nlu_dkk}
    &\left.\frac{\partial}{\partial t_{(k)}}\mathsf{NLU}(\mathrm{Sm}(t),k)\right\vert_{\mathrm{Sm}(t)=q}\\
    \nonumber =&q_{(k)},
\end{align}
\begin{align}
    \label{eqn:nlu_dik}
    &\left.\frac{\partial}{\partial t_{(i)}}\mathsf{NLU}(\mathrm{Sm}(t),k)\right\vert_{\mathrm{Sm}(t)=q}\quad(i\neq k)\\
    \nonumber =&-\frac{q_{(k)}\cdot q_{(i)}}{1-q_{(k)}}.
\end{align}
If the classes are not evenly distributed in the dataset, this may not satisfy the condition that the assumed equilibrium is a stationary point of $\sum_{k}q_{(k)}\cdot\mathsf{NLU}(\mathrm{Sm}(t),k)$. To achieve this, we need to properly weight the NLU by class. We can do this by scaling Equation~\ref{eqn:nlu_dkk} and Equation~\ref{eqn:nlu_dik} to match Equation~\ref{eqn:nll_d}. We define the weighted negative log unlikelihood function as:
\begin{equation}
    \mathsf{NLU}_q(p,k)=-\frac{1-q_{(k)}}{q_{(k)}}\ln (1-p_{(k)}).
\end{equation}

Then we have, at the assumed equilibrium:
\begin{align}
    &\left.\frac{\partial}{\partial t_{(i)}}\sum_{k}q_{(k)}\cdot\mathsf{NLU}_q(\mathrm{Sm}(t),k)\right\vert_{\mathrm{Sm}(t)=q}\\
    \nonumber =&\sum_{k\neq i}-q_{(k)}\cdot\frac{1-q_{(k)}}{q_{(k)}}\cdot\frac{q_{(k)}\cdot q_{(i)}}{1-q_{(k)}}+q_{(i)}\cdot\frac{1-q_{(i)}}{q_{(i)}}\cdot q_{(i)}\\
    \nonumber =&\sum_{k\neq i}-q_{(k)}q_{(i)}+q_{(i)}(1-q_{(i)})\\
    \nonumber =&-(1-q_{(i)})q_{(i)}+q_{(i)}(1-q_{(i)})\\
    \nonumber =&0.
\end{align}
And the L1 norm of the gradient would be:
\begin{align}
    &||\frac{\partial}{\partial t}\mathsf{NLU}_q(\mathrm{Sm}(t),k)||_1\\
    \nonumber =&\sum_i \frac{1-q_{(k)}}{q_{(k)}}\cdot\mathrm{Sm}(t,k)\left(1+\sum_{i\neq k}\frac{\mathrm{Sm}(t,i)}{1-\mathrm{Sm}(t,k)}\right)\\
    \nonumber =&2\cdot\frac{1-q_{(k)}}{q_{(k)}}\cdot\mathrm{Sm}(t,k),
\end{align}
which equals to Equation~\ref{eqn:nll_grad} at the assumed equilibrium, and has the desired property that a smaller value of $\mathrm{Sm}(t,k)$ gives a smaller gradient. Evaluating the second derivative at the assumed equilibrium gives
\begin{align}
    &\left.\frac{\partial^2}{\partial t_{(k)}^2}\mathsf{NLU}(\mathrm{Sm}(t),k)\right\vert_{\mathrm{Sm}(t)=q}\\
    \nonumber =&q_{(k)}(1-q_{(k)}),
\end{align}
\begin{align}
    &\left.\frac{\partial^2}{\partial t_{(i)}^2}\mathsf{NLU}(\mathrm{Sm}(t),k)\right\vert_{\mathrm{Sm}(t)=q}\quad(i\neq k)\\
    \nonumber =&\frac{q_{(k)}q_{(i)}(2q_{(i)}+q_{(k)}-1)}{(1-q_{(k)})^2},
\end{align}
\allowdisplaybreaks
\begin{align}
    &\left.\frac{\partial^2}{\partial t_{(i)}^2}\sum_k q_{(k)} \mathsf{NLU}_q(\mathrm{Sm}(t),k)\right\vert_{\mathrm{Sm}(t)=q}\\
    \nonumber =&q_{(i)}(1-q_{(i)})^2+\sum_{k\neq i}\frac{q_{(k)}q_{(i)}(2q_{(i)}+q_{(k)}-1)}{1-q_{(k)}}\\
    \nonumber =&q_{(i)}\left((1-q_{(i)})^2+\sum_{k\neq i}(\frac{2q_{(k)}q_{(i)}}{1-q_{(k)}}-q_{(k)})\right)\\
    \nonumber >&q_{(i)}\left((1-q_{(i)})^2+\sum_{k\neq i}q_{(k)}(2q_{(i)}-1)\right)\\
    \nonumber =&q_{(i)}\left((1-q_{(i)})^2+(1-q_{(i)})(2q_{(i)}-1)\right)\\
    \nonumber =&q_{(i)}(1-q_{(i)})(1-q_{(i)}+2q_{(i)}-1)\\
    \nonumber =&q_{(i)}^2(1-q_{(i)})\\
    \nonumber >&0.
\end{align}
So the assumed equilibrium is indeed a local minimum of the adversarial loss function we want to minimize.

\subsection{Sample-Space Classification}
The Stage I training procedure of generating samples from random labels for classification is not straightforward to understand, and here we give an explanation.

The distribution of the encoder's output in the code space has few constraints, and there can be different networks that give very different distribution in the code space but are nevertheless essentially equivalent. For example, assume that the last layer of the encoder and the first layer of the generator are linear and the dimension of the code space is $d$. Then we can take an invertible $d\times d$ matrix $M$. We multiply the last layer weights of the encoder by $M$ on the right and multiply the first layer weights of the generator by $M^{-1}$ on the left. In terms of reconstruction, the modified network gives the exact same result as the original, but the code distribution in the code space has been transformed.

This becomes a problem with adversarial training: if the classifier operates in the code space, then to avoid being successfully classified, instead of removing information about the labeled factors from its output, the unknown factor encoder can change its output distribution to confuse the classifier, which results in the code distribution fluctuating constantly in the code space.

In contrast, in the sample space, the distribution of generated samples is anchored to the distribution of training samples and cannot change freely. So, operating the classifier in the sample space can potentially reduce fluctuation in the code space and improve stability. Then the question is which samples should be the input to the classifier.

The reconstructed sample cannot serve as the input to the classifier, since it must contain full information of the input sample, including those about the labeled factors, which is in conflict with the adversarial objective of making the classifier unable to classify by the labeled factor.

\begin{figure}
  \centering
  \includegraphics[width=0.4\linewidth]{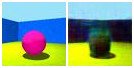}
  \caption{Ill-formed sample generated from mean labeled code.}
  \label{fig:bad_mean}
\end{figure}

Another choice is to combine the unknown code with some kind of ``neutral'' labeled code, for example, the mean of all label embeddings. The problem is that the ``mean'' labeled code may not be ``typical'': as can be seen in Figure \ref{fig:corr1}, in the \textit{3D Shapes} dataset, the network learns that the ten classes of each color attribute are arranged like a circle, but no samples are distributed near the center of the circle. In this case, the mean labeled code does not produce a well-formed sample. An example is shown in Figure \ref{fig:bad_mean}: on the left is the input. The floor hue is the unknown factor. In the generated sample on the right, all labeled factors have been replaced by the mean embedding, and the generated sample is ill-formed.

The result is that the encoder can encode labeled information about the input sample without being detected: the generator can easily recognize an invalid mean labeled code, and when it receives one it generates ill-formed samples so that the classifier cannot classify the generated sample, regardless of what the encoder has encoded.

So the unknown code used for generating the input to the classifier must be a typical code but at the same time independent from the input sample. Thus, we use the embedding of a random label chosen independently from the input sample.

\subsection{Continuous-Valued Factor Disentanglement}
When presenting our method, we assumed that all labels are discrete, class-type labels. Here we discuss the treatment of continuous-valued factors.

Continuous-valued labels are usually associated with regression problems. So it is reasonable to first attempt to use regressors in place of the classifiers. But there are obvious problems with this approach. Consider training samples $x^{(i)}$ each associated with a single, real-valued label $y^{(i)}$. For each $x^{(i)}$, compute $\overline{x}^{(i)\prime}$ as in Equation 3 in the main text. The regressor $C$ should minimize some kind of distance between $C(\overline{x}^{(i)\prime})$ and $y^{(i)}$, say squared distance $(C(\overline{x}^{(i)\prime})-y^{(i)})^2$. Then, in the assumed equilibrium where the encoder does not encode any information about the labeled factor, the regressor can only make a guess. To minimize the expected loss, the best guess should be the mean of all $y^{(i)}$. Let $y^*=\sum_{i=1}^{n} y^{(i)}$. Then if there exists training sample $x^*$ whose label is $y^*$ or very close to $y^*$, any adversarial training would not guarantee to prevent the encoder from encoding full information of $x^*$: there is no way to distinguish whether the regressor is giving a $y^*$ because it has detected labeled information in its input, or it is giving a $y^*$ because it detected no such information and is making a guess.

While we have not stated explicitly, we have already provided the solution to working with continuous-valued factors: note that in the \textit{3D Chairs} dataset, the rotation angle is not a true category-type factor, but a quantized continuous-valued factor. Treating it as a category-type factor gives satisfactory results. Similarly, for any continuous-valued factor, we can always divide its range into a suitable number of buckets and quantize the factor into discrete labels. Generally, a few dozen buckets would work fine.
\section{Data and Metrics}

\subsection{Data}

The image size and list of factors of the datasets are given in Table \ref{tab:dataset}, with the number of possible values of each factor and the length of code we use for the encoder of that factor.

\begin{table}
\caption{Datasets used for evaluation and comparison.}
\vspace{-1.5em}
\label{tab:dataset}
\begin{center}
\begin{tabular}{cccc}
  \toprule
  Dataset                        & Factor & \# of Values & Code Size \\
  \midrule
  \multirow{2}{*}{\parbox{55pt}{\centering MNIST\\$28\times28\times1$}}         & Class & 10 & 10 \\
                                 & (Style) & N/A & 64 \\
  \midrule
  \multirow{2}{*}{\parbox{55pt}{\centering F-MNIST\\$28\times28\times1$}}  & Class & 10 & 10 \\
                                 & (Style) & N/A & 64 \\
  \midrule
  \multirow{3}{*}{\parbox{55pt}{\centering 3D Chairs\\$128\times128\times3$}}     & Model & 1393 & 512 \\
                                 & Elevation & 2 & 2 \\
                                 & Azimuth & 31 & 2 \\
  \midrule
  \multirow{6}{*}{\parbox{55pt}{\centering 3D Shapes\\$64\times64\times3$}}     & Floor hue & 10 & 8 \\
                                 & Wall hue & 10 & 8 \\
                                 & Object hue & 10 & 8 \\
                                 & Scale & 8 & 8 \\
                                 & Shape & 4 & 8 \\
                                 & Orientation & 15 & 8 \\
  \bottomrule
\end{tabular}
\end{center}
\end{table}

\subsection{Metrics}

The Mutual Information Gap (MIG) was originally proposed for the unsupervised setting. We made some adjustments to the computation of MIG to suit the weakly-supervised setting and to allow multi-dimensional code spaces for each factor.

Let $N$ be the number of factors, $L_i$ be the (discrete) random variable representing the label of factor $i$ and $X_i$ be the vector-valued random variable representing the output of the encoder for factor $i$, $i = 1, 2, \ldots, N$. Let $X_i^{(k)}$ be the $k$-th entry of $X_i$, which is a real-valued random variable. The normalized mutual information between $L_i$ and $X_j^{(k)}$ is defined as in \cite{DBLP:conf/nips/ChenLGD18}:
\begin{equation}
    \hat{I}(L_i;X_j^{(k)})=\frac{I(L_i;X_j^{(k)})}{H(L_i)}.
\end{equation}
Then, the multi-dimensional mutual information between $L_i$ and $X_j$ is defined by taking the maximum of $\hat{I}(L_i;X_j^{(k)})$ over $k$:
\begin{equation}
    \hat{I}(L_i;X_j)=\max_{k}\hat{I}(L_i;X_j^{(k)}).
\end{equation}

One might argue that it is mathematically more meaningful to take the normalized mutual information between $L_i$ and the whole $X_j$ instead. But we found that, as the dimensionality of the code space increases, the number of samples required to accurately estimate $H(X_j)$ and $H(X_j|L_i)$ increases exponentially. And when the number of samples is insufficient, even a randomly initialized encoder would be incorrectly computed to have normalized mutual information close to one, which makes the evaluation meaningless. So we have to settle with our current definition.

Then, the MIG is computed as
\begin{equation}
    \label{eqn:mig}
    \mathrm{MIG}=\frac{1}{N}\sum_{i}(\hat{I}(L_i;X_i)-\max_{j\neq i}\hat{I}(L_i;X_j)).
\end{equation}

As we have noted, in Table 2 in the main text we are only concerned with the disentanglement between the combined unknown factor and each individual labeled factor. To reflect this, in the computation of MIG here, in Equation \ref{eqn:mig} we only take the average over $i$ where factor $i$ is labeled.

Special procedures were taken to compute the MIG for \cite{DBLP:conf/iclr/GabbayH20}: the definition of MIG requires the distribution of $X_j$ to have continuous support, for otherwise all the normalized mutual information would be equal to one and the MIG would be zero. The output of \cite{DBLP:conf/iclr/GabbayH20} is in the form of an exact code, rather than a normal distribution as in VAE-based methods, so the $X_j$'s will have discrete support, making MIG non-applicable. Note that during the training of \cite{DBLP:conf/iclr/GabbayH20}, Gaussian noise with fixed standard deviation was added to the content embedding, which effectively turns discrete codes into a distribution with full support. So in the computation of MIG, we similarly add Gaussian noise with a fixed standard deviation. The standard deviation is chosen using the following procedure: for all possible pairs of $(\sigma_1, \sigma_2)$ where $\sigma_1, \sigma_2\in\{10^{-5},2\times 10^{-5}, 5\times 10^{-5}, 10^{-4}, \ldots, 1, 2, 5, 10\}$, we compute MIG by adding $\mathcal{N}(0,\sigma_1^2I)$ to the content (``unknown factor'' in our terminology) code and $\mathcal{N}(0,\sigma_2^2I)$ to the class (``labeled factor'') code. The values of $\sigma_1$ $\sigma_2$ are chosen so that the average MIG under two settings on the \textit{3D chairs} dataset (rotation unknown and model unknown) is maximized. By this we determine that $\sigma_1=0.05$ and $\sigma_2=0.02$.

\section{Ablation Analysis}
\label{sec:ablation}

\subsection{Stability of Stage I}

\begin{figure}
  \centering
  \includegraphics[width=0.75\linewidth]{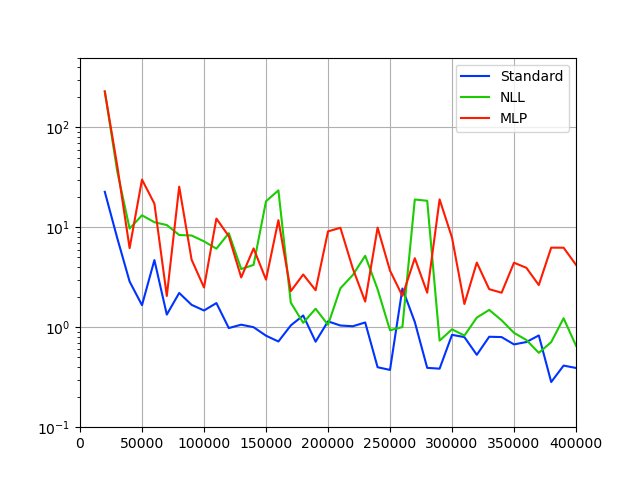}
  \caption{Average squared code distance.}
  \label{fig:code_distance}
\end{figure}

\begin{table}
\caption{Comparison of result with different Stage II configurations on MNIST.}
\centering
 \begin{tabular}{ccc}
  \toprule
  Configuration  & MSE $\downarrow$ & MIG $\uparrow$\\
  \midrule
  Standard            & \textbf{0.0086} & \textbf{0.978} \\
  Non-adversarial $R$ & 0.0103 & 0.916 \\
  No unknown code condition & 0.0402 & 0.930 \\
  \bottomrule
 \end{tabular} 
 \label{tab:ablation_stage2}
\end{table}

\begin{figure*}
    \centering
    \begin{subfigure}{0.3\linewidth}
        \includegraphics[width=\linewidth,trim=10 0 10 0]{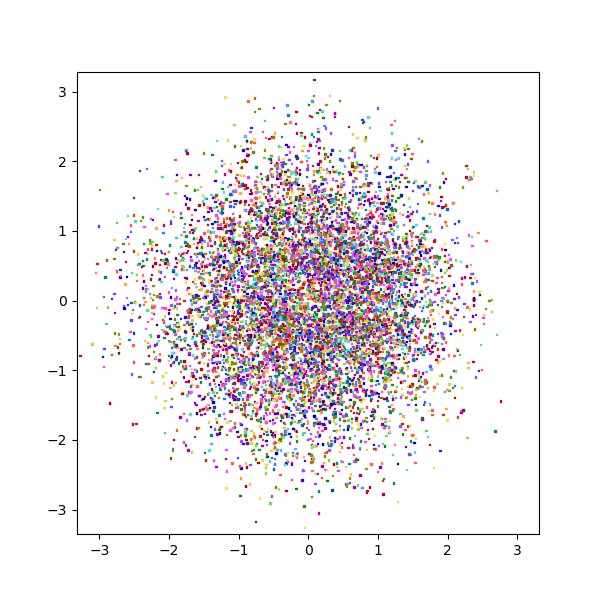}
        \caption{$\beta$-VAE~\cite{DBLP:conf/iclr/HigginsMPBGBML17}}
    \end{subfigure}
    \begin{subfigure}{0.3\linewidth}
        \includegraphics[width=\linewidth,trim=10 0 10 0]{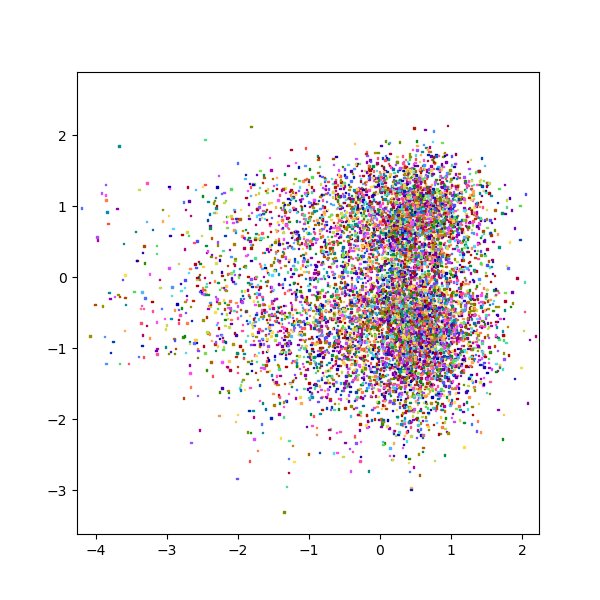}
        \caption{Factor-VAE~\cite{DBLP:conf/icml/KimM18}}
    \end{subfigure}
    \begin{subfigure}{0.3\linewidth}
        \includegraphics[width=\linewidth,trim=10 0 10 0]{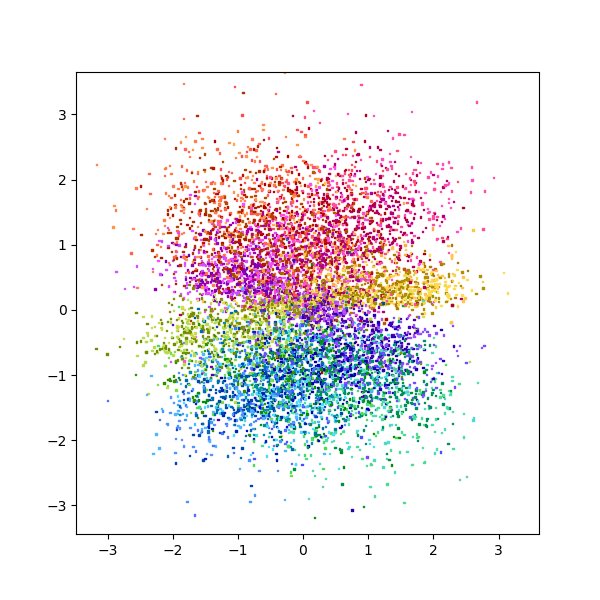}
        \caption{$\beta$-TCVAE~\cite{DBLP:conf/nips/ChenLGD18}}
    \end{subfigure}
    \raisebox{-63pt}{\includegraphics[width=0.0475\linewidth,trim=20 10 20 10]{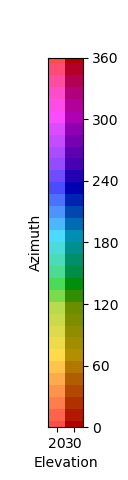}}
    \\
    \begin{subfigure}{0.3\linewidth}
        \includegraphics[width=\linewidth,trim=10 0 10 0]{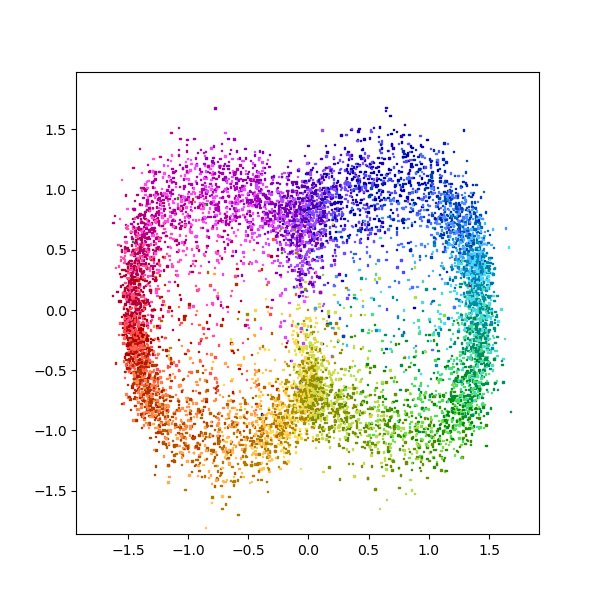}
        \caption{Pairwise-VAE~\cite{DBLP:conf/aaai/ChenB20}}
    \end{subfigure}
    \begin{subfigure}{0.3\linewidth}
        \includegraphics[width=\linewidth,trim=10 0 10 0]{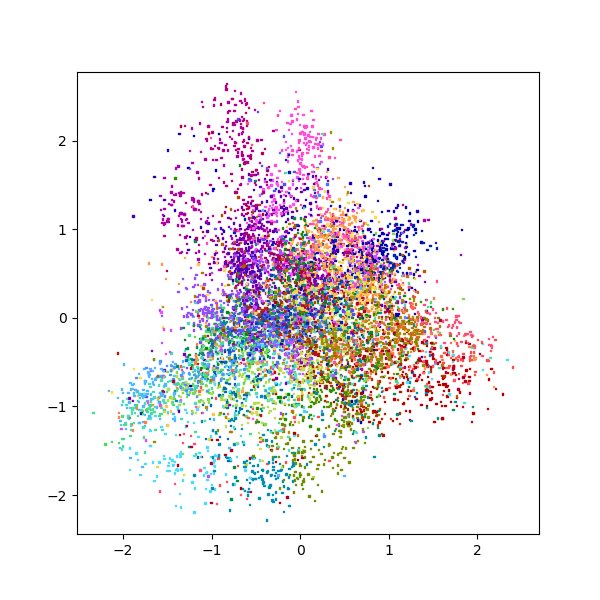}
        \caption{Lord~\cite{DBLP:conf/iclr/GabbayH20}}
    \end{subfigure}
    \begin{subfigure}{0.3\linewidth}
        \includegraphics[width=\linewidth,trim=10 0 10 0]{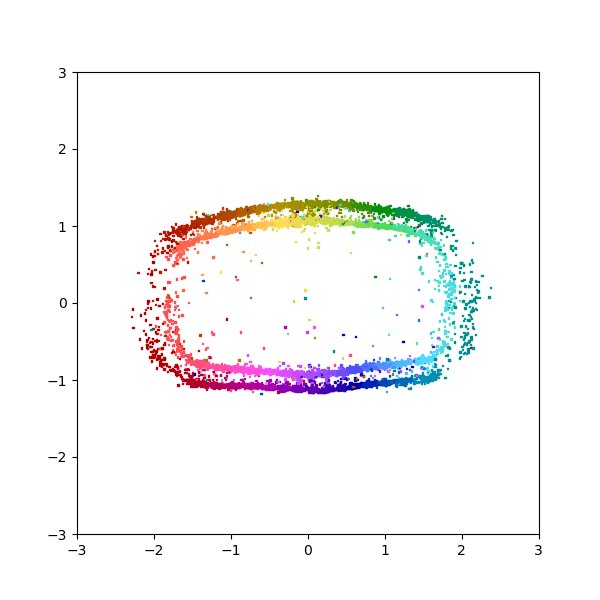}
        \caption{Ours}
    \end{subfigure}
    \raisebox{-63pt}{\includegraphics[width=0.0475\linewidth,trim=20 10 20 10]{figures/visualization/color_bar.png}}
    \caption{Visualizing the disentanglement with test sample distributions.}
    \label{fig:supp_plots}
    \vspace{-0.3cm}
\end{figure*}

Our proposed methods of Negative Log Unlikelihood and sample-space classification aim to improve the stability of encoder-classifier adversarial training. Here we evaluate the effectiveness of these two schemes. Specifically, we track the change of code distribution. We take snapshots of the network at fixed intervals during training. The whole test dataset is encoded, and we compute the average squared distance in the code space, from the code of each sample to the code of the same sample in the previous snapshot. In stable training, the encoder should keep the distance small while still finding a good distribution.

We train three variants of Stage I on the 3D chairs dataset with rotation unknown: one standard variant (proposed), one where the adversarial objective is maximizing the NLL loss of the classifier, and one where the classifier is an MLP operating in the code space, with four hidden layers of size 512. 

The code distribution is computed every 10,000 iterations until iteration 400,000. The average squared code distance in the unknown code space every 10,000 iterations apart is shown in Figure \ref{fig:code_distance}, in logarithm scale. We can see that both NLU and sample-space classification contributed to reducing the fluctuation in code space.

\subsection{Adversarial Classifier and Condition on Unlabeled Code in Stage II}

We evaluate the effectiveness of adversarial classifiers in Stage II compared to non-adversarial ones and examine the necessity of the code distance loss term. We train three variants of Stage II on the \textit{MNIST} dataset: one standard variant (proposed), one without NLU term (remove NLU term from Equation 5e in the main text) so that the classifiers do not try to distinguish generated samples from real ones in the same class, and one without code distance term (remove code distance from Equation 5g in the main text) so that the network does not explicitly preserve the unknown factor.

The initial network weights of $E$ and $G$ for three configurations are inherited from the same Stage I training run so that the result is only affected by Stage II training. We compute the mean squared reconstruction loss and mutual information gap for the final models in Table \ref{tab:ablation_stage2}. By using the adversarial Stage II classifier and adding the code distance term, we are able to improve both disentanglement (MIG) and reconstruction (MSE).


\section{Effect of Factor Correlation}
\label{sec:correlation}
In the datasets used for evaluation, the factors are independent of each other. In particular, in the 3D Shapes and the 3D Chairs datasets, every combination of labels occurred exactly once. In this section we would like to explore the behavior of our method when some of the factors are correlated. For better control, we construct a dataset where the correlation is exactly known: in the 3D Shapes dataset, each of the three color factors has $10$ possible values, numbered $0$ to $9$. We take the subset of the dataset consisting of all images whose \textit{floor hue} and \textit{wall hue} differ by $0$ or $\pm 1$, modulo $10$.

\subsection{Correlation Between Two Labeled Factors}
The first case is when correlation exists between two of the labeled factors. The semantics of the labeled factors in our networks is enforced to strictly follow the semantics of the labels, so it is expected that our network would behave in the same way as when the labeled factors did not have correlation.

\begin{figure}
    \centering
    \begin{subfigure}{0.45\linewidth}
        \includegraphics[width=\linewidth]{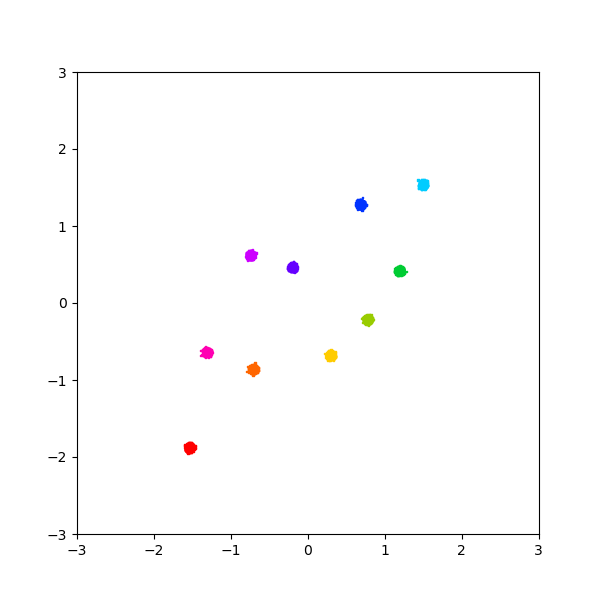}
        \caption{Floor hue}
    \end{subfigure}
    \begin{subfigure}{0.45\linewidth}
        \includegraphics[width=\linewidth]{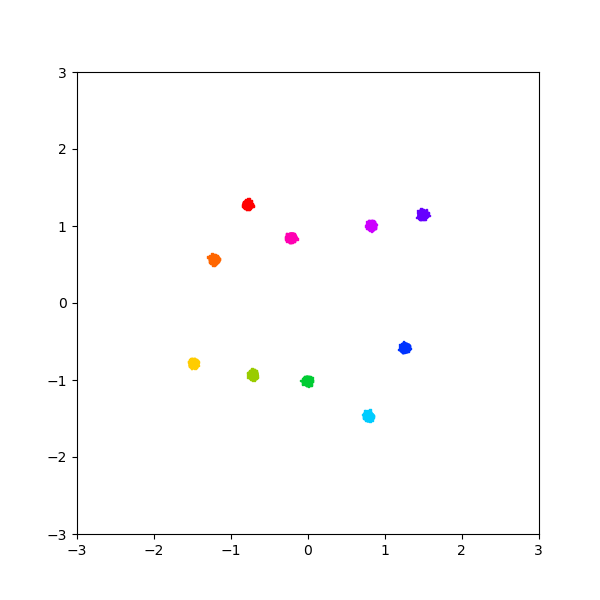}
        \caption{Wall hue}
    \end{subfigure}
    \caption{Distribution of test samples in the code space of the two correlated labeled factors.}
    \label{fig:corr1}
\end{figure}

We train our model on the correlated subset with \textit{object hue} being the unknown factor, so that the two correlated factors are both labeled. The distribution of test samples in the two correlated factors is shown in Figure \ref{fig:corr1}. As can be seen, our network successfully learns to encode \textit{floor hue} and \textit{wall hue} as labeled.

While the behavior of the network remains the same, the MIG decreases: due to correlation the mutual information between \textit{floor hue} and \textit{wall hue} is now $\ln 10-\ln 3$ instead of $0$, and a perfect set of encoders would produce an MIG of
\begin{equation}
    \frac{1}{6}\left(2\left(1-\frac{\ln10-\ln3}{\ln10}\right)+4\right)\approx 0.8257
\end{equation}

In comparison, our method gives an MIG of $0.8026$.

\subsection{Correlation Between Labeled and Unknown Factors}
The situation is more complicated when there is correlation between the unlabeled factor and the labeled factors. Our goal is for the unknown encoder to not encode any information about the labeled factors, which is to say, the conditional distribution of the unknown encoder, given the labeled factors, should be the same regardless of the value of the labeled factors. We train our model with \textit{floor hue} being the unknown factor. In this case, since for any value of \textit{wall hue} there are exactly three possible values of \textit{floor hue}, it can be predicted that our unknown encoder should discover a factor with three discrete values, such that for any \textit{wall hue}, each of the three \textit{floor hues} is encoded as a different value. Note that this discovered factor is not necessarily the ``hue difference'' of the floor and the wall: there is no guarantee that the three values of this factor correspond to the hue difference being $-1$, $0$ and $1$ consistently, independent of other factors. The mapping from hue difference to the value of the discovered factor can vary according to \textit{wall hue}. The only guarantee is that for the same \textit{wall hue}, different \textit{floor hues} correspond to different values of the discovered factor.

\begin{figure}
    \centering
    \includegraphics[width=0.45\linewidth]{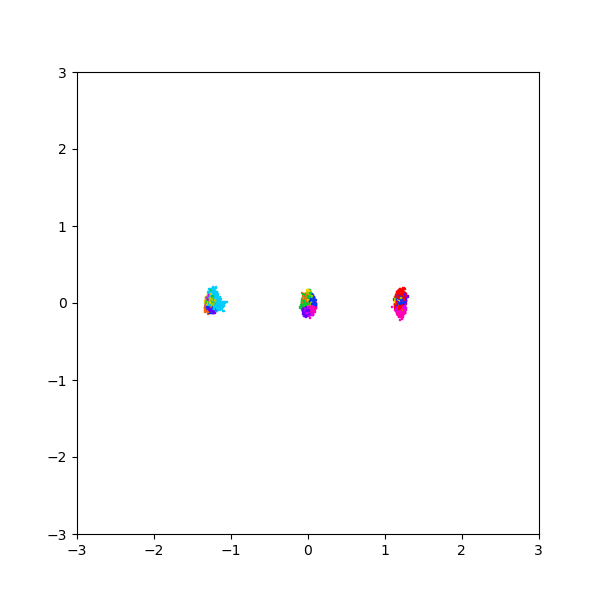}
    \caption{Distribution of test samples in the code space of the unknown encoder when the unknown factor and the labeled factors are correlated.}
    \label{fig:corr2}
\end{figure}

The distribution of the samples in the unknown encoder's code space, colored by \textit{floor hue}, is shown in Figure \ref{fig:corr2}. Three clusters can be clearly seen.

In general, we can conclude that if the intended semantics of the unknown factor is correlated to the labeled factors, then the factor discovered by our method would have different semantics. This may or may not be a desirable outcome, but it shows that our method is highly effective in ensuring the independence between the discovered unknown factor and the labeled factors.
\begin{figure*}
    \centering
    \hspace{0.02\linewidth}
    \begin{subfigure}{0.45\linewidth}
        \includegraphics[width=\linewidth]{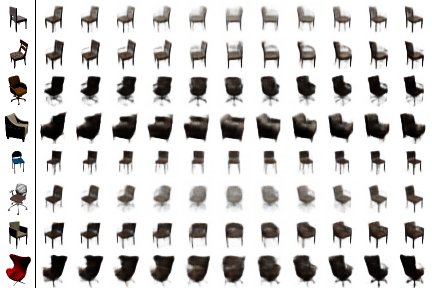}
        \caption{$\beta$-VAE~\cite{DBLP:conf/iclr/HigginsMPBGBML17}}
    \end{subfigure}
    \hfill
    \begin{subfigure}{0.45\linewidth}
        \includegraphics[width=\linewidth]{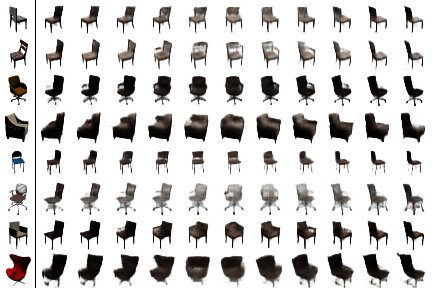}       
        \caption{Factor-VAE~\cite{DBLP:conf/icml/KimM18}}
    \end{subfigure}
    \hspace{0.02\linewidth}
    
    \hspace{0.02\linewidth}
    \begin{subfigure}{0.45\linewidth}
        \includegraphics[width=\linewidth]{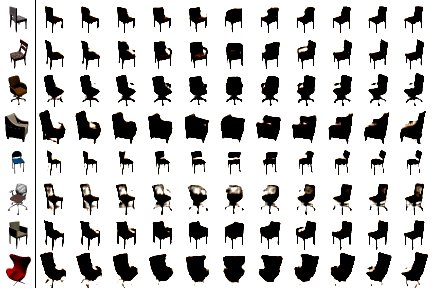}
        \caption{$\beta$-TCVAE~\cite{DBLP:conf/nips/ChenLGD18}}
    \end{subfigure}
    \hfill
    \begin{subfigure}{0.45\linewidth}
        \includegraphics[width=\linewidth]{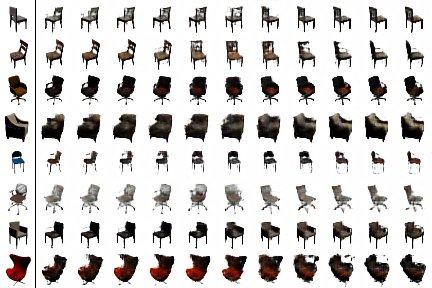}
        \caption{Pairwise-VAE~\cite{DBLP:conf/aaai/ChenB20}}
    \end{subfigure}
    \hspace{0.02\linewidth}
    
    \hspace{0.02\linewidth}
    \begin{subfigure}{0.45\linewidth}
        \includegraphics[width=\linewidth]{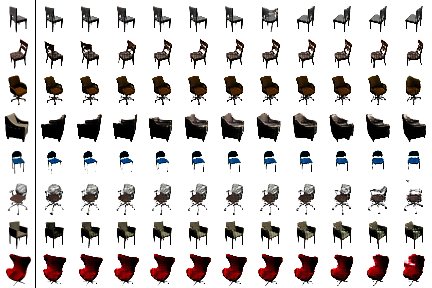}
        \caption{Lord~\cite{DBLP:conf/iclr/GabbayH20}}
    \end{subfigure}
    \hfill
    \begin{subfigure}{0.45\linewidth}
        \includegraphics[width=\linewidth]{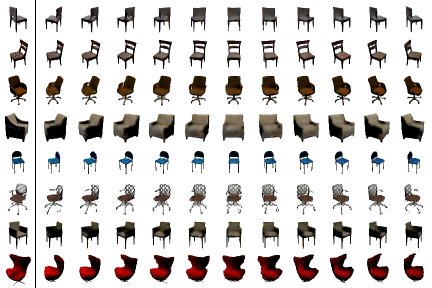}
        \caption{Ours}
    \end{subfigure}
    \hspace{0.02\linewidth}
    \caption{Additional results of manipulation comparison on \textit{3D Chairs} by uniformly sampling the latent codes depicting the azimuth rotation. The leftmost column shows the inputs.}
    \label{fig:res}
    \vspace{-0.3cm}
\end{figure*}

\begin{figure*}[t]
  \centering
  \includegraphics[width=\linewidth]{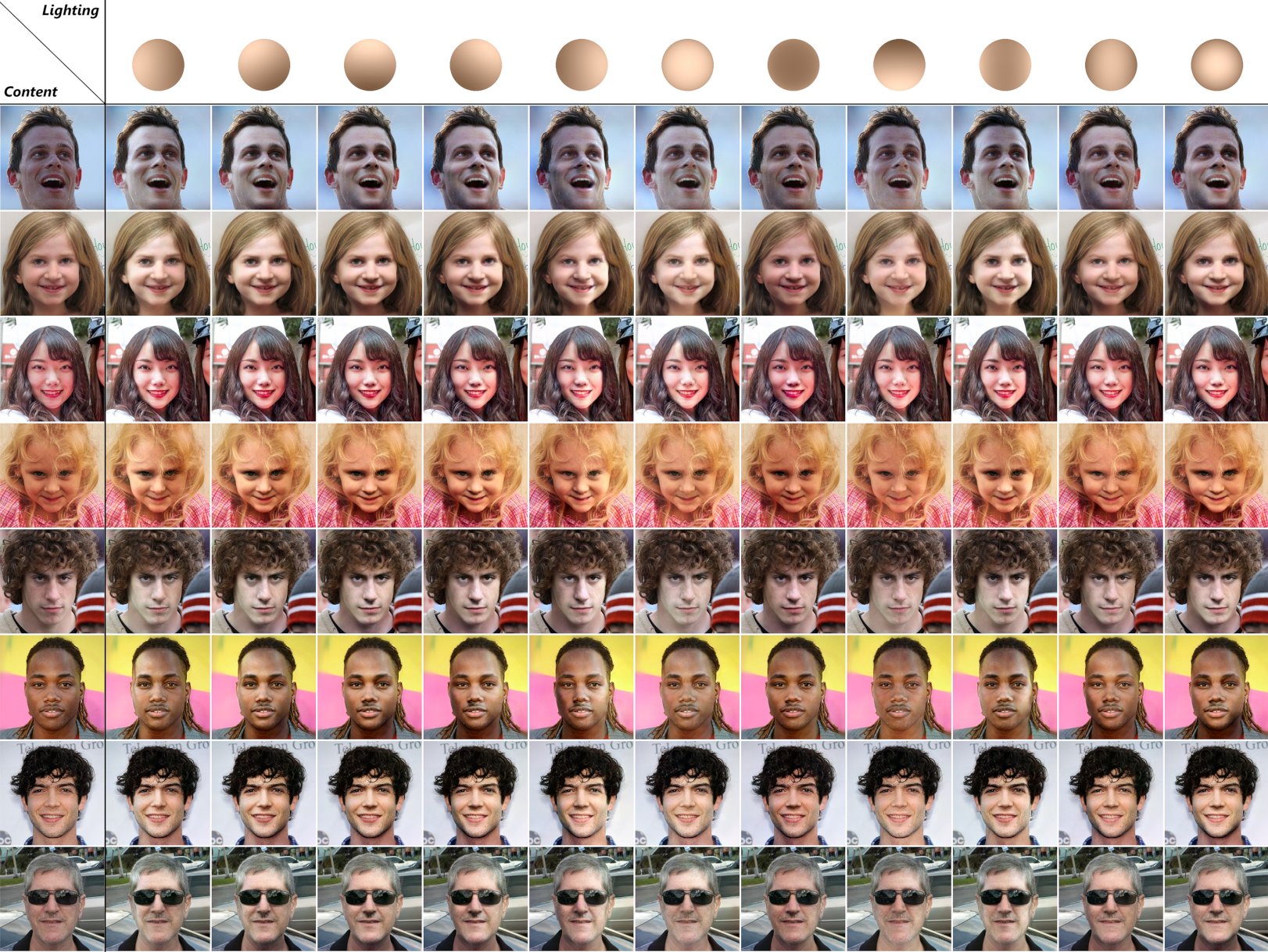}
  \caption{Additional portrait relighting results.}
  \label{fig:more_lighting}
\end{figure*}

\section{Comparisons}
\label{sec:cmp}

\subsection{Visualization}

In Figure \ref{fig:supp_plots}, for each of the methods compared, we plot the distribution of test samples of the \textit{3D Chairs} dataset in the code space, projected onto the two dimensions having the largest mutual information with the rotation label. In $\beta$-VAE~\cite{DBLP:conf/iclr/HigginsMPBGBML17} and Factor-VAE~\cite{DBLP:conf/icml/KimM18}, no clear color pattern can be observed. In $\beta$-TCVAE and Lord~\cite{DBLP:conf/iclr/GabbayH20}, samples with the same rotation are somewhat close together but there is no meaningful order between different rotations. Pairwise-VAE~\cite{DBLP:conf/aaai/ChenB20} and our method can arrange the azimuth angle correctly into a ring, but the structure is more clear in our method, and also, we can distinguish two slightly different elevation angles, which are not distinguished by any other method.

\subsection{Results}
We provide more results of manipulating the latent code related to azimuth rotations and generate the images using different methods in Figure \ref{fig:res}. 

\section{Downstream Tasks}
\label{sec:supp_app}

\subsection{Portrait Relighting}
In this task, the labeled factor is lighting, represented by the coefficients of spherical harmonics up to second-order, which are $9$ real-valued numbers. We show that our disentanglement framework can handle such continuous labels. More portrait relighting results are shown in Figure \ref{fig:more_lighting}.

\subsection{Anime Style Transfer}
\begin{figure*}
  \centering
  \includegraphics[width=0.85\linewidth]{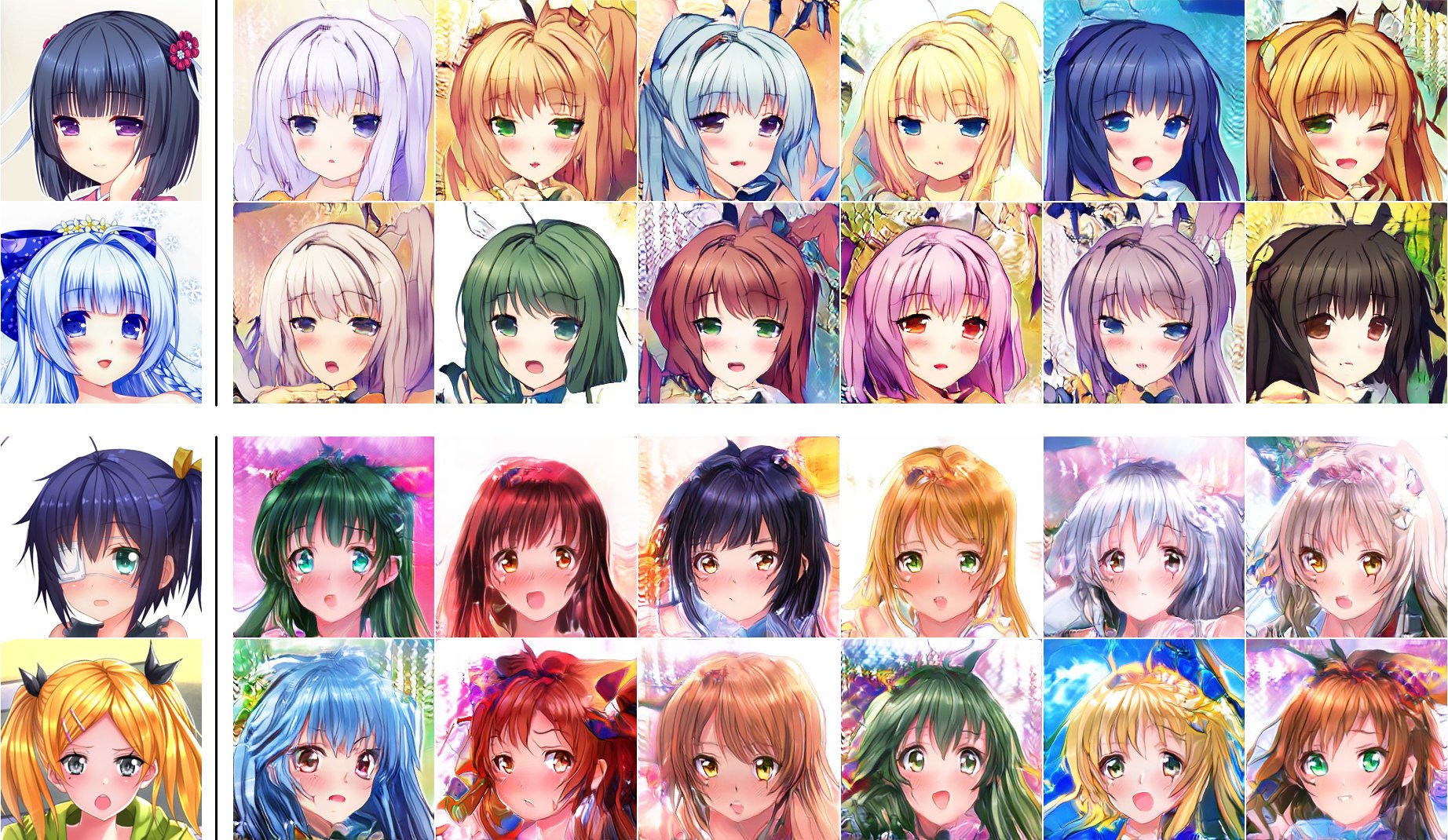}
  \caption{Generated samples with fixed style and random content. Left column shows four input examples of two different artists.}
  \label{fig:anime_fix_style}
\end{figure*}

\begin{figure*}
  \centering
  \includegraphics[width=0.85\linewidth]{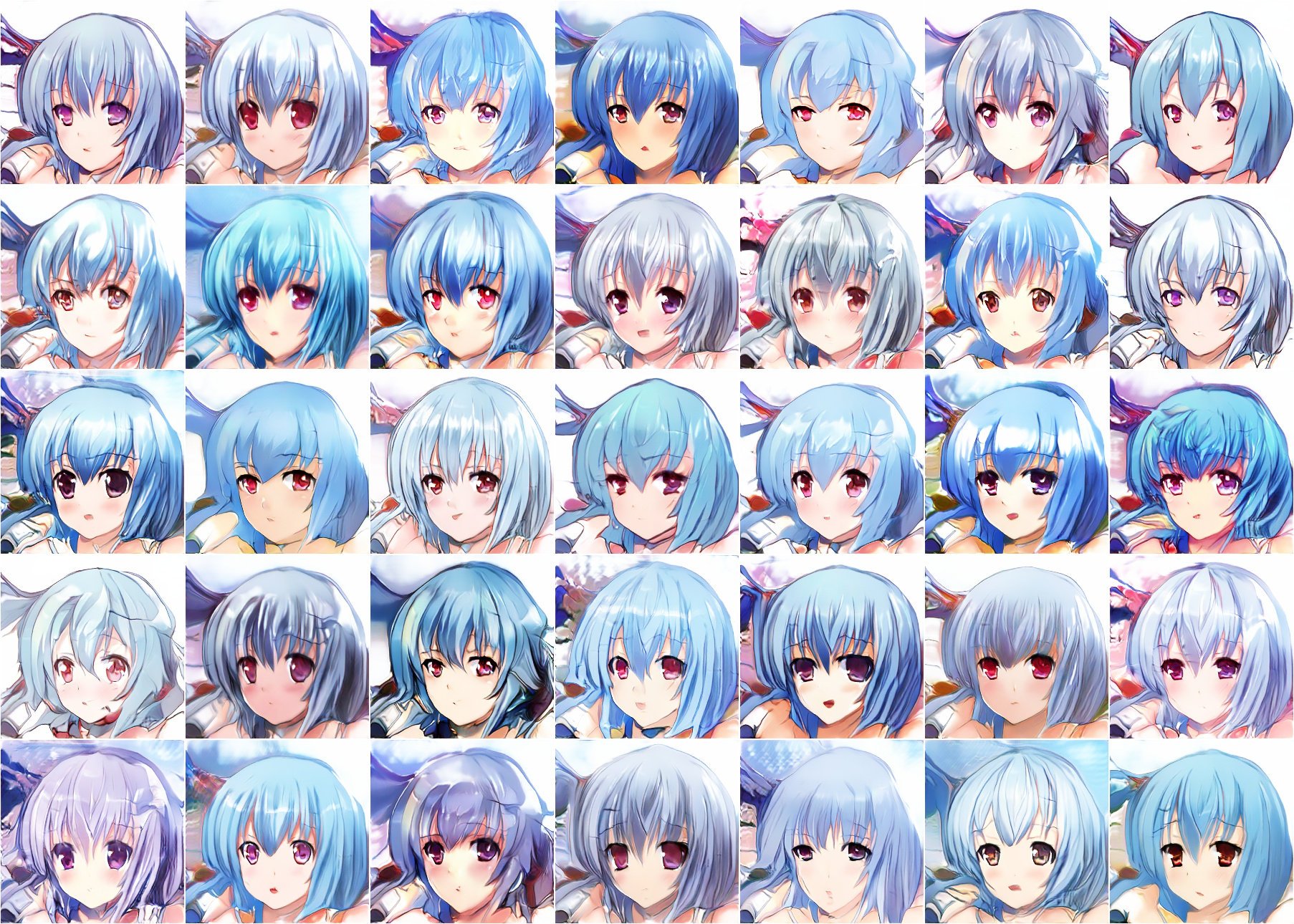}
  \caption{Generated samples with fixed content and random style. The content code is fixed for all results while style codes randomly sample from the style distribution.}
  \label{fig:anime_fix_content}
\end{figure*}

Figure~\ref{fig:anime_fix_style} and Figure~\ref{fig:anime_fix_content} present more results of anime style transfer generated by fixing either style or content. In Figure~\ref{fig:anime_fix_style}, we show the styles of the results are consistently faithful to the input. In Figure \ref{fig:anime_fix_content}, we explore the diversity of styles our network can learn and demonstrate the ability to generate the same content in different styles where facial shapes, appearances, and aspect ratios are captured.

 \begin{table}{}
\centering
    \begin{tabular}{cccc}
        \toprule
        Method & ID $\uparrow$ & Pose $\uparrow$ & Exp. $\downarrow$ \\
        \midrule
        X2face~\cite{DBLP:conf/eccv/WilesKZ18} & 0.635 & 0.302 & 0.448 \\
        First-order~\cite{DBLP:conf/nips/SiarohinLT0S19} & \underline{0.770} & 0.822 & 0.274 \\
        Ours \textit{w/o} disentanglement & 0.715 & \textbf{0.862} & \textbf{0.208}\\
        Ours \textit{w/} disentanglement & \textbf{0.776} & \underline{0.840} & \underline{0.243}\\
        \bottomrule
    \end{tabular}
    \caption{Quantitative comparison of methods for cross-subject face reenactment on the VoxCeleb2 testing dataset between our method with and without landmark disentanglement, and two one-shot methods~\cite{DBLP:conf/eccv/WilesKZ18}, and~\cite{DBLP:conf/nips/SiarohinLT0S19}. Note that our results without disentanglement are generated using the ground truth landmarks from the driving frame so their poses/expressions in the results should be the closest to those in the driving frame, but their identities are very different from the target person. Our method with disentanglement achieves the best in identity preservation and the second-best in poses/expressions.}
    \label{tab:compare_one}
\end{table}


\subsection{Landmark-Based Face Reenactment}
The landmark-based face reenactment can generate the face motion of one target person from a single image with another source subject’s facial expressions and head pose from a driving video. The generated results should preserve the source poses/expressions while matching the target identity. With only the identity label known, we can disentangle unknown poses/expressions from identities and synthesize new landmarks by changing identities in the source landmarks to the target.

\begin{figure*}[h]
  \centering
  \setlength{\tabcolsep}{0.2mm}{
\begin{tabular}{cc|cc|cc}
  \includegraphics[width=0.15\linewidth]{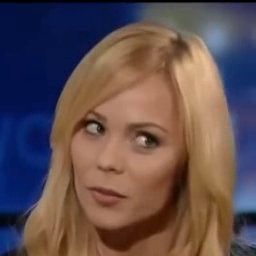} &
  \includegraphics[width=0.15\linewidth]{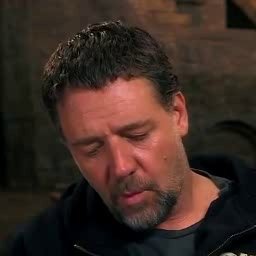} &
  \includegraphics[width=0.15\linewidth]{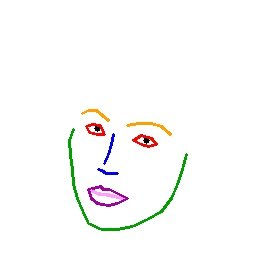} &
  \includegraphics[width=0.15\linewidth]{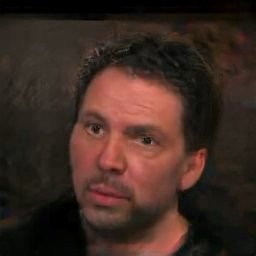} &
  \includegraphics[width=0.15\linewidth]{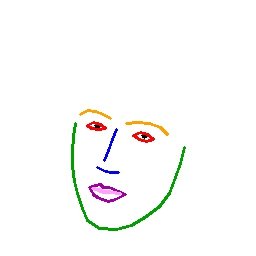} &
  \includegraphics[width=0.15\linewidth]{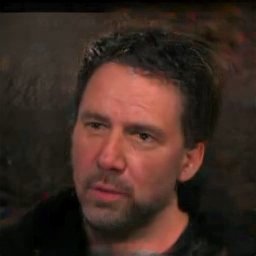} \\
  \includegraphics[width=0.15\linewidth]{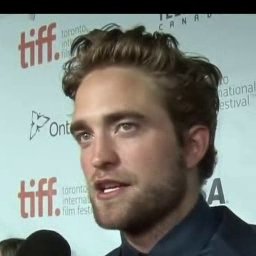} &
  \includegraphics[width=0.15\linewidth]{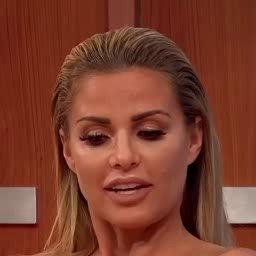} &
  \includegraphics[width=0.15\linewidth]{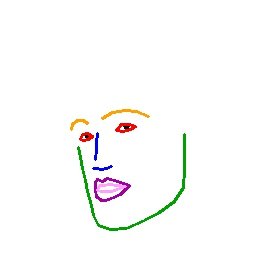} &
  \includegraphics[width=0.15\linewidth]{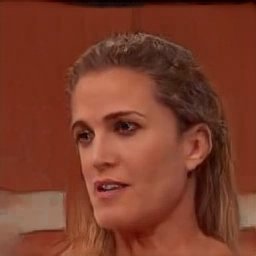} &
  \includegraphics[width=0.15\linewidth]{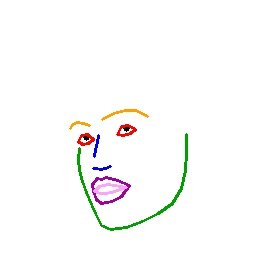} &
  \includegraphics[width=0.15\linewidth]{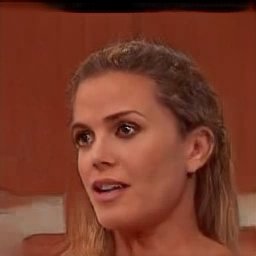} \\
  \includegraphics[width=0.15\linewidth]{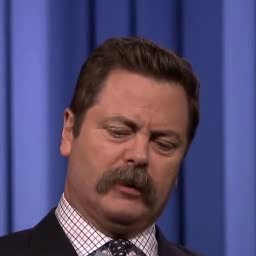} &
  \includegraphics[width=0.15\linewidth]{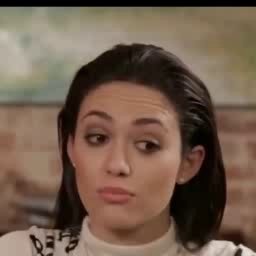} &
  \includegraphics[width=0.15\linewidth]{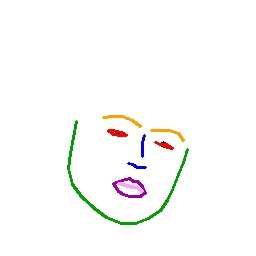} &
  \includegraphics[width=0.15\linewidth]{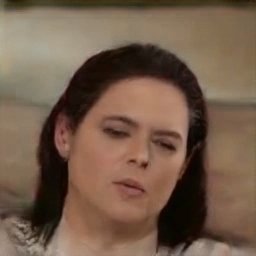} &
  \includegraphics[width=0.15\linewidth]{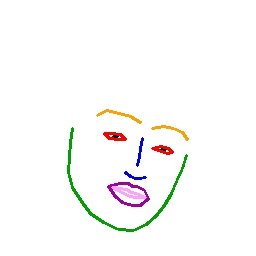} &
  \includegraphics[width=0.15\linewidth]{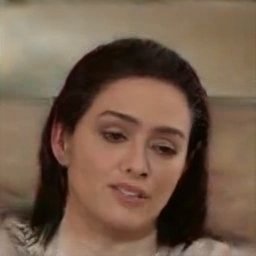} \\
  \includegraphics[width=0.15\linewidth]{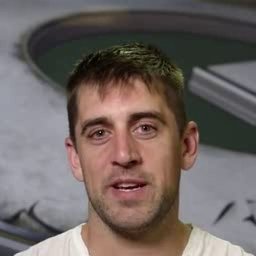} &
  \includegraphics[width=0.15\linewidth]{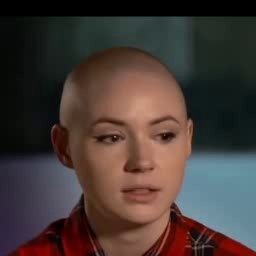} &
  \includegraphics[width=0.15\linewidth]{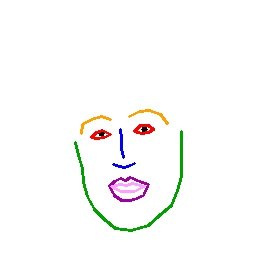} &
  \includegraphics[width=0.15\linewidth]{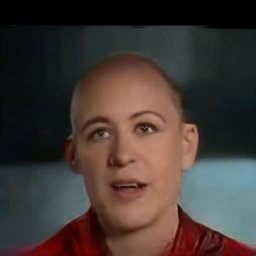} &
  \includegraphics[width=0.15\linewidth]{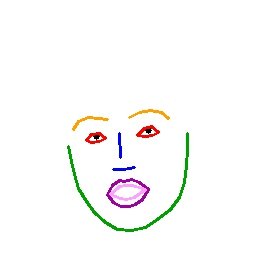} &
  \includegraphics[width=0.15\linewidth]{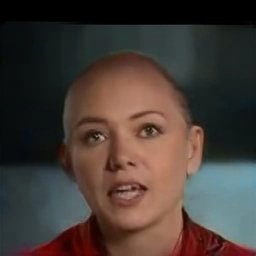} \\
  \includegraphics[width=0.15\linewidth]{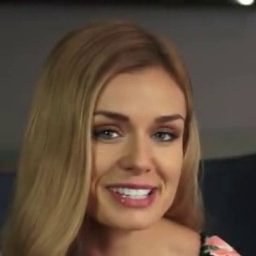} &
  \includegraphics[width=0.15\linewidth]{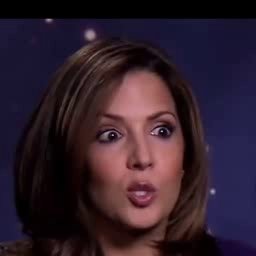} &
  \includegraphics[width=0.15\linewidth]{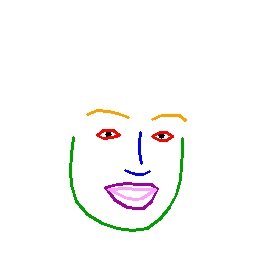} &
  \includegraphics[width=0.15\linewidth]{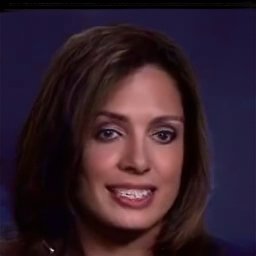} &
  \includegraphics[width=0.15\linewidth]{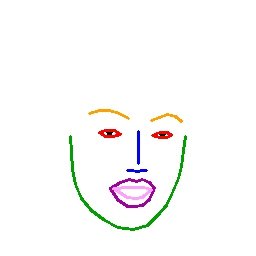} &
  \includegraphics[width=0.15\linewidth]{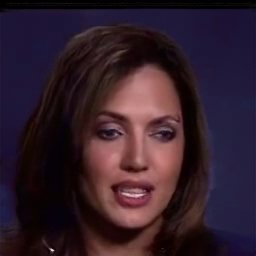} \\
  \includegraphics[width=0.15\linewidth]{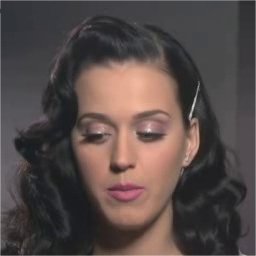} &
  \includegraphics[width=0.15\linewidth]{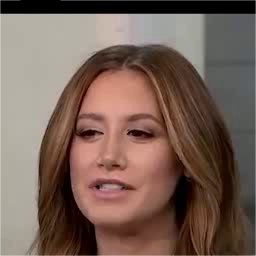} &
  \includegraphics[width=0.15\linewidth]{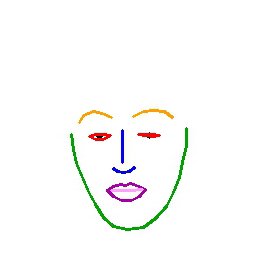} &
  \includegraphics[width=0.15\linewidth]{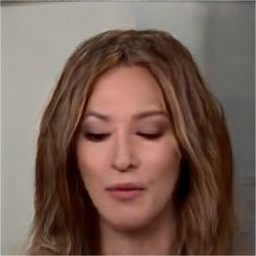} &
  \includegraphics[width=0.15\linewidth]{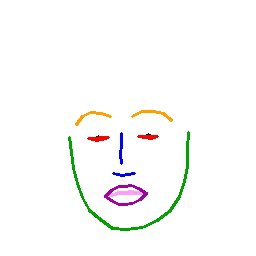} &
  \includegraphics[width=0.15\linewidth]{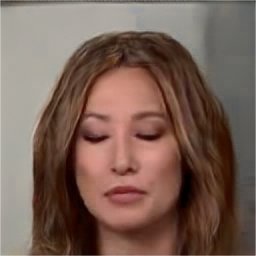} \\
  \includegraphics[width=0.15\linewidth]{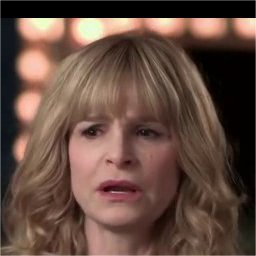} &
  \includegraphics[width=0.15\linewidth]{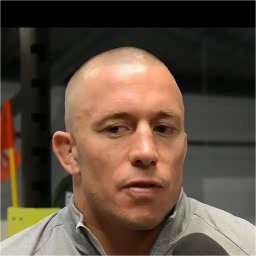} &
  \includegraphics[width=0.15\linewidth]{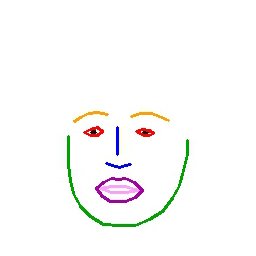} &
  \includegraphics[width=0.15\linewidth]{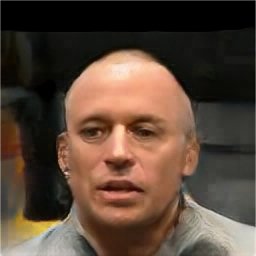} &
  \includegraphics[width=0.15\linewidth]{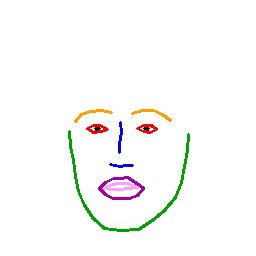} &
  \includegraphics[width=0.15\linewidth]{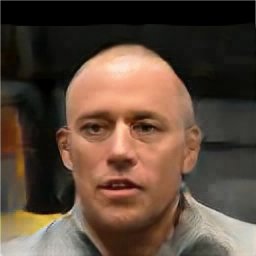} \\
  \hline
  Source & Target & \textit{w/o} & Result & \textit{w/} & Result \\
  &  & Disentanglement &  & Disentanglement & 
  \end{tabular}}
\caption{Qualitative comparison on face image reenactment between the translation results without (\textit{w/o}) and with (\textit{w/}) facial landmark disentanglement.}
	\label{fig:face_disentangle}
\end{figure*}

\begin{figure*}[htp]
  \centering
  \setlength{\tabcolsep}{0.2mm}{
\begin{tabular}{ccccc}
  \includegraphics[width=0.14\linewidth]{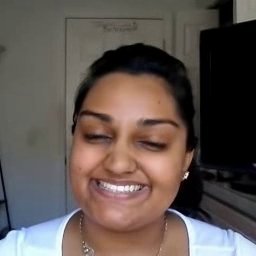} &
  \includegraphics[width=0.14\linewidth]{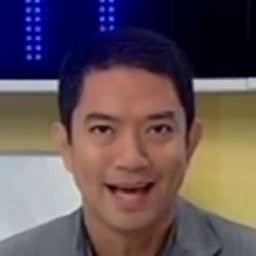} &
  \includegraphics[width=0.14\linewidth]{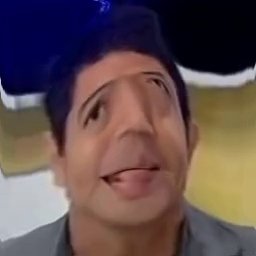} &
  \includegraphics[width=0.14\linewidth]{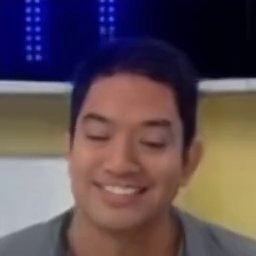} &
  \includegraphics[width=0.14\linewidth]{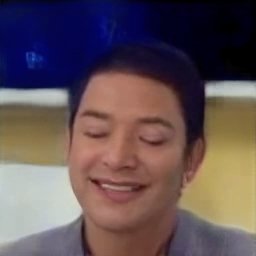} \\
  \includegraphics[width=0.14\linewidth]{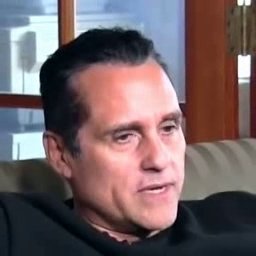} &
  \includegraphics[width=0.14\linewidth]{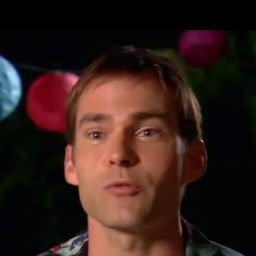} &
  \includegraphics[width=0.14\linewidth]{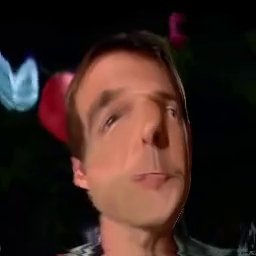} &
  \includegraphics[width=0.14\linewidth]{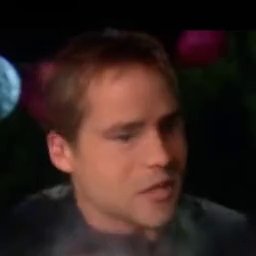} &
  \includegraphics[width=0.14\linewidth]{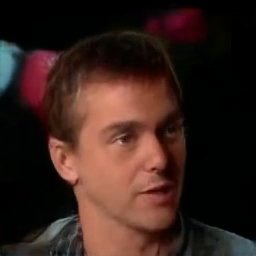} \\
  \includegraphics[width=0.14\linewidth]{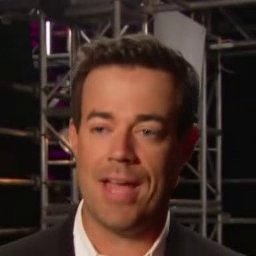} &
  \includegraphics[width=0.14\linewidth]{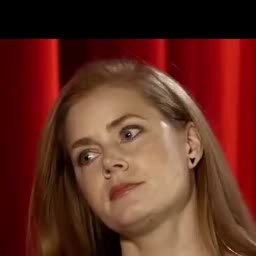} &
  \includegraphics[width=0.14\linewidth]{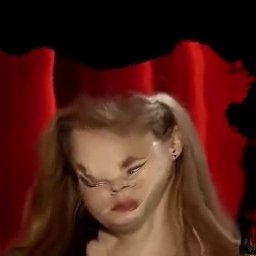} &
  \includegraphics[width=0.14\linewidth]{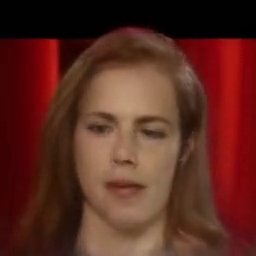} &
  \includegraphics[width=0.14\linewidth]{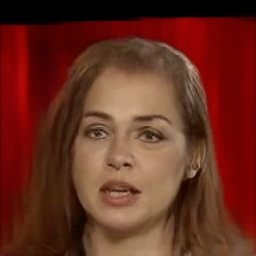} \\
  \includegraphics[width=0.14\linewidth]{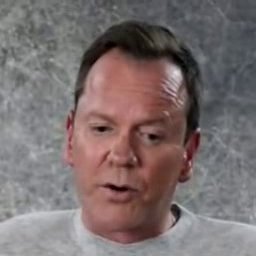} &
  \includegraphics[width=0.14\linewidth]{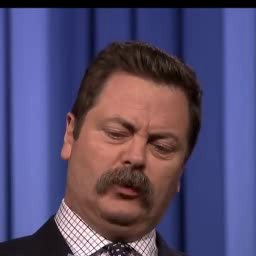} &
  \includegraphics[width=0.14\linewidth]{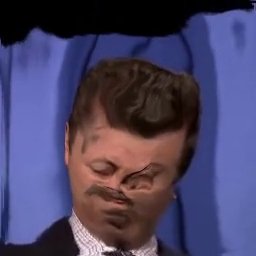} &
  \includegraphics[width=0.14\linewidth]{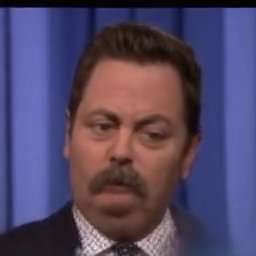} &
  \includegraphics[width=0.14\linewidth]{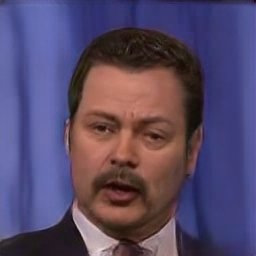}\\
  \includegraphics[width=0.14\linewidth]{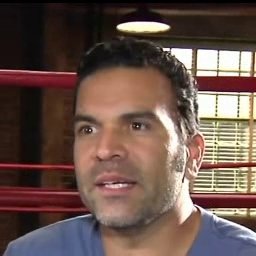} &
  \includegraphics[width=0.14\linewidth]{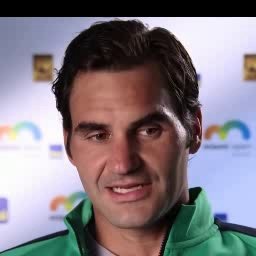} &
  \includegraphics[width=0.14\linewidth]{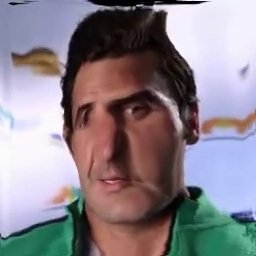} &
  \includegraphics[width=0.14\linewidth]{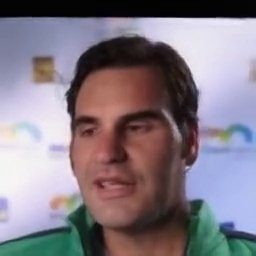} &
  \includegraphics[width=0.14\linewidth]{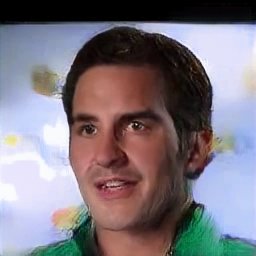} \\
  \includegraphics[width=0.14\linewidth]{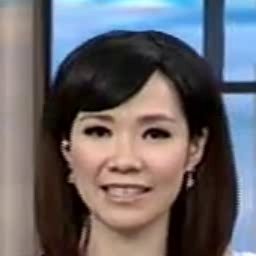} &
  \includegraphics[width=0.14\linewidth]{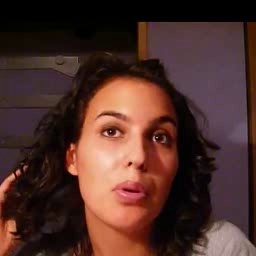} &
  \includegraphics[width=0.14\linewidth]{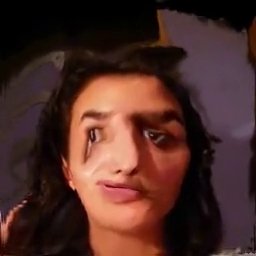} &
  \includegraphics[width=0.14\linewidth]{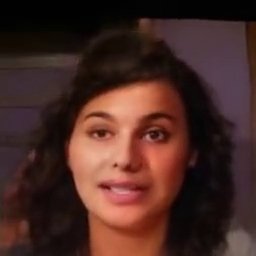} &
  \includegraphics[width=0.14\linewidth]{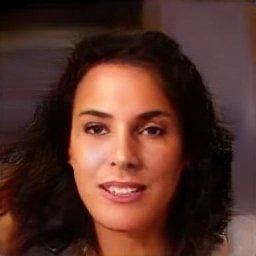} \\
  \includegraphics[width=0.14\linewidth]{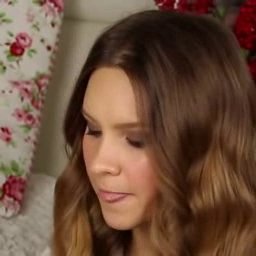} &
  \includegraphics[width=0.14\linewidth]{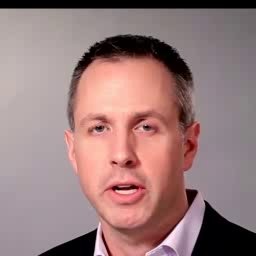} &
  \includegraphics[width=0.14\linewidth]{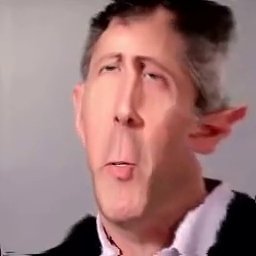} &
  \includegraphics[width=0.14\linewidth]{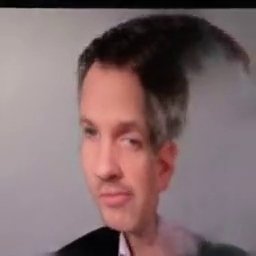} &
  \includegraphics[width=0.14\linewidth]{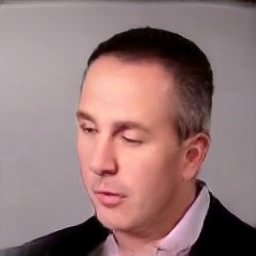} \\
  \includegraphics[width=0.14\linewidth]{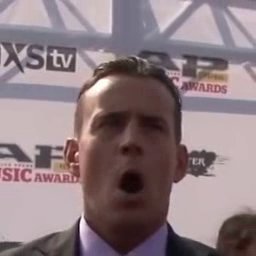} &
  \includegraphics[width=0.14\linewidth]{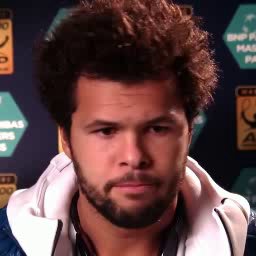} &
  \includegraphics[width=0.14\linewidth]{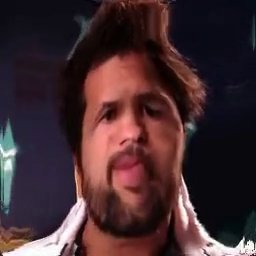} &
  \includegraphics[width=0.14\linewidth]{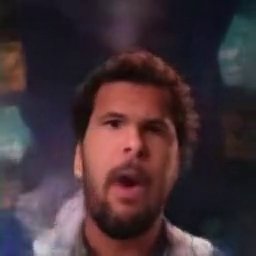} &
  \includegraphics[width=0.14\linewidth]{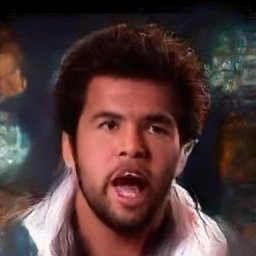} \\

  Source & Target & X2face~\cite{DBLP:conf/eccv/WilesKZ18} & First-order~\cite{DBLP:conf/nips/SiarohinLT0S19} & Ours \\
  \end{tabular}}
  \caption{Qualitative comparison on face image reenactment between our method and face motion transfer networks: X2face~\cite{DBLP:conf/eccv/WilesKZ18}, and First-order~\cite{DBLP:conf/nips/SiarohinLT0S19} using images from Voxceleb2~\cite{chung2018voxceleb2} and FaceForensics++ \cite{DBLP:conf/iccv/RosslerCVRTN19}.}
	\label{fig:face_comp}
\end{figure*}

We train the landmark-to-image network on the training dataset from VoxCeleb2~\cite{chung2018voxceleb2} which is processed by cropping a $256\times 256$ face image and extracting its landmarks from each frame. In total, the training data contains $52,112$ videos for $1,000$ randomly selected subjects. We test on a video dataset of $8,000$ frames for $80$ pairwise subjects ($100$ frames per video) randomly sampled from the VoxCeleb2 testing data,

We present more qualitative comparison in Figure~\ref{fig:face_disentangle} between with and without landmark disentanglement and in Figure~\ref{fig:face_comp} against two of the state-of-the-art one-shot face reenactment methods, \textit{i.e.} X2face~\cite{DBLP:conf/eccv/WilesKZ18} and First-order~\cite{DBLP:conf/nips/SiarohinLT0S19}. Figure~\ref{fig:face_disentangle} compares the results using source landmarks without disentanglement and synthesized landmarks with disentanglement and shows that landmark disentanglement leads to better identity preservation. Figure~\ref{fig:face_comp} shows quantitative comparisons with two one-shot face reenactment baselines and demonstrates our method can better generalize to unseen subjects with more consistent quality under a large variety of poses/expressions.

Besides, we conduct a quantitative comparison as shown in Table~\ref{tab:compare_one} to measure the accuracy of disentanglement. We evaluate the results using three metrics (\textit{ID}, \textit{Pose}, and \textit{Exp.}) for identity, pose, and expression respectively. \textit{ID} measures the identity similarity between the resulting face and the target person by computing cosine similarity between their embedding vectors of the face recognition network VGGFace2~\cite{DBLP:conf/fgr/CaoSXPZ18}. \textit{Pose} measures the similarity between the resulting head pose and the source subject's pose in the driving frame by computing cosine similarity between their head rotations in radians around the X, Y, and Z axes estimated by OpenFace~\cite{DBLP:conf/fgr/BaltrusaitisZLM18}. \textit{Exp.} measures the difference between the resulting expression and the source expression in the driving frame by computing the L2 distance of their intensities of corresponding facial action units detected by OpenFace. Note that our results generated without disentanglement directly use the landmarks from the driving frame so their poses/expressions in the results are the closest to those in the driving frame but their identities mismatch the target person. In contrast, our method with disentanglement largely increases the identity accuracy while achieving comparable accuracy in poses/expressions.

\subsection{Skeleton-Based Body Motion Retargeting}
We train our skeleton disentanglement network on the dataset of Mixamo~\cite{mixamo}. It contains synthetic 3D skeletons of approximately $800$ unique motion sequences each of $36$ distinct characters ground truth labels for motion, view, and pose. We follow the same setting in~\cite{DBLP:conf/cvpr/YangZW00ZL20} to use $32$ of these characters with $800$ sequences of each one for training and the rest for testing. There is no overlap in motion sequence and character between training and testing. In training, we project the 3D joint coordinates on-the-fly onto 2D with viewing angles chosen randomly. The labeled factors are the identity of the character and the view angle and the unknown factor is the motion. The network sees only one frame at a time instead of a motion sequence. 

In Table~\ref{tab:skeleton_cmp}, we quantitatively compare with the state-of-the-art methods ( \textit{i.e.}, \cite{DBLP:conf/cvpr/YangZW00ZL20} and \cite{DBLP:journals/tog/AbermanWLCC19}), which are task-specific, focusing on skeleton disentanglement. The method of \cite{DBLP:conf/cvpr/YangZW00ZL20} is unsupervised while the method of \cite{DBLP:journals/tog/AbermanWLCC19} utilizes full supervision. We perform evaluations on the same held-out test set from Mixamo (with ground truth available) using MSE and MAE as the metrics, reported in the original scale of the data. Our method, using weak supervision, outperforms both methods in terms of numerical joint position error but does not rely on any domain-specific prior knowledge.

\begin{table}
\caption{Quantitative comparison with the state-of-the-art methods for skeleton disentanglement on Mixamo.}
\centering
 
  \begin{tabular}{ccc}
  \toprule
  Method & MSE $\downarrow$ & MAE $\downarrow$ \\
  \midrule
  \cite{DBLP:conf/cvpr/YangZW00ZL20} & 0.0131 & 0.0673 \\
  \cite{DBLP:journals/tog/AbermanWLCC19} & 0.0151 & 0.0749 \\
  Ours & \textbf{0.0056} & \textbf{0.0498} \\
  \bottomrule
 \end{tabular} 
 
 \label{tab:skeleton_cmp}
 \end{table}

\begin{figure*}
    \centering
    \begin{subfigure}{0.8\linewidth}
        \includegraphics[width=\linewidth]{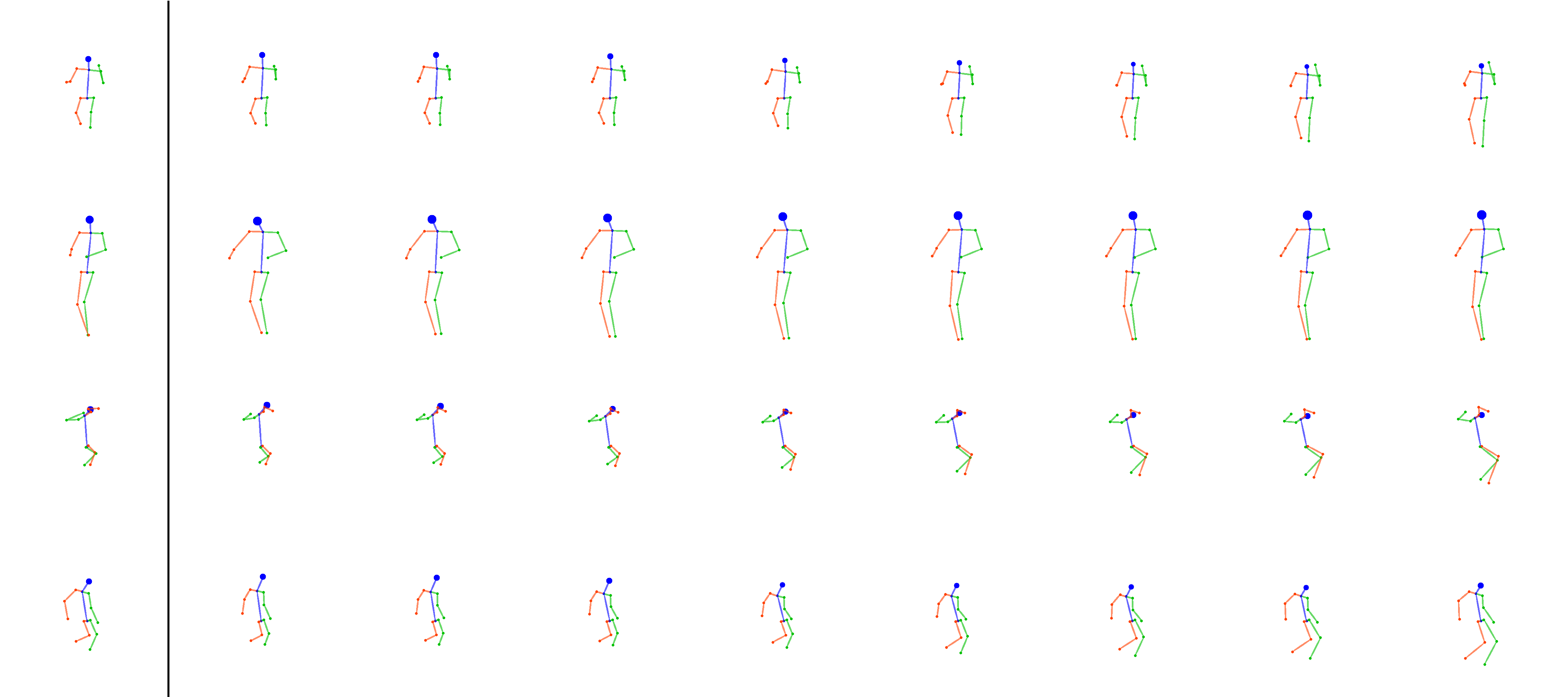}
        \caption{Identity synthesis with view and motion fixed}
    \end{subfigure}\hfill
    \begin{subfigure}{0.8\linewidth}
        \includegraphics[width=\linewidth]{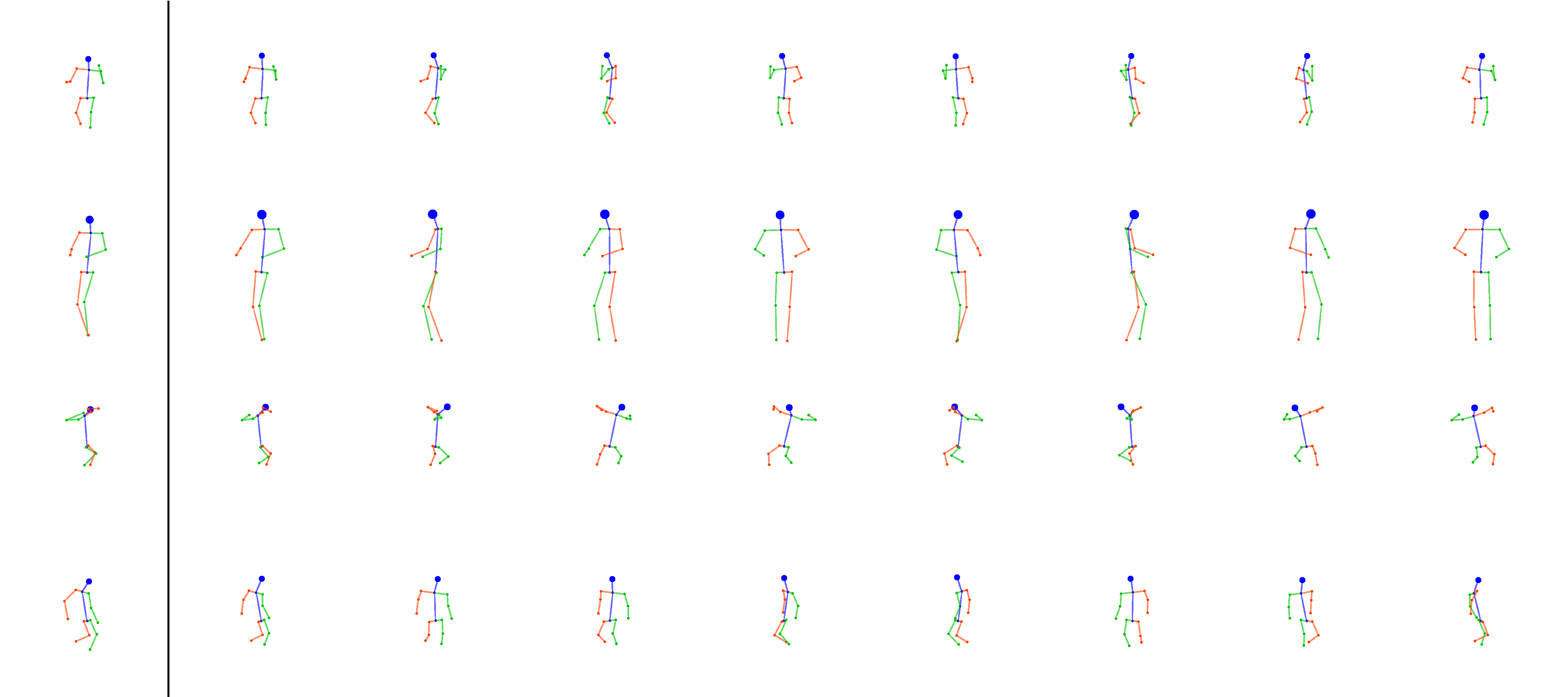}
        \caption{View synthesis with identity and motion fixed}
    \end{subfigure}\hfill
    \begin{subfigure}{0.8\linewidth}
        \includegraphics[width=\linewidth]{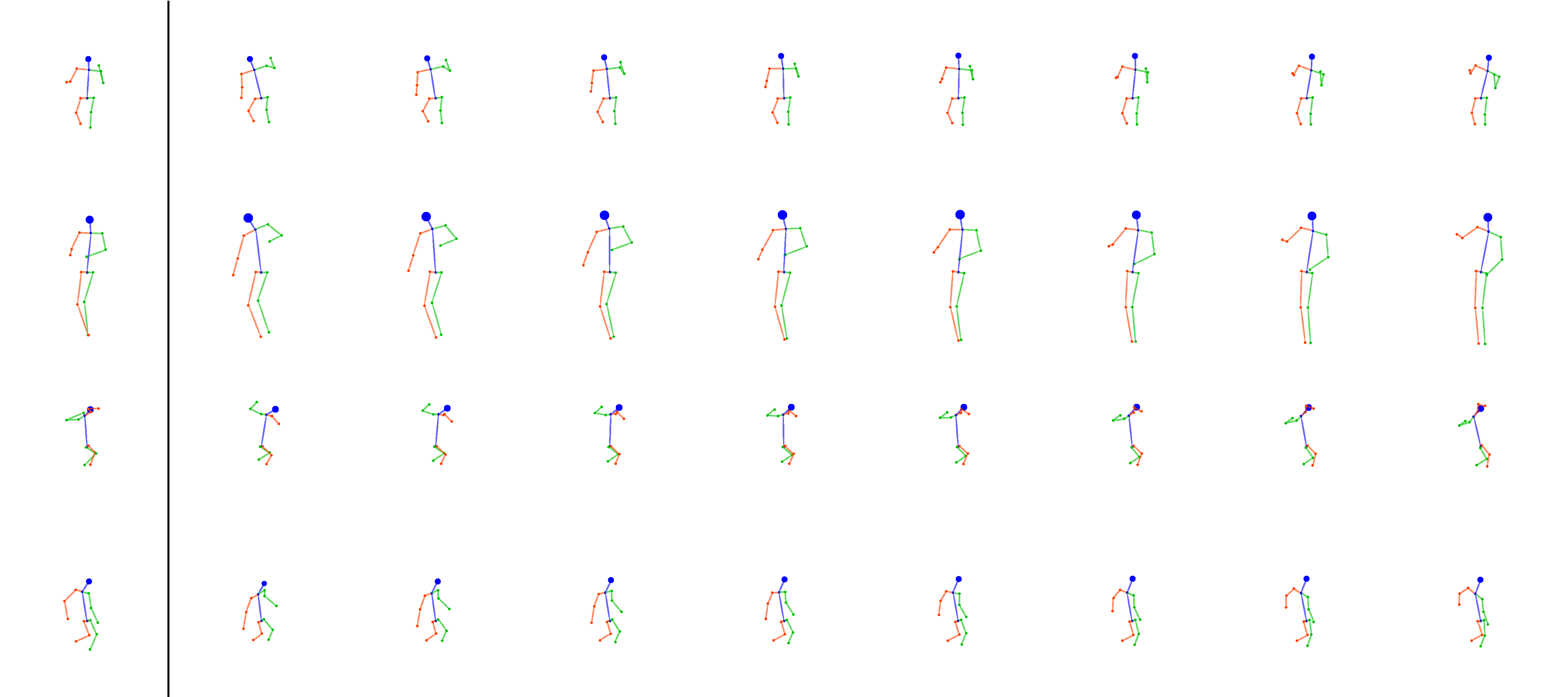}
        \caption{Motion synthesis with identity and view fixed}
    \end{subfigure}\hfill
    \caption{Novel synthesis of identity, view and motion. Left column shows the input and the rest represents the generated skeletons by individually controlling identity, view and motion while fixing the others.}
    \label{fig:app_skeleton_control}
    \vspace{-0.3cm}
\end{figure*}

Figure \ref{fig:app_skeleton_control} shows that we can independently control identity, view and motion, synthesizing 2D skeletons with novel identity, view and motion while preserving the remaining factors unchanged.

\begin{figure*}[ht]
    \centering
    \setlength{\tabcolsep}{0.5mm}
    \begin{tabular}{c|cccccccc} &
    \includegraphics[width=0.105\linewidth]{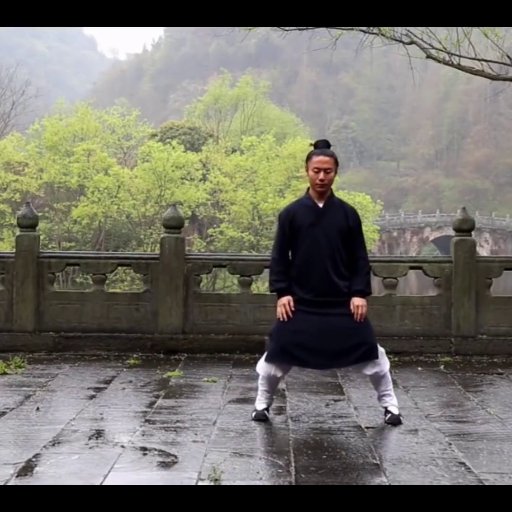} &
    \includegraphics[width=0.105\linewidth]{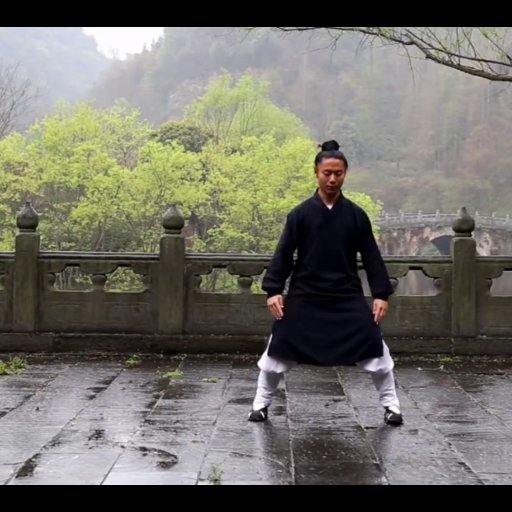} &
    \includegraphics[width=0.105\linewidth]{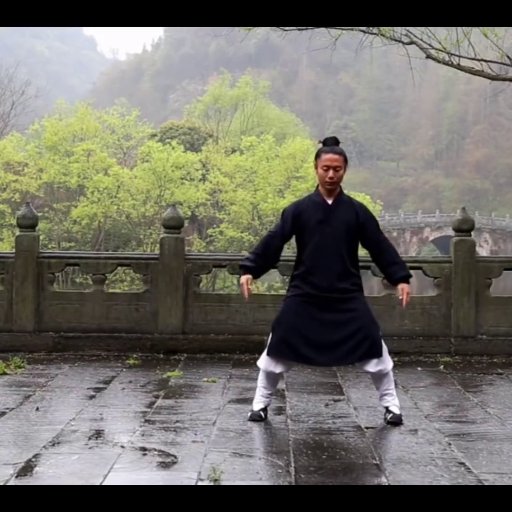} &
    \includegraphics[width=0.105\linewidth]{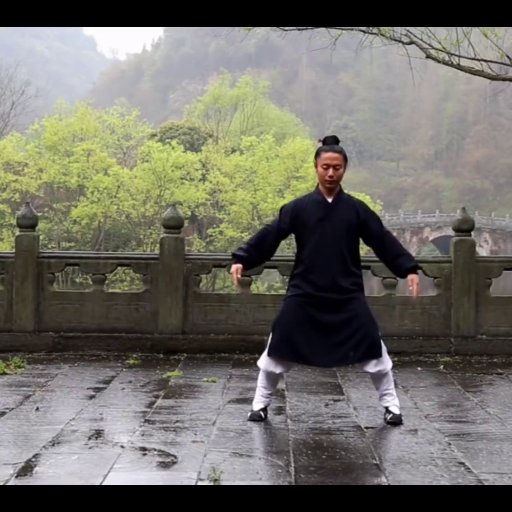} &
    \includegraphics[width=0.105\linewidth]{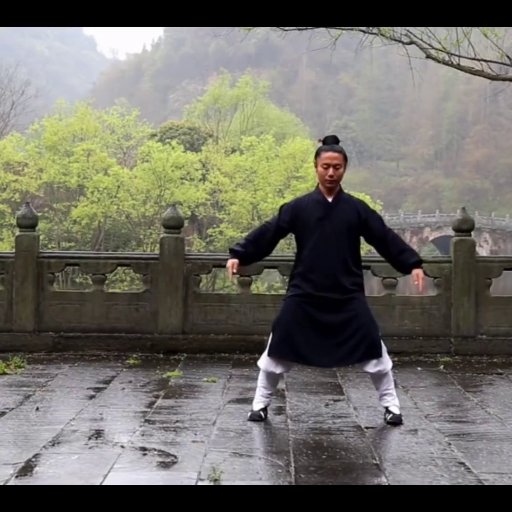} &
    \includegraphics[width=0.105\linewidth]{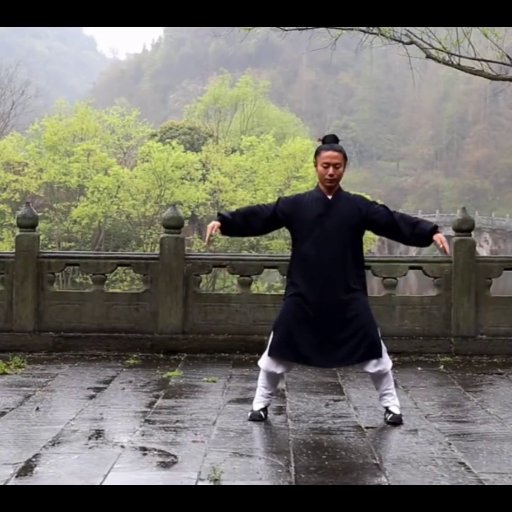} &
    \includegraphics[width=0.105\linewidth]{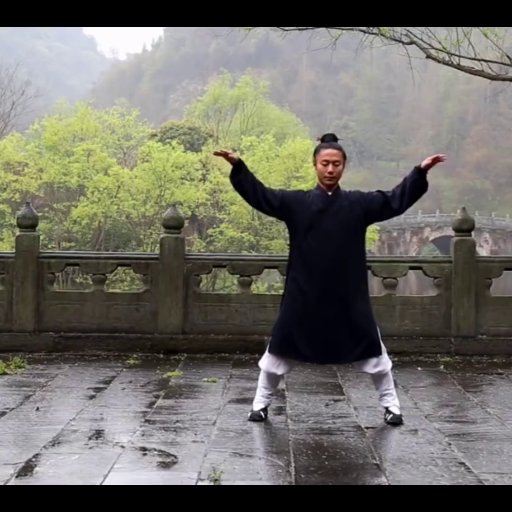} &
    \includegraphics[width=0.105\linewidth]{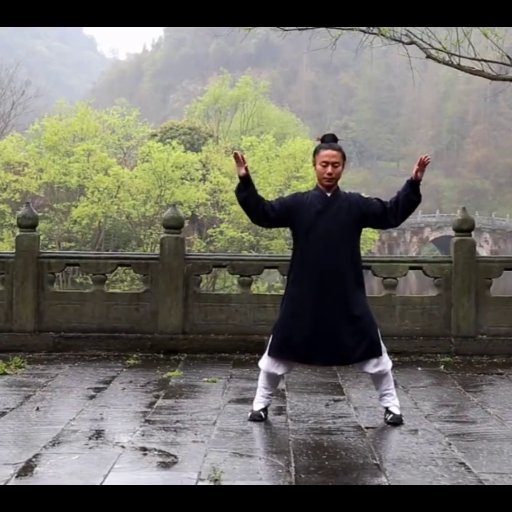} \\ &
    \includegraphics[width=0.105\linewidth]{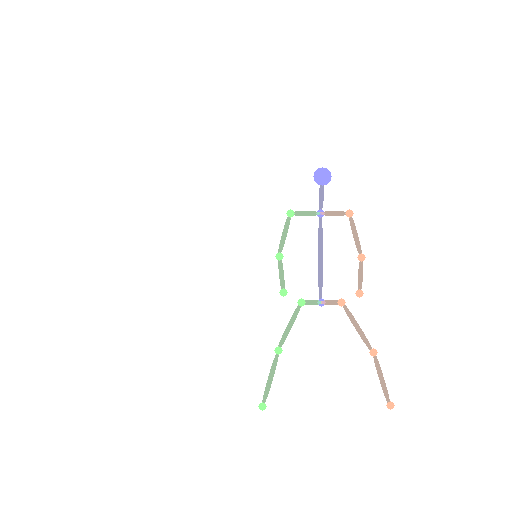} &
    \includegraphics[width=0.105\linewidth]{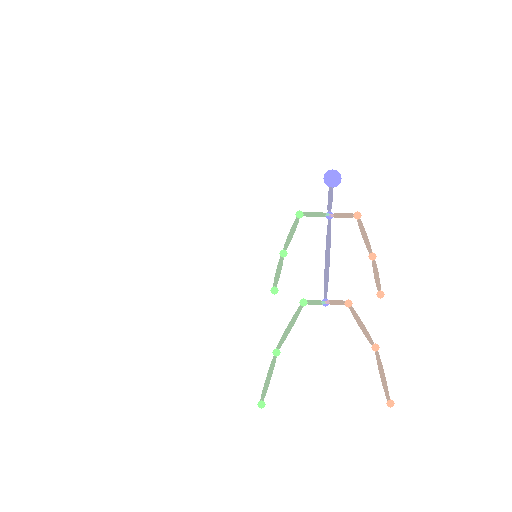} &
    \includegraphics[width=0.105\linewidth]{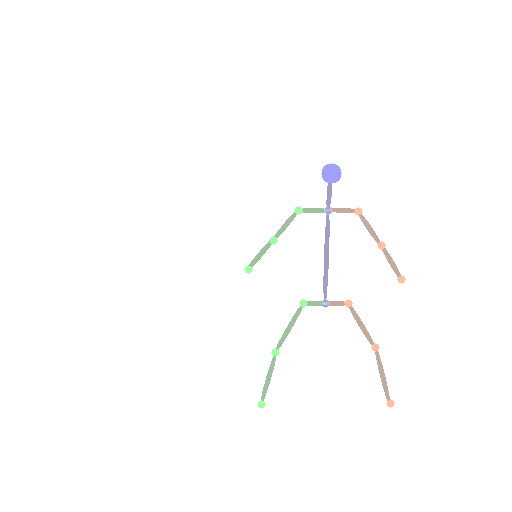} &
    \includegraphics[width=0.105\linewidth]{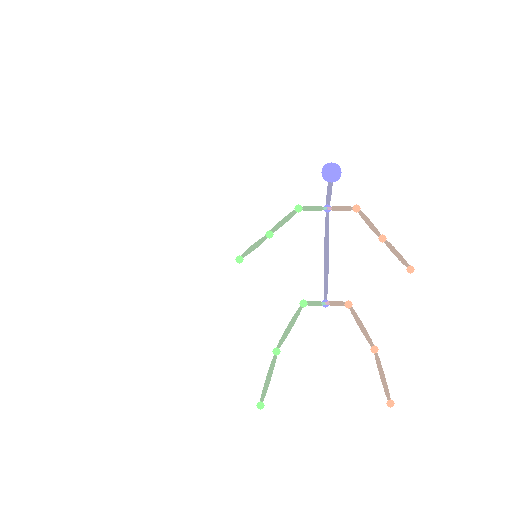} &
    \includegraphics[width=0.105\linewidth]{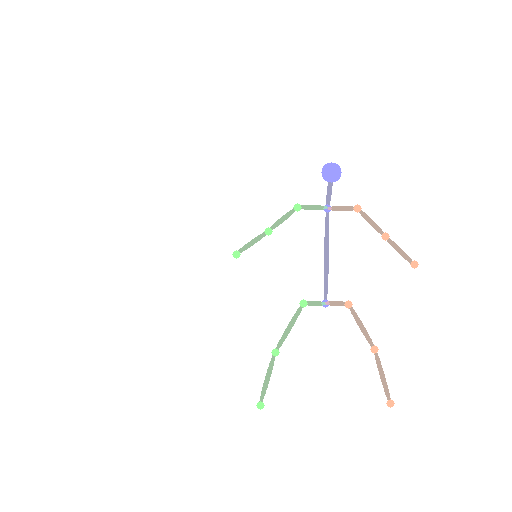} &
    \includegraphics[width=0.105\linewidth]{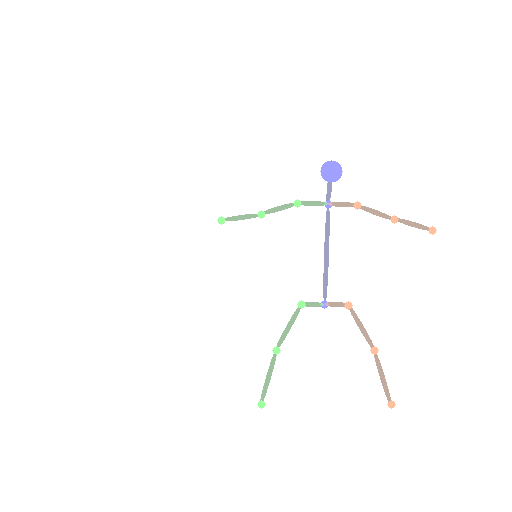} &
    \includegraphics[width=0.105\linewidth]{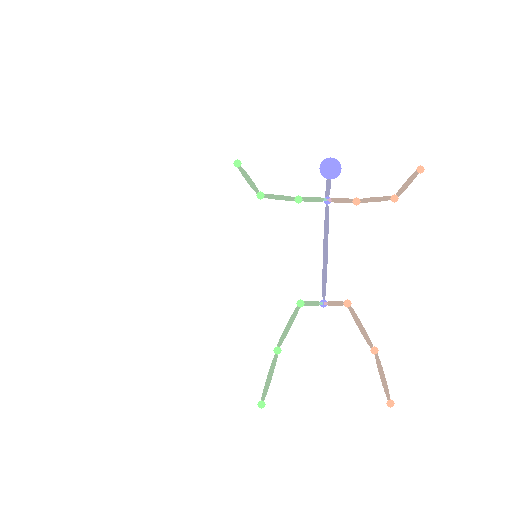} &
    \includegraphics[width=0.105\linewidth]{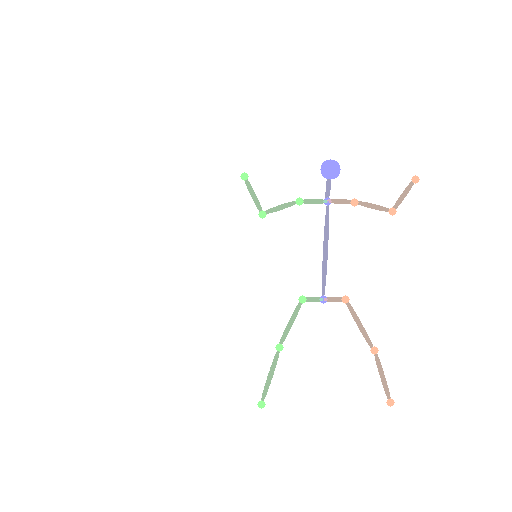} \\ \cline{2-9}
    \includegraphics[width=0.105\linewidth]{figures/application_retargeting/test.png} &
    \includegraphics[width=0.105\linewidth]{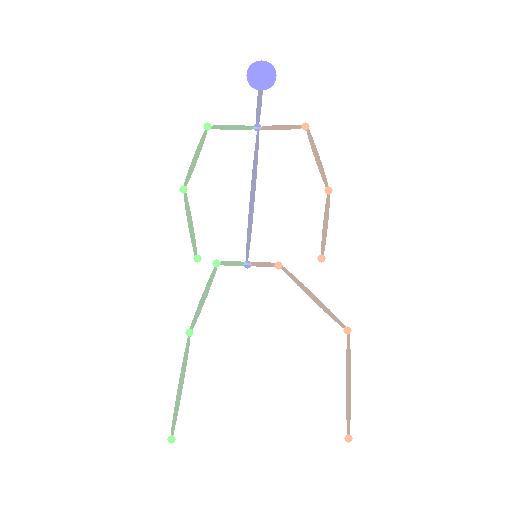} &
    \includegraphics[width=0.105\linewidth]{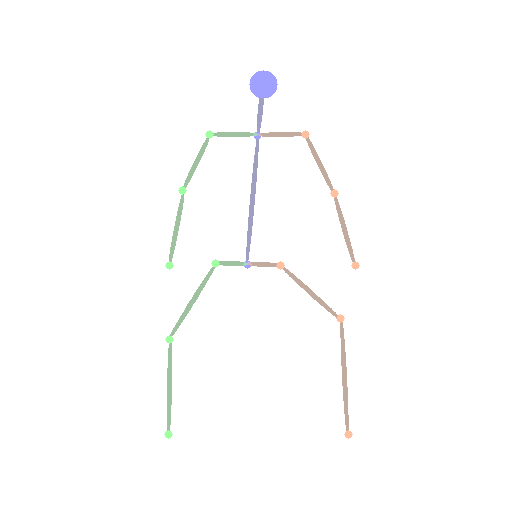} &
    \includegraphics[width=0.105\linewidth]{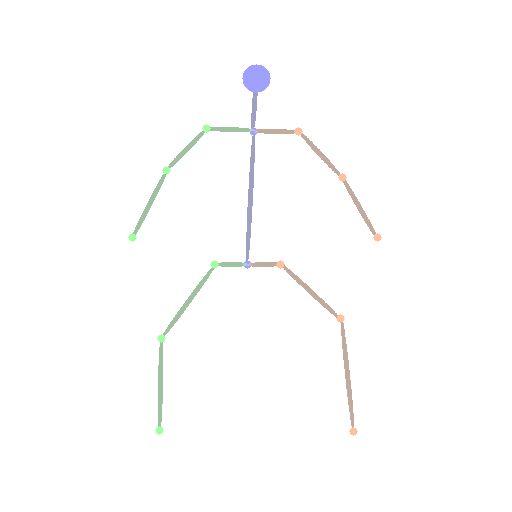} &
    \includegraphics[width=0.105\linewidth]{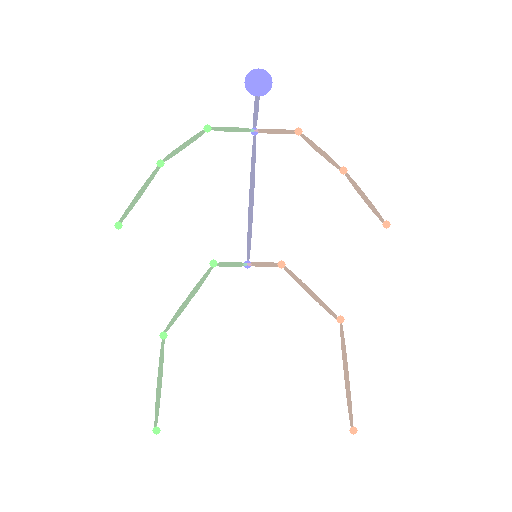} &
    \includegraphics[width=0.105\linewidth]{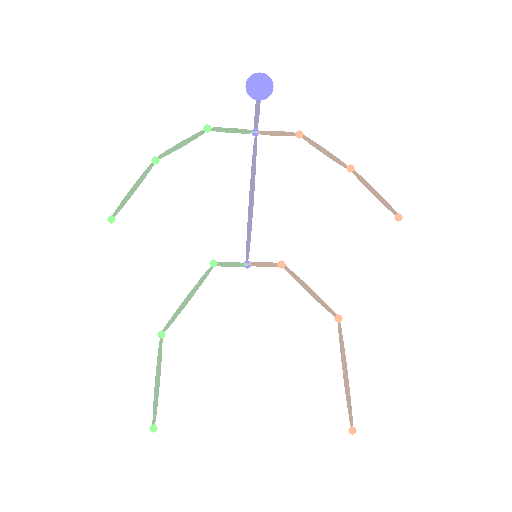} &
    \includegraphics[width=0.105\linewidth]{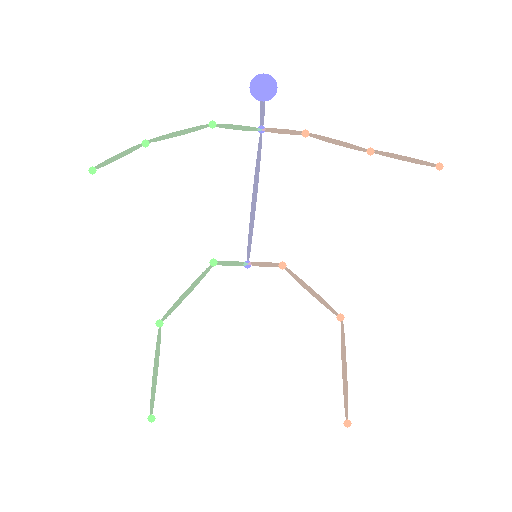} &
    \includegraphics[width=0.105\linewidth]{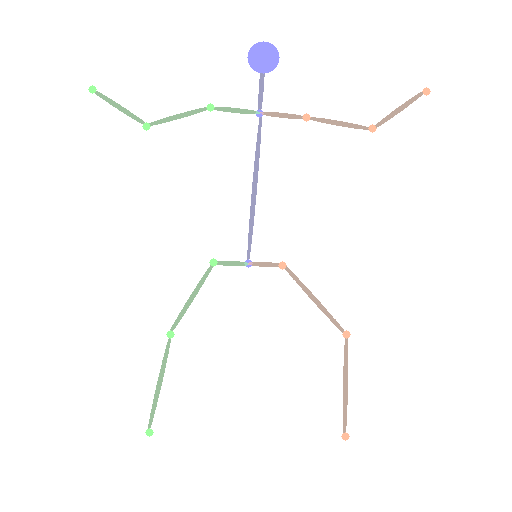} &
    \includegraphics[width=0.105\linewidth]{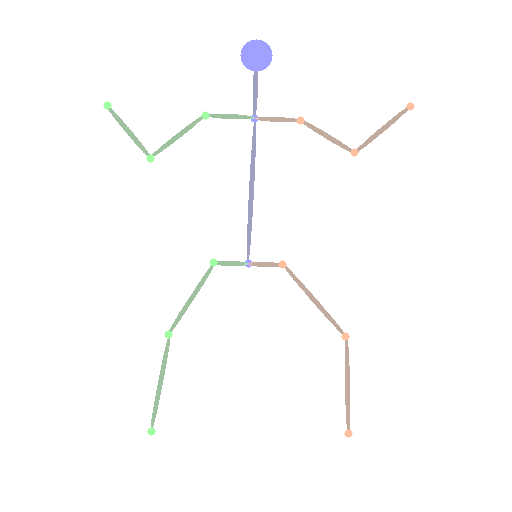} \\ 
    \includegraphics[width=0.105\linewidth]{figures/application_retargeting/input.jpg} &
    \includegraphics[width=0.105\linewidth]{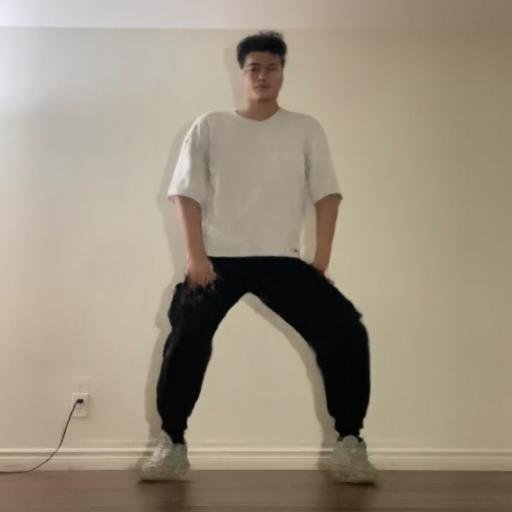} &
    \includegraphics[width=0.105\linewidth]{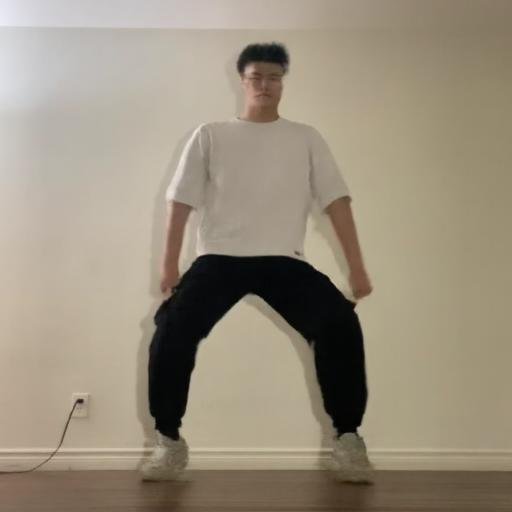} &
    \includegraphics[width=0.105\linewidth]{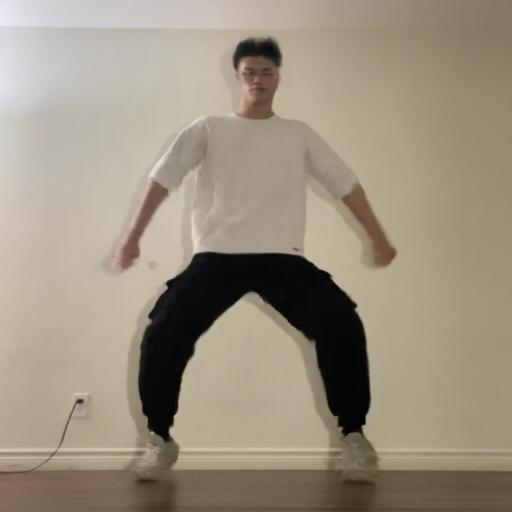} &
    \includegraphics[width=0.105\linewidth]{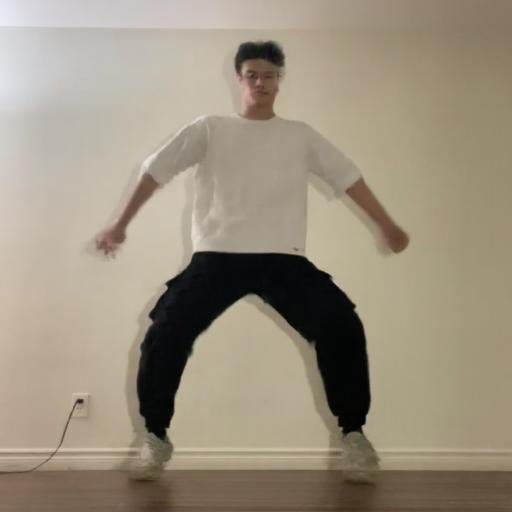} &
    \includegraphics[width=0.105\linewidth]{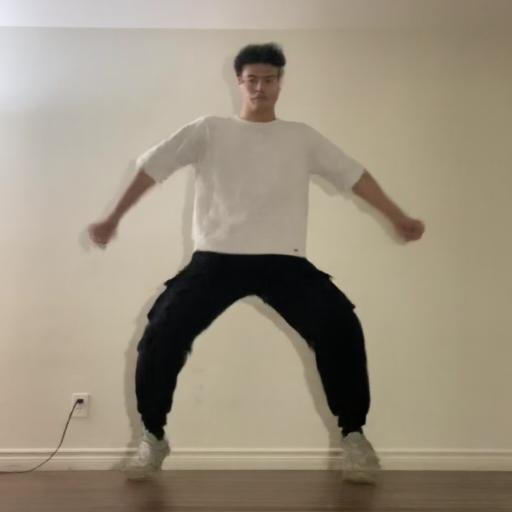} &
    \includegraphics[width=0.105\linewidth]{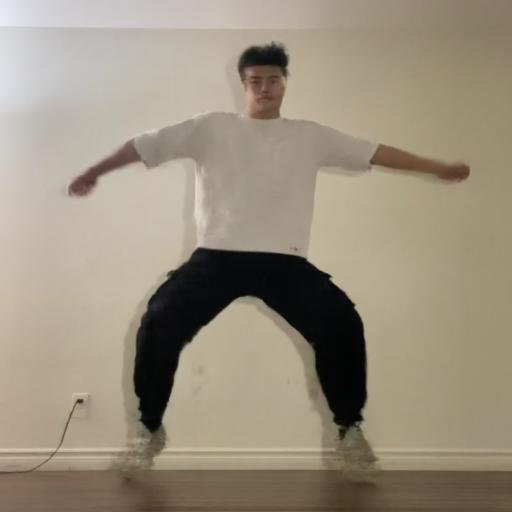} &
    \includegraphics[width=0.105\linewidth]{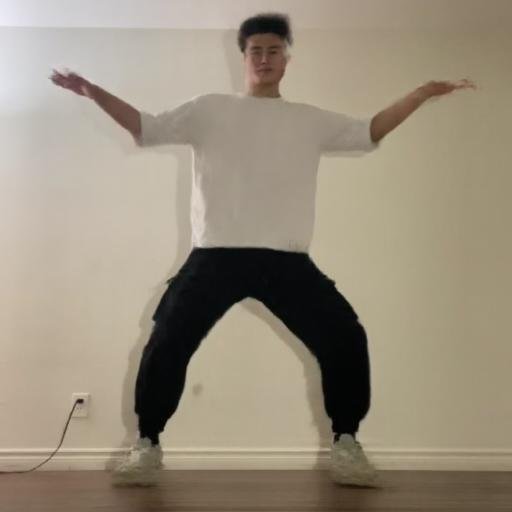} &
    \includegraphics[width=0.105\linewidth]{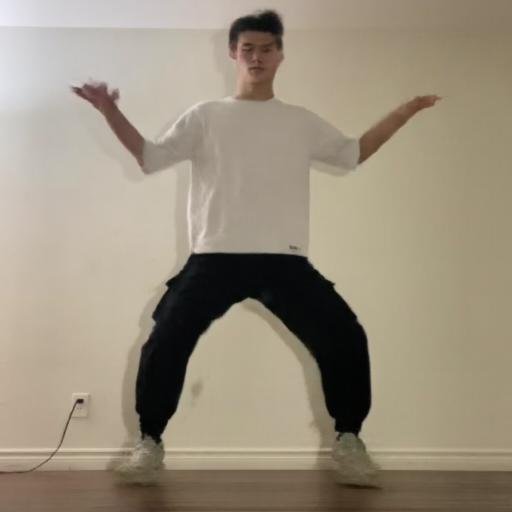} \\
    \multicolumn{9}{c}{(c)} \\ &
    \includegraphics[width=0.105\linewidth]{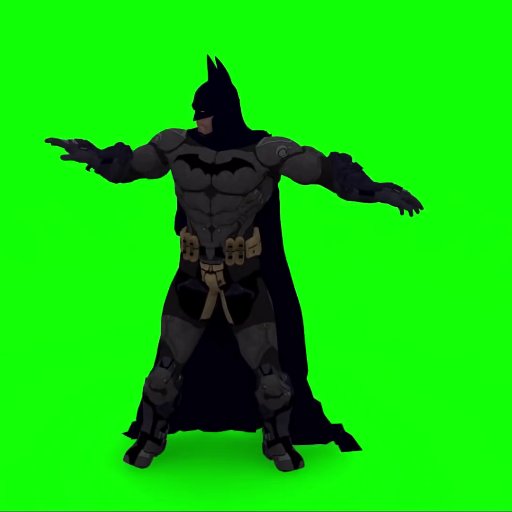} &
    \includegraphics[width=0.105\linewidth]{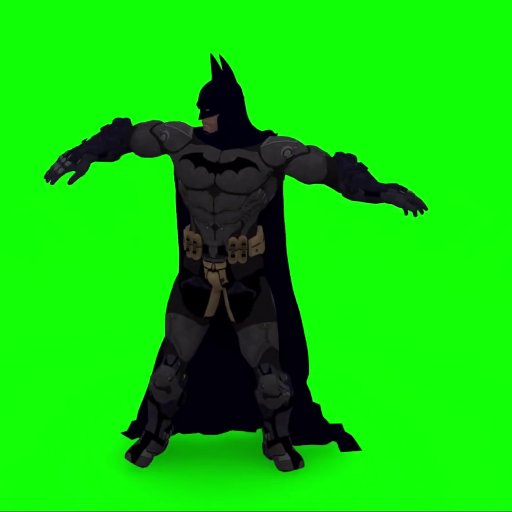} &
    \includegraphics[width=0.105\linewidth]{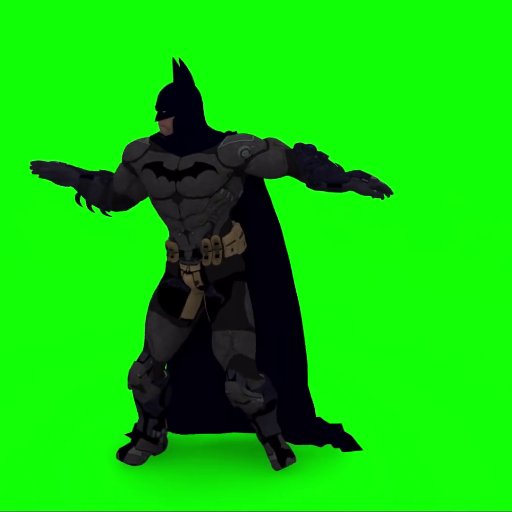} &
    \includegraphics[width=0.105\linewidth]{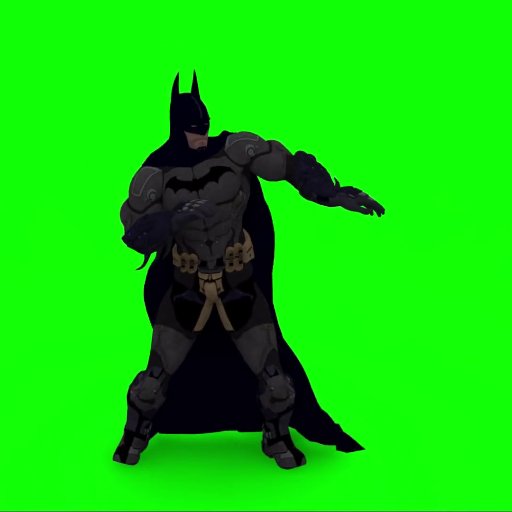} &
    \includegraphics[width=0.105\linewidth]{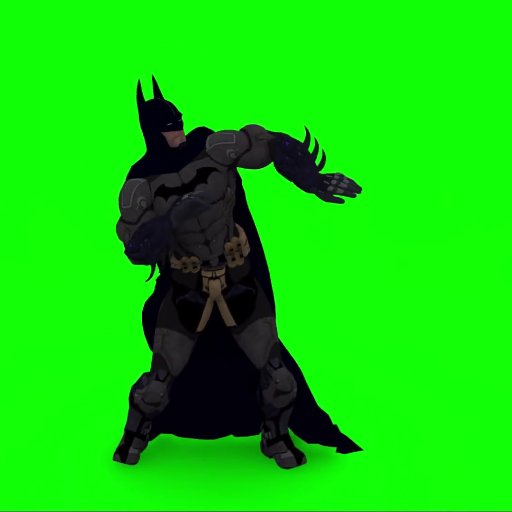} &
    \includegraphics[width=0.105\linewidth]{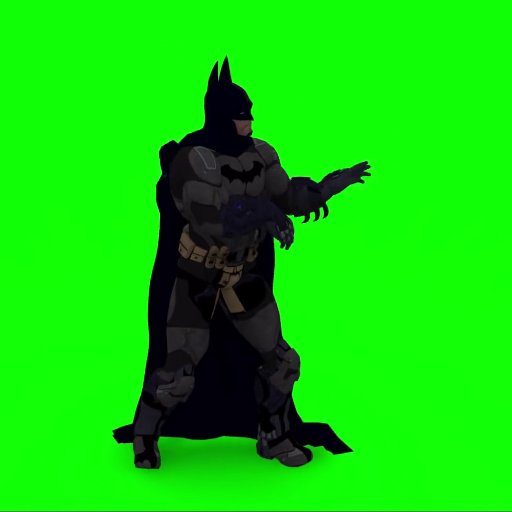} &
    \includegraphics[width=0.105\linewidth]{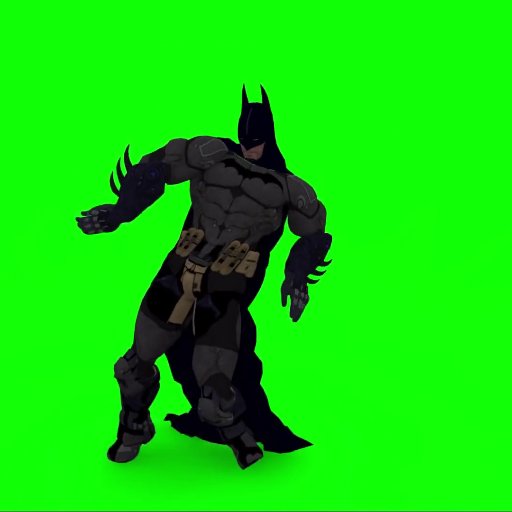} &
    \includegraphics[width=0.105\linewidth]{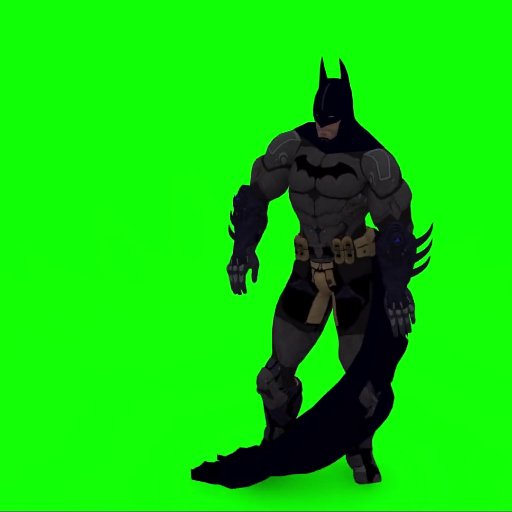} \\ &
    \includegraphics[width=0.105\linewidth]{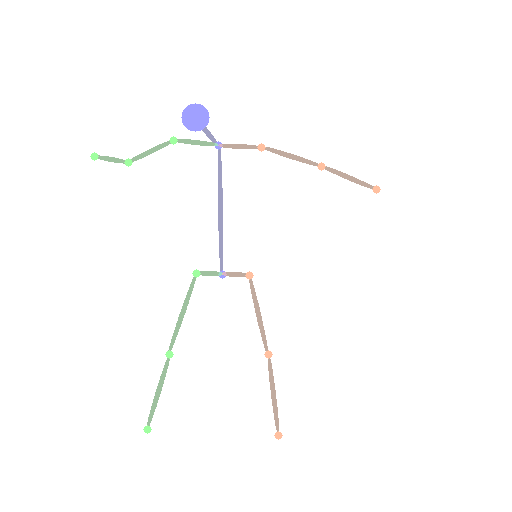} &
    \includegraphics[width=0.105\linewidth]{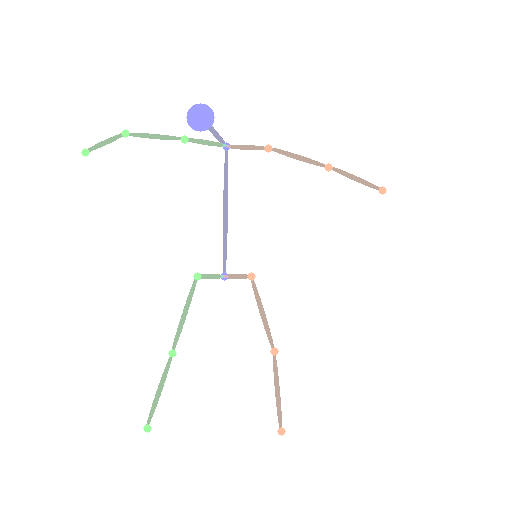} &
    \includegraphics[width=0.105\linewidth]{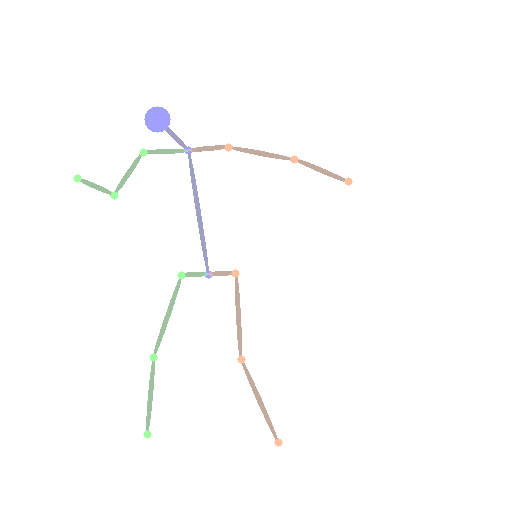} &
    \includegraphics[width=0.105\linewidth]{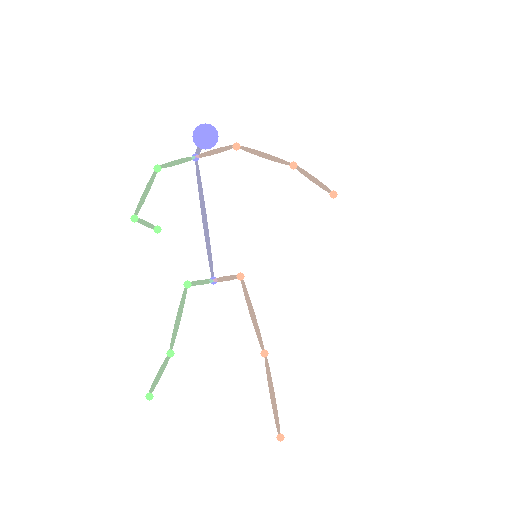} &
    \includegraphics[width=0.105\linewidth]{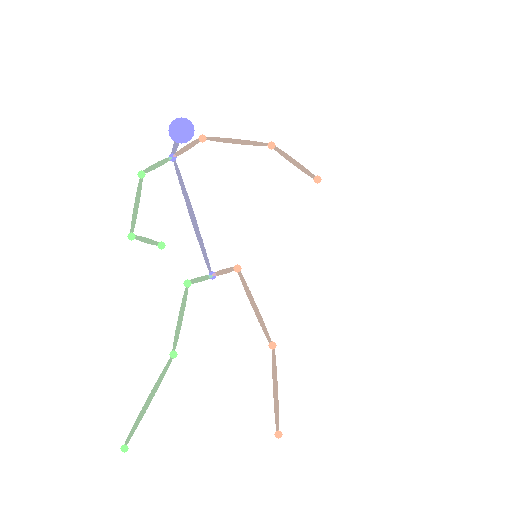} &
    \includegraphics[width=0.105\linewidth]{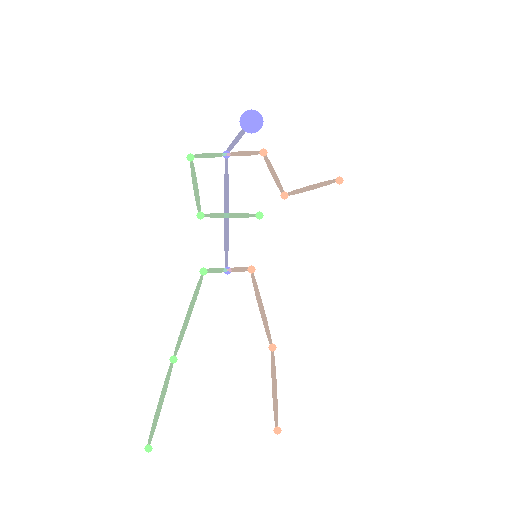} &
    \includegraphics[width=0.105\linewidth]{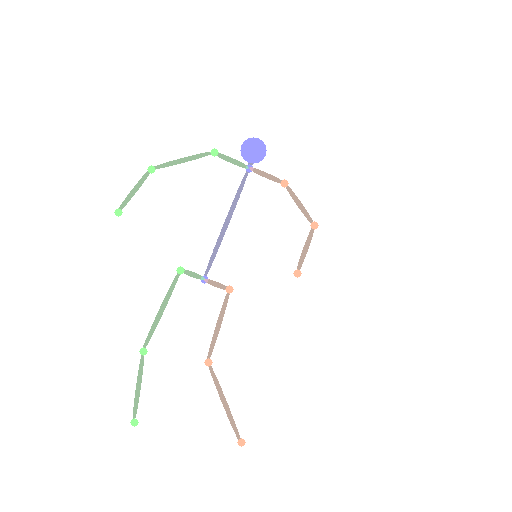} &
    \includegraphics[width=0.105\linewidth]{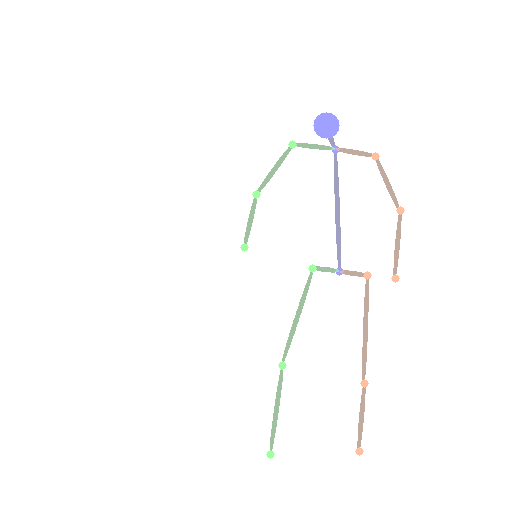} \\ \cline{2-9}
    \includegraphics[width=0.105\linewidth]{figures/application_retargeting/test.png} &
    \includegraphics[width=0.105\linewidth]{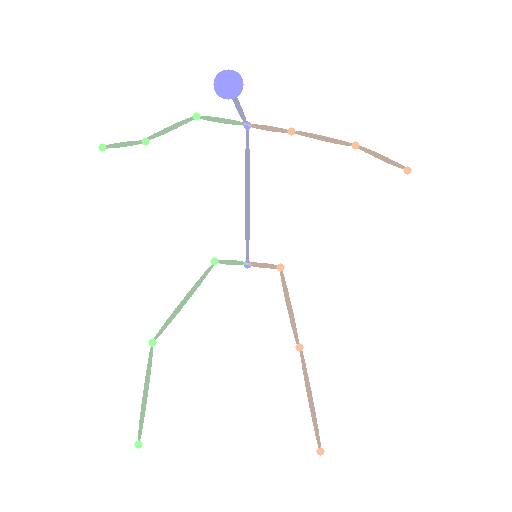} &
    \includegraphics[width=0.105\linewidth]{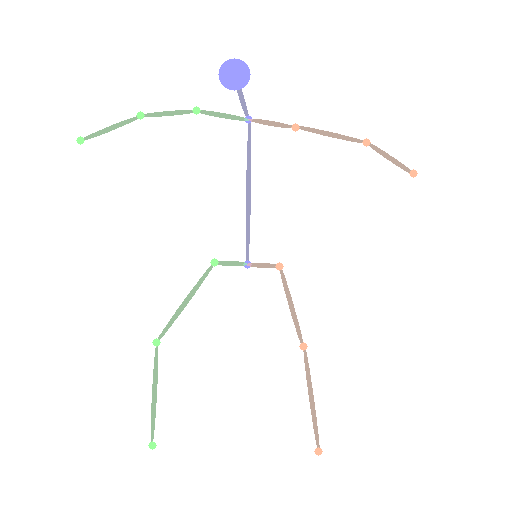} &
    \includegraphics[width=0.105\linewidth]{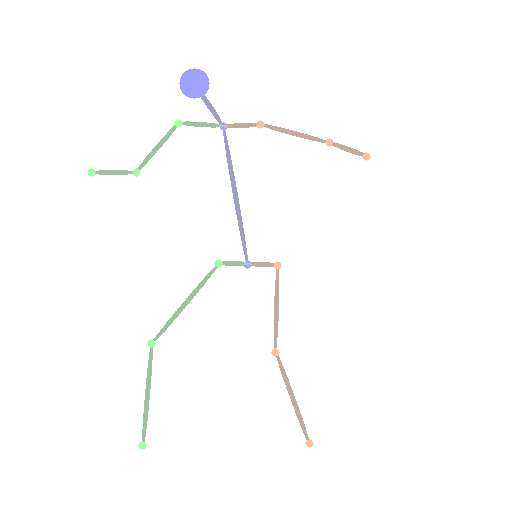} &
    \includegraphics[width=0.105\linewidth]{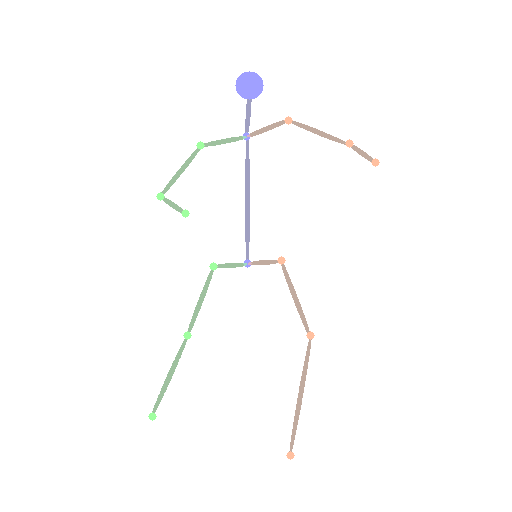} &
    \includegraphics[width=0.105\linewidth]{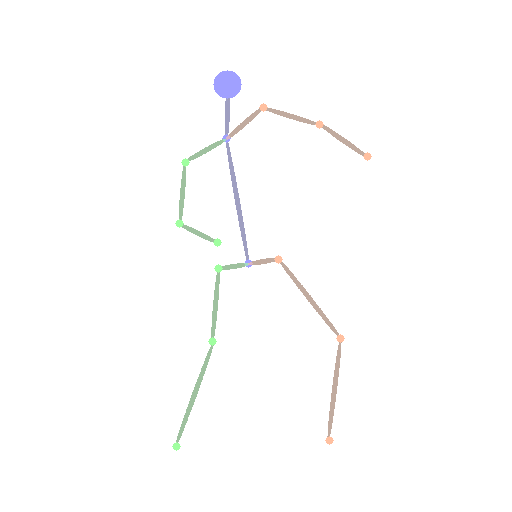} &
    \includegraphics[width=0.105\linewidth]{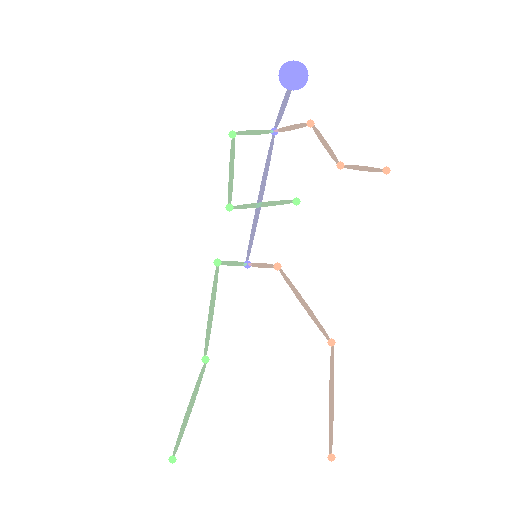} &
    \includegraphics[width=0.105\linewidth]{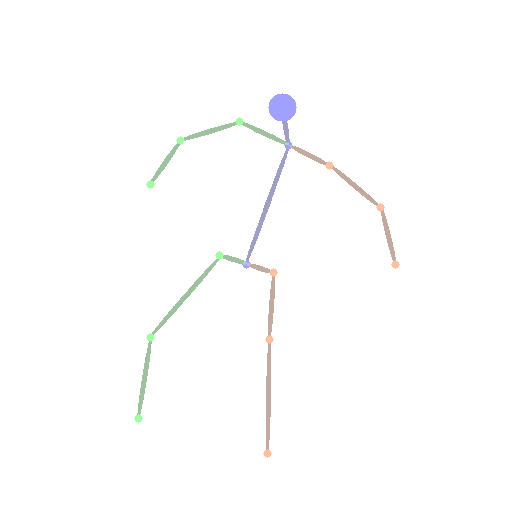} &
    \includegraphics[width=0.105\linewidth]{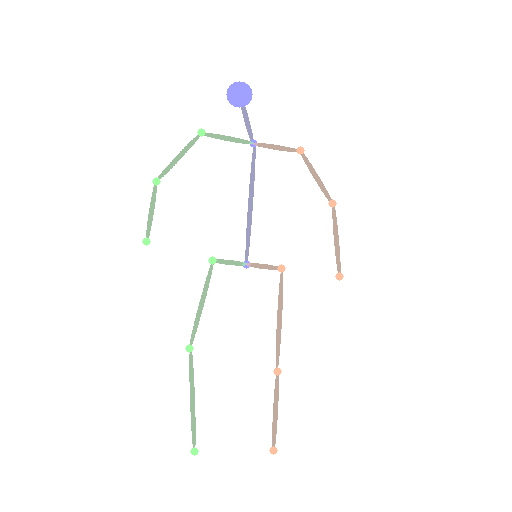} \\
    \includegraphics[width=0.105\linewidth]{figures/application_retargeting/input.jpg} &
    \includegraphics[width=0.105\linewidth]{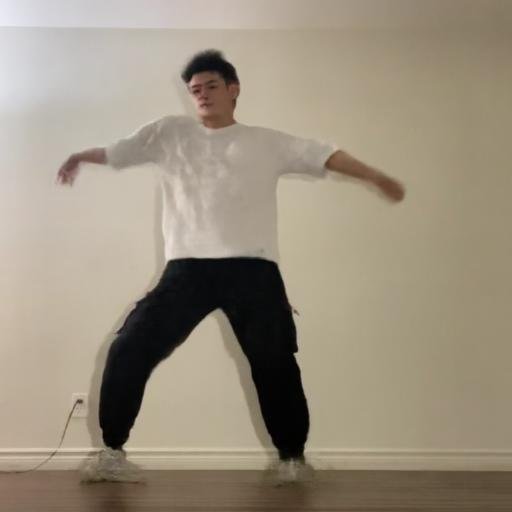} &
    \includegraphics[width=0.105\linewidth]{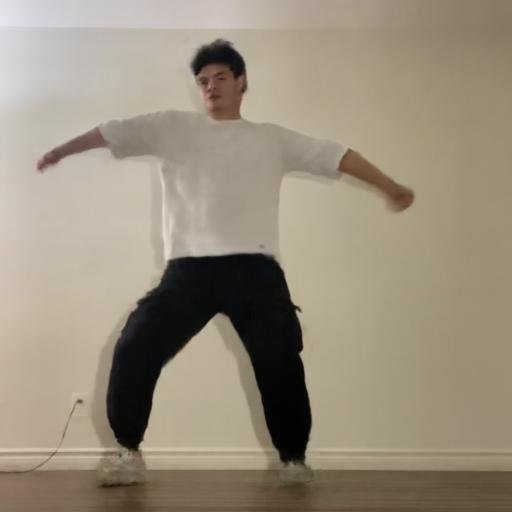} & 
    \includegraphics[width=0.105\linewidth]{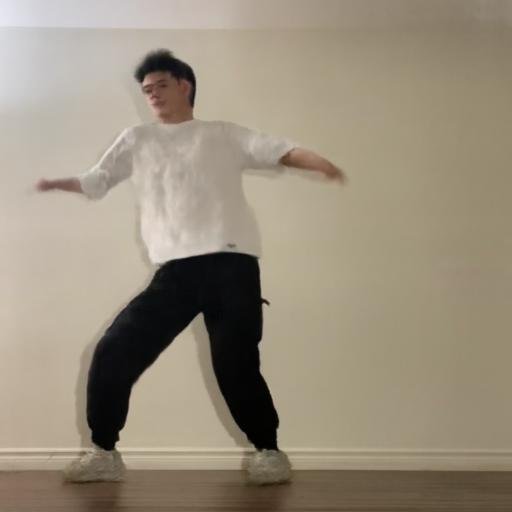} & 
    \includegraphics[width=0.105\linewidth]{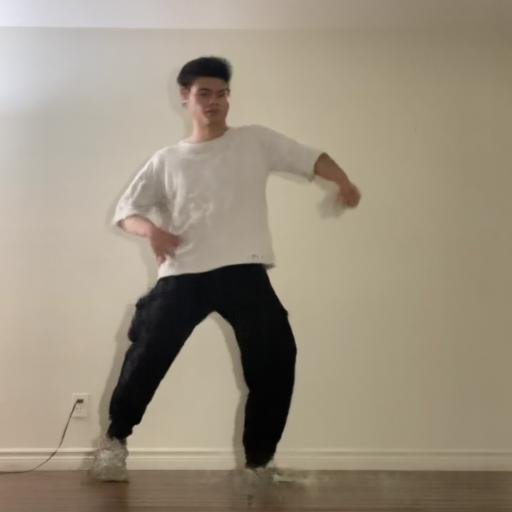} & 
    \includegraphics[width=0.105\linewidth]{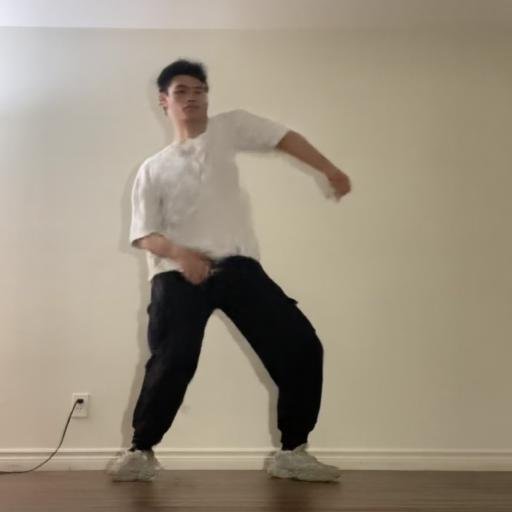} & 
    \includegraphics[width=0.105\linewidth]{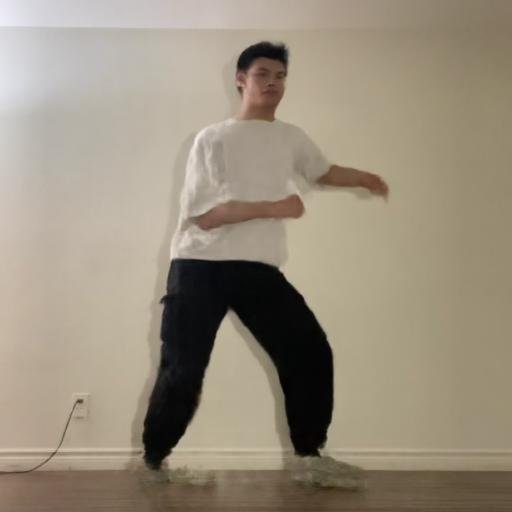} & 
    \includegraphics[width=0.105\linewidth]{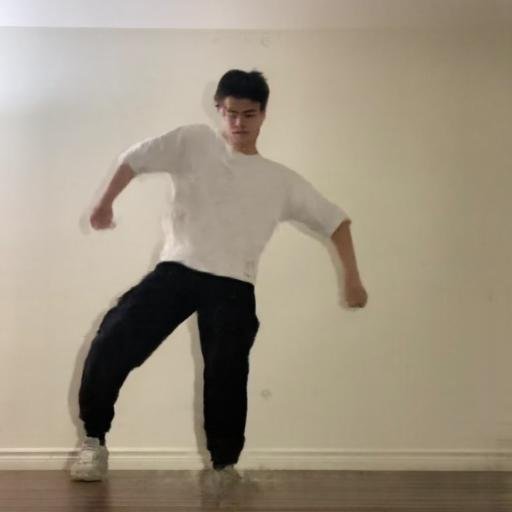} & 
    \includegraphics[width=0.105\linewidth]{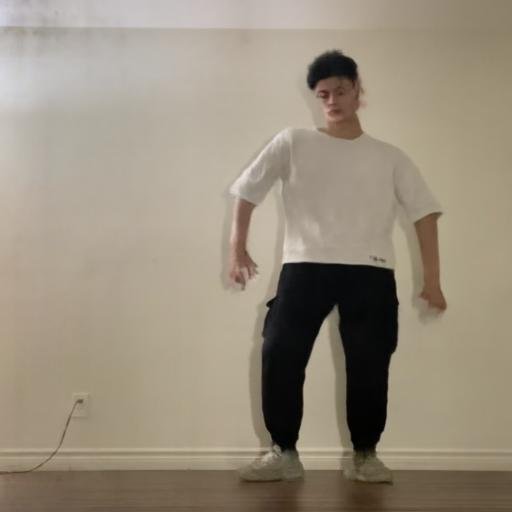} \\
    \multicolumn{9}{c}{(d)} \\
    \end{tabular}
    \caption{Motion retargeting results from different driving subjects to the same target person. In each case, left column shows the target person and his skeleton, right columns (from top to bottom) represent input source frames, extracted source skeletons, transformed skeletons, and generated target frames.}
    \label{fig:motion_retarget2}
\end{figure*}

Figure~\ref{fig:motion_retarget2} shows more results of 2D skeleton-based motion retargeting. Although driven by different subjects, the target skeleton identity is completely preserved in the generated results.
\begin{figure*}
    \centering
    \setlength{\tabcolsep}{0.5mm}
    \begin{tabular}{c|cccccccc} &
    \includegraphics[width=0.105\linewidth]{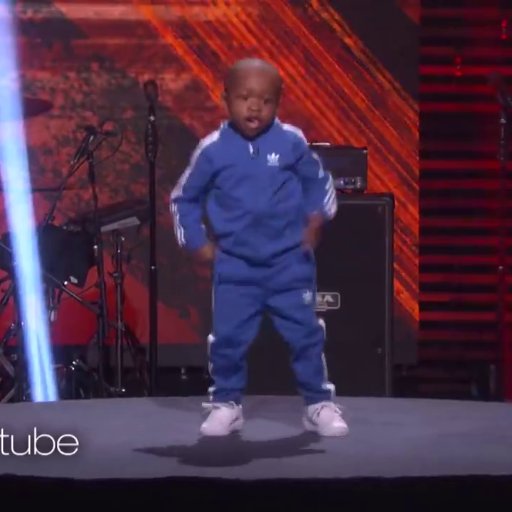} &
    \includegraphics[width=0.105\linewidth]{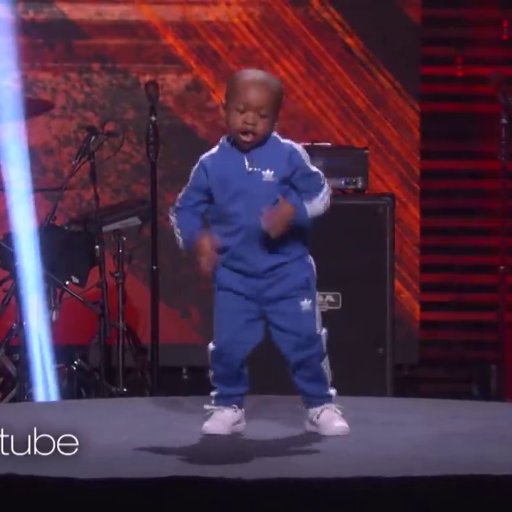} &
    \includegraphics[width=0.105\linewidth]{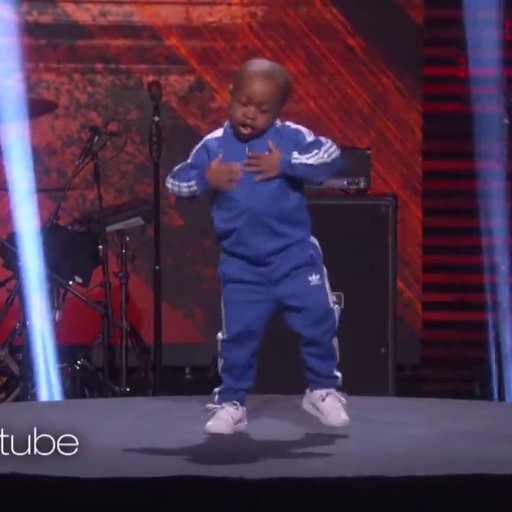} &
    \includegraphics[width=0.105\linewidth]{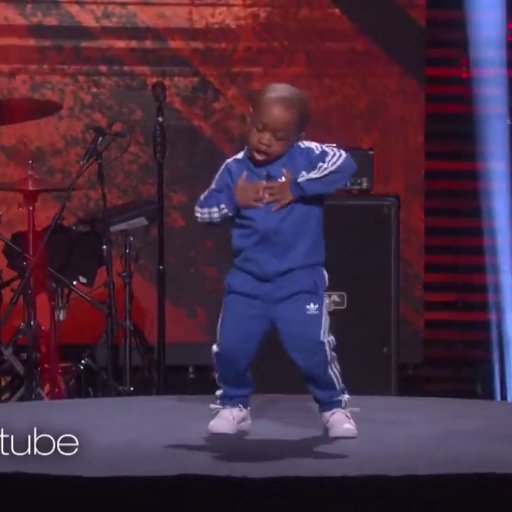} &
    \includegraphics[width=0.105\linewidth]{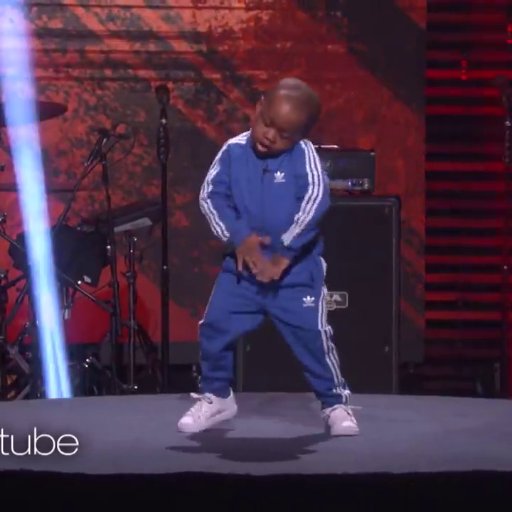} &
    \includegraphics[width=0.105\linewidth]{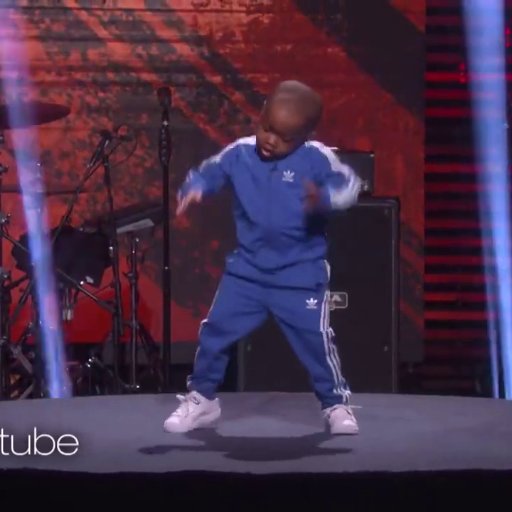} &
    \includegraphics[width=0.105\linewidth]{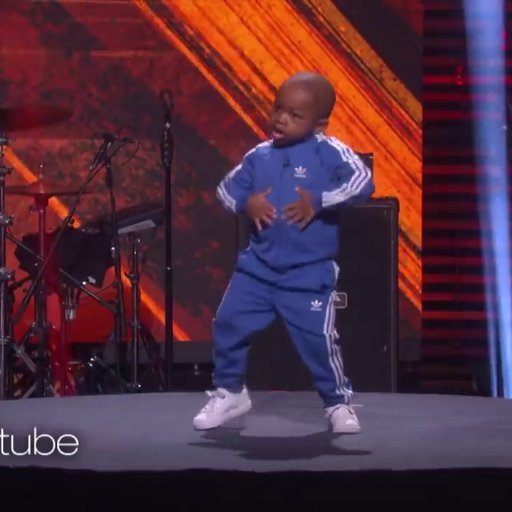} &
    \includegraphics[width=0.105\linewidth]{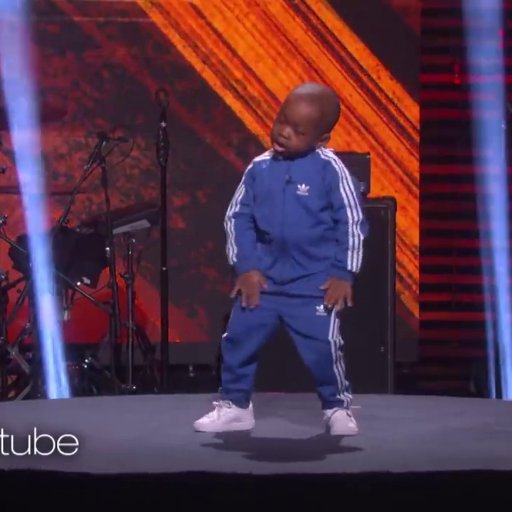} \\ &
    \includegraphics[width=0.105\linewidth]{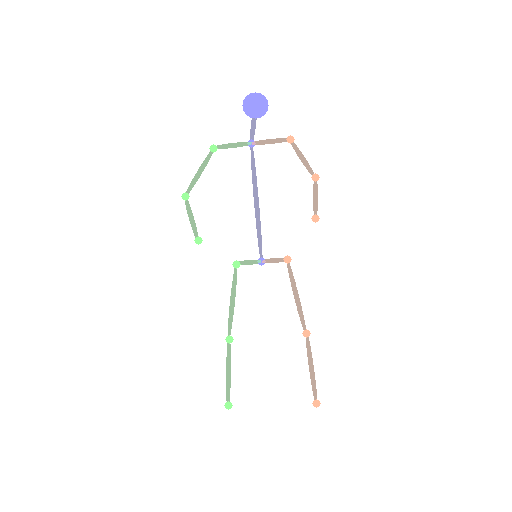} &
    \includegraphics[width=0.105\linewidth]{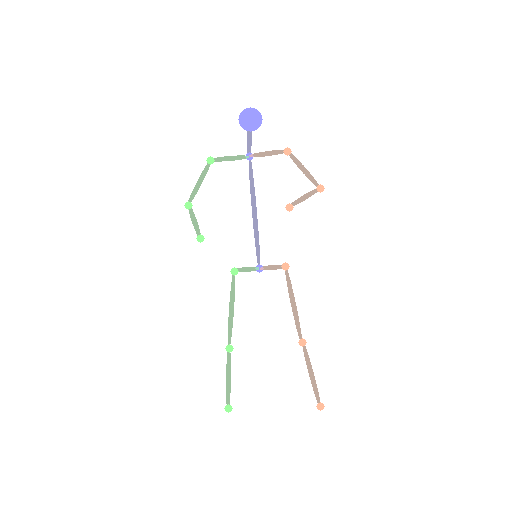} &
    \includegraphics[width=0.105\linewidth]{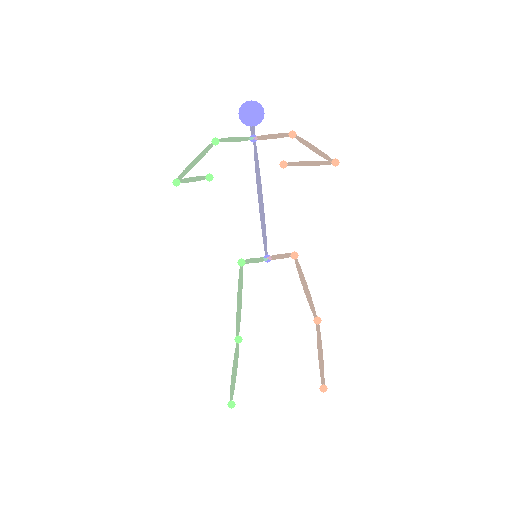} &
    \includegraphics[width=0.105\linewidth]{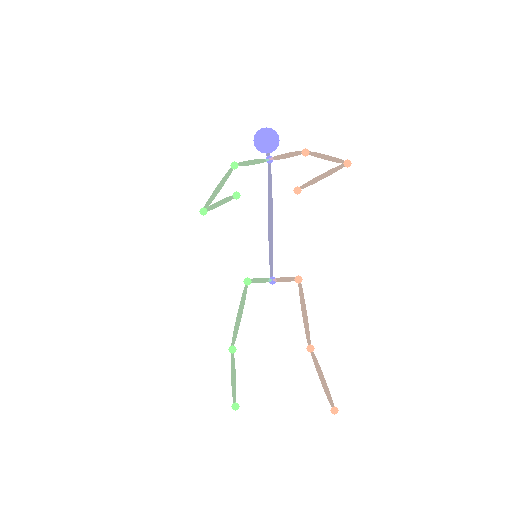} &
    \includegraphics[width=0.105\linewidth]{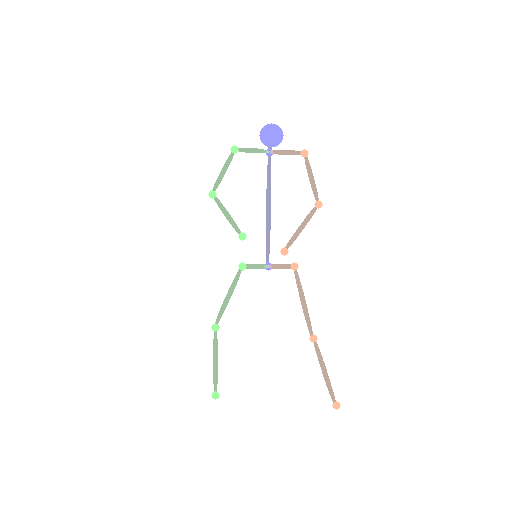} &
    \includegraphics[width=0.105\linewidth]{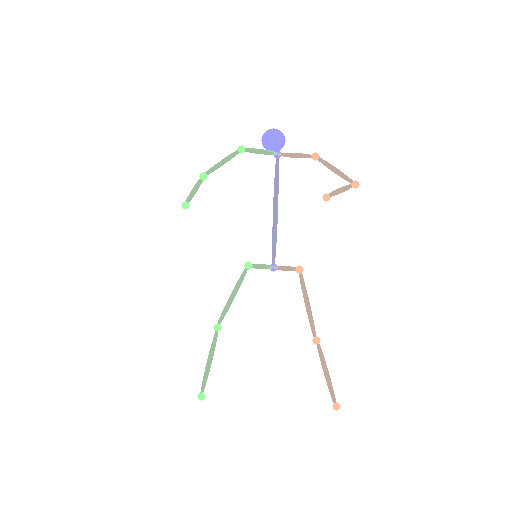} & 
    \includegraphics[width=0.105\linewidth]{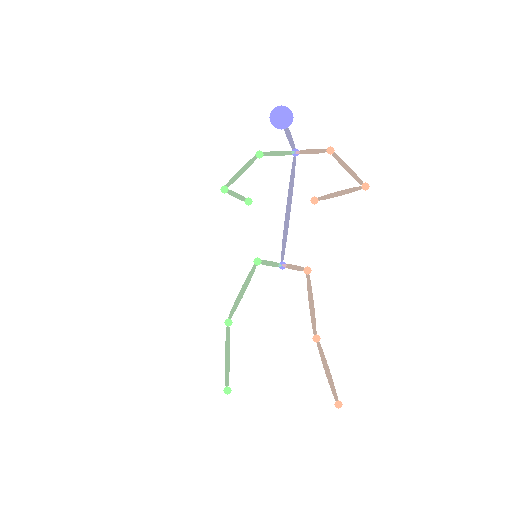} & 
    \includegraphics[width=0.105\linewidth]{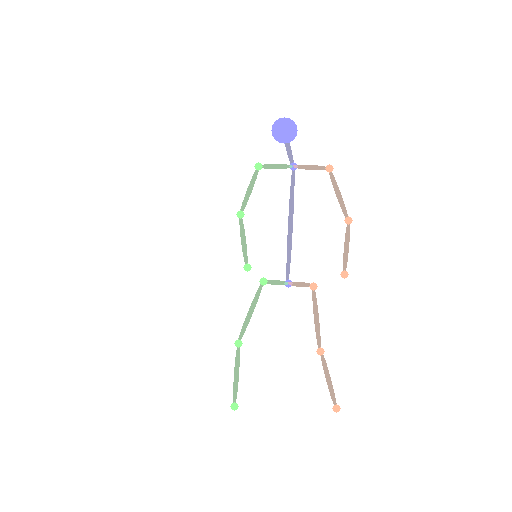} \\ \cline{2-9}
    \includegraphics[width=0.105\linewidth]{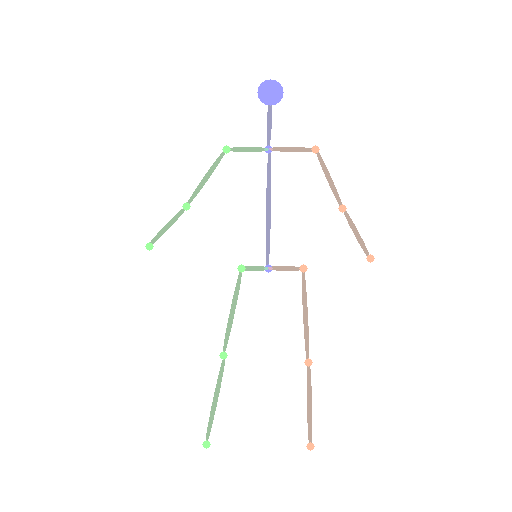} &
    \includegraphics[width=0.105\linewidth]{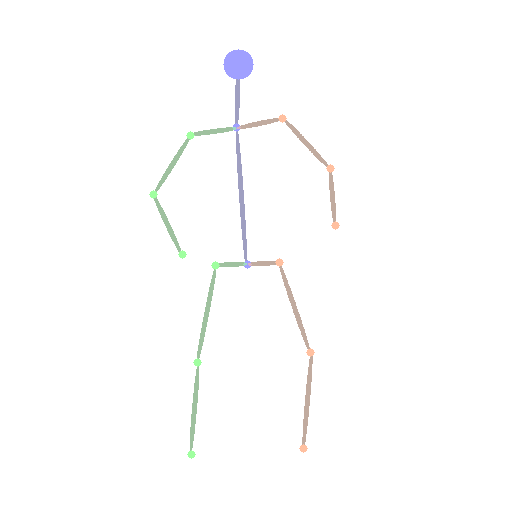} &
    \includegraphics[width=0.105\linewidth]{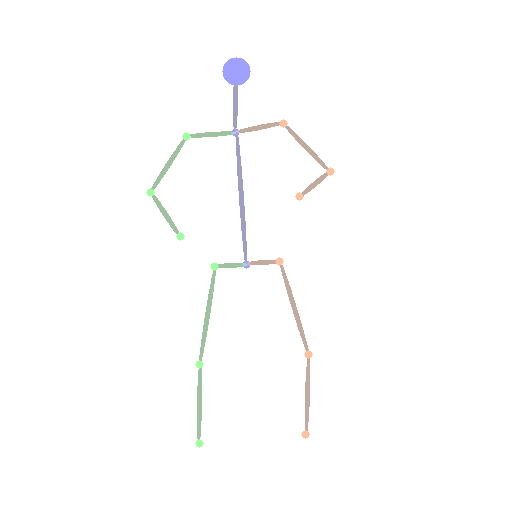} &
    \includegraphics[width=0.105\linewidth]{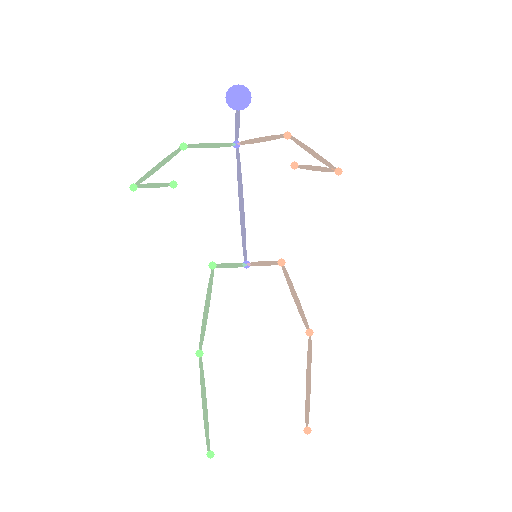} &
    \includegraphics[width=0.105\linewidth]{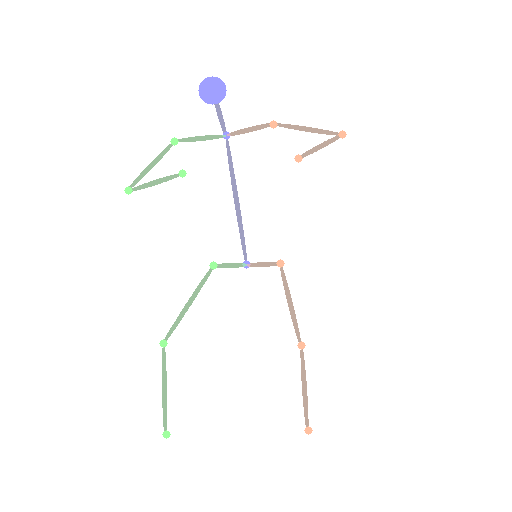} &
    \includegraphics[width=0.105\linewidth]{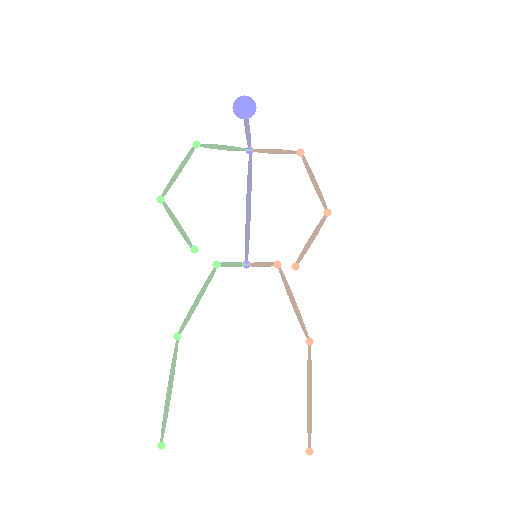} &
    \includegraphics[width=0.105\linewidth]{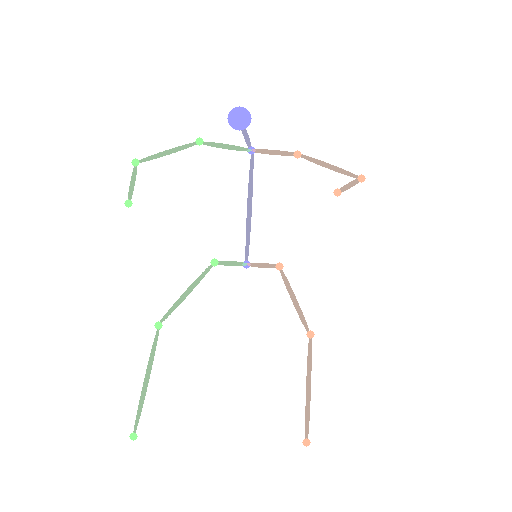} &
    \includegraphics[width=0.105\linewidth]{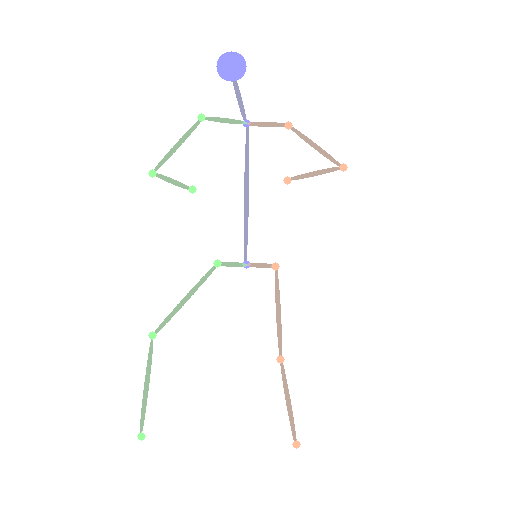} &
    \includegraphics[width=0.105\linewidth]{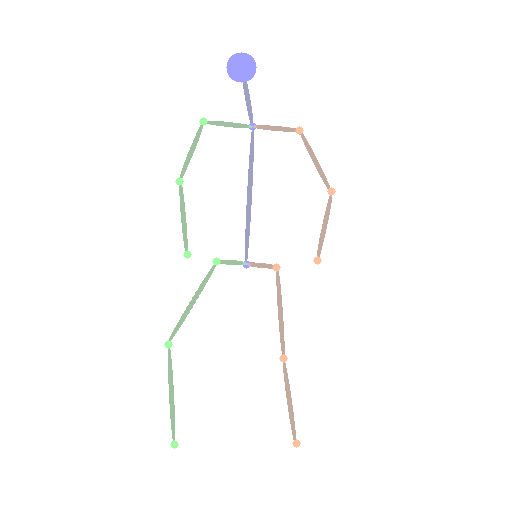} \\
    \includegraphics[width=0.105\linewidth]{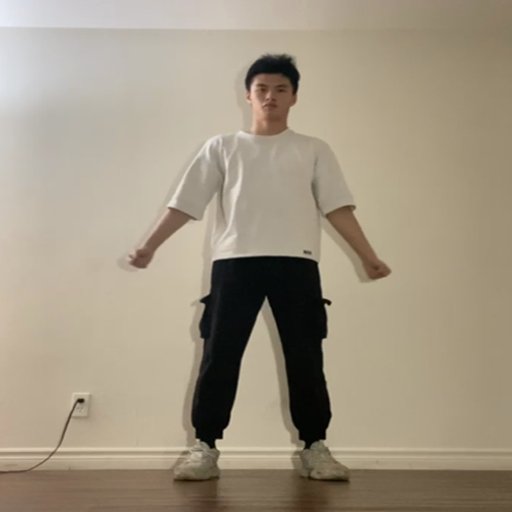} &
    \includegraphics[width=0.105\linewidth]{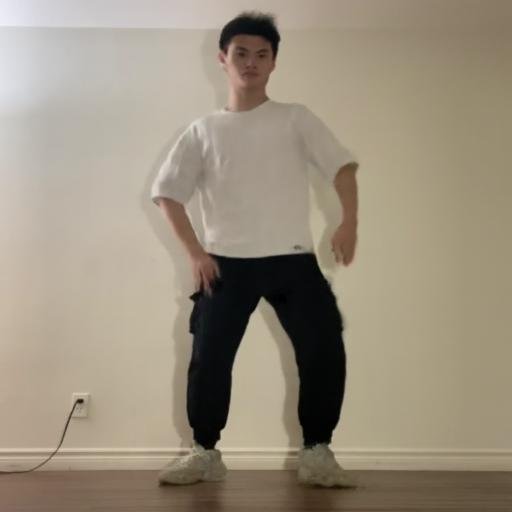} &
    \includegraphics[width=0.105\linewidth]{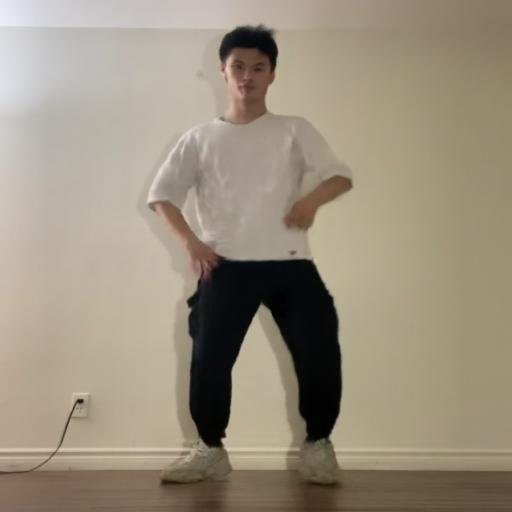} &
    \includegraphics[width=0.105\linewidth]{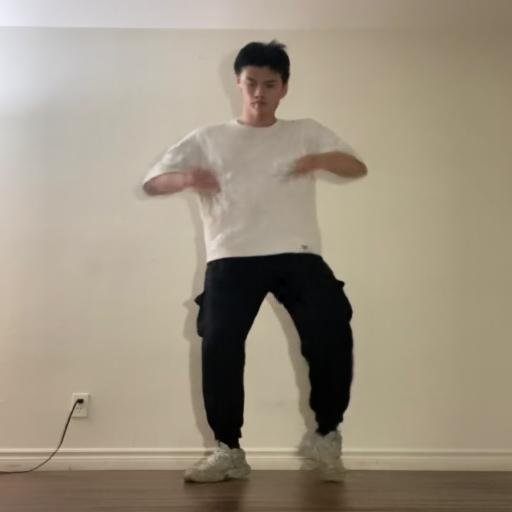} &
    \includegraphics[width=0.105\linewidth]{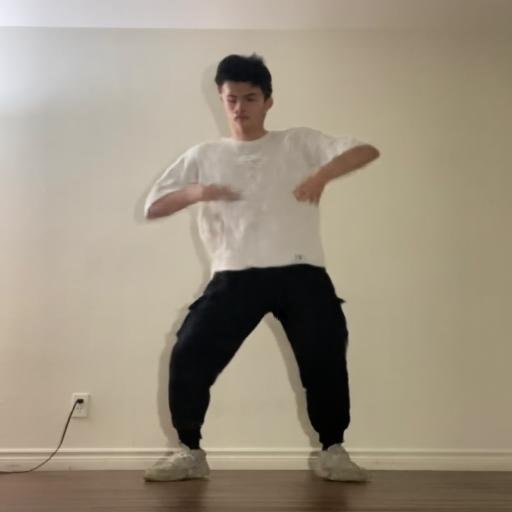} &
    \includegraphics[width=0.105\linewidth]{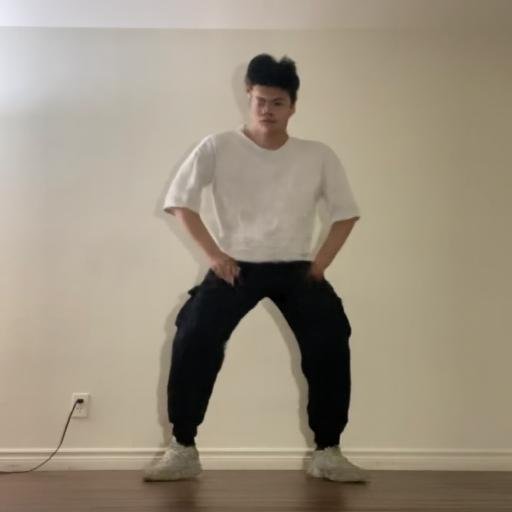} &
    \includegraphics[width=0.105\linewidth]{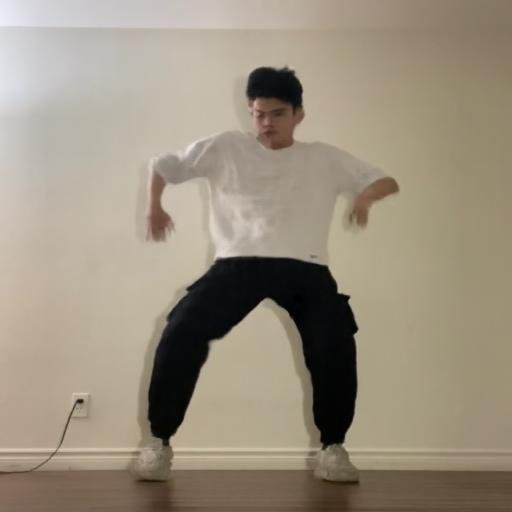} &
    \includegraphics[width=0.105\linewidth]{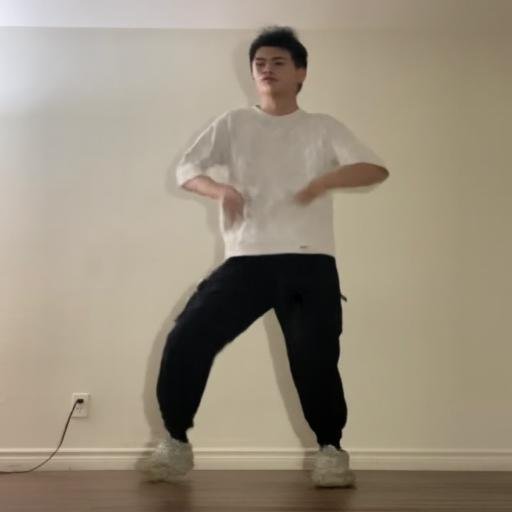} &
    \includegraphics[width=0.105\linewidth]{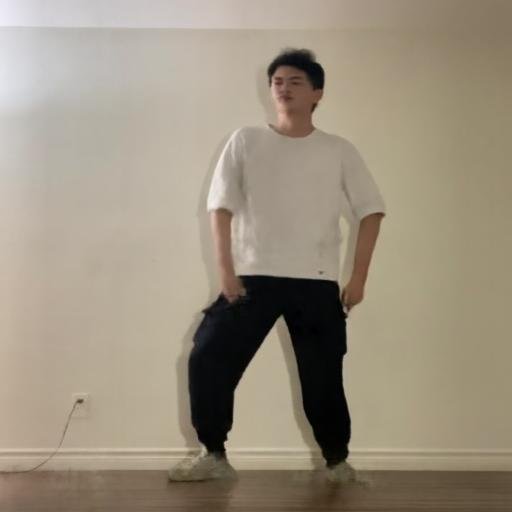} \\
    \multicolumn{9}{c}{(a)} \\ &
    \includegraphics[width=0.105\linewidth]{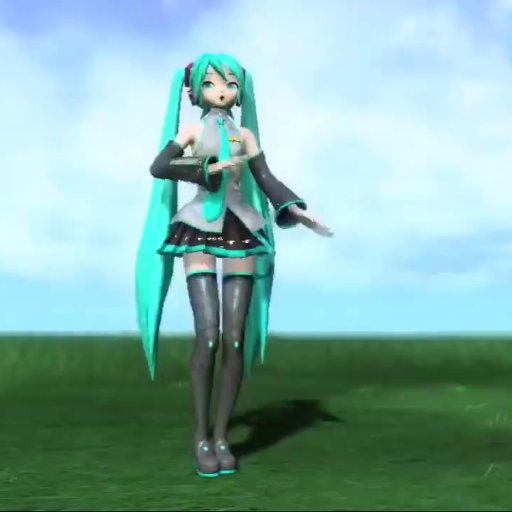} &
    \includegraphics[width=0.105\linewidth]{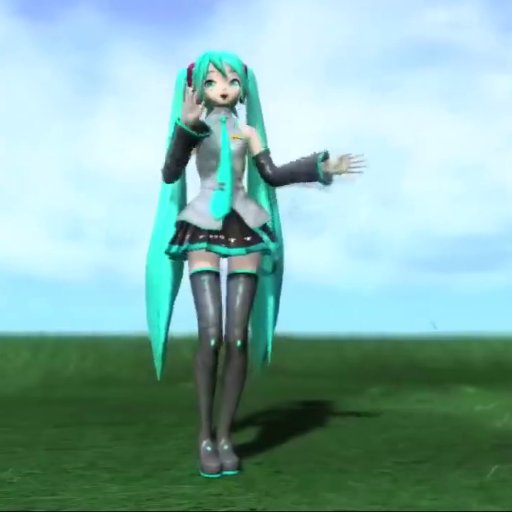} &
    \includegraphics[width=0.105\linewidth]{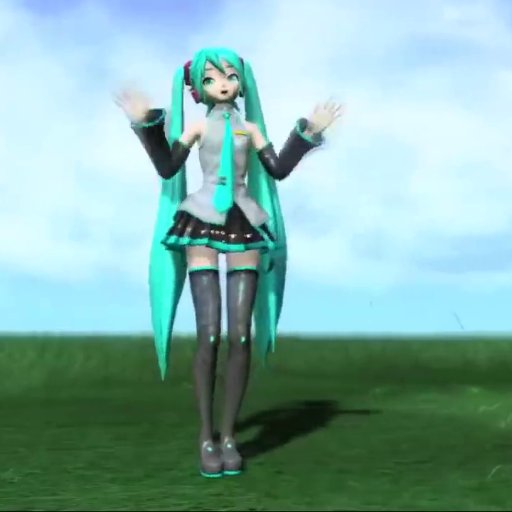} &
    \includegraphics[width=0.105\linewidth]{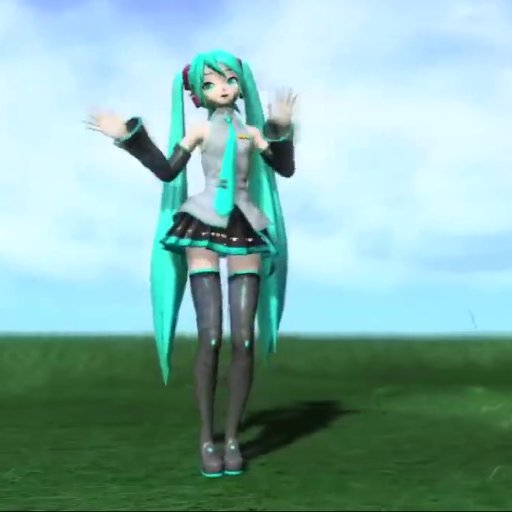} &
    \includegraphics[width=0.105\linewidth]{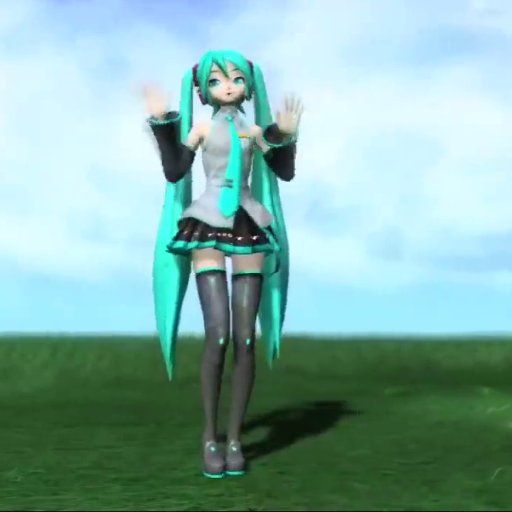} &
    \includegraphics[width=0.105\linewidth]{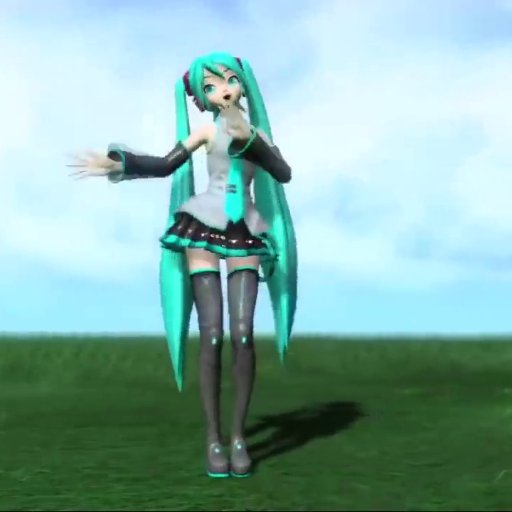} &
    \includegraphics[width=0.105\linewidth]{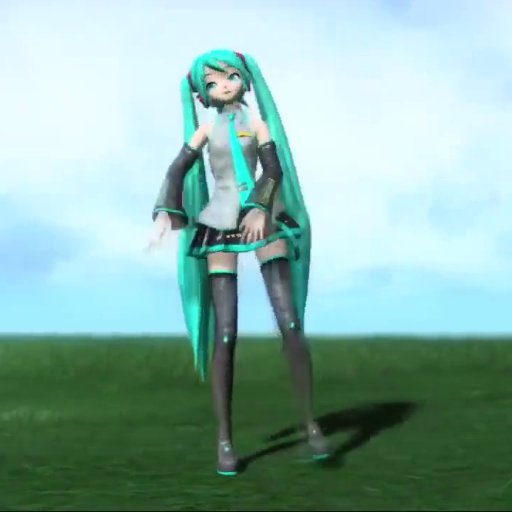} &
    \includegraphics[width=0.105\linewidth]{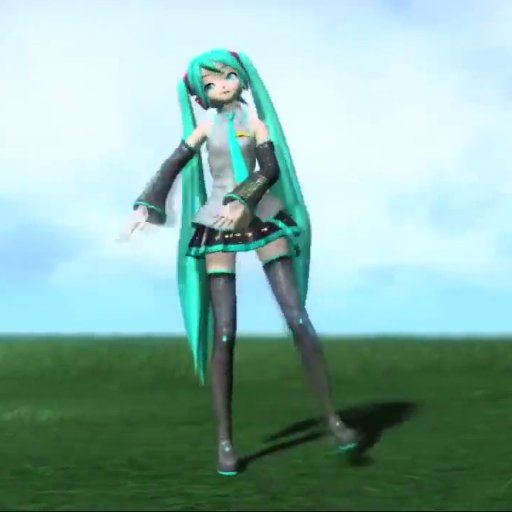} \\ &
    \includegraphics[width=0.105\linewidth]{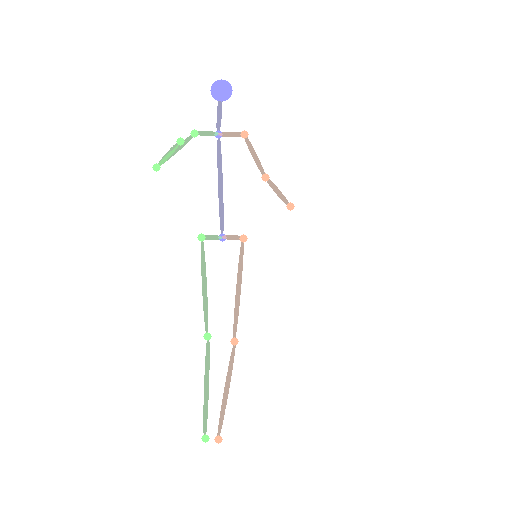} &
    \includegraphics[width=0.105\linewidth]{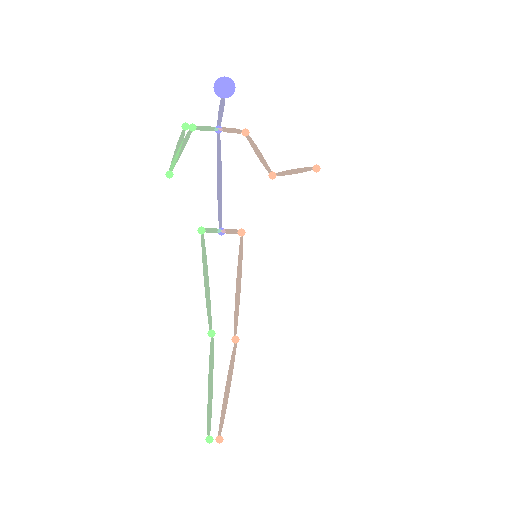} &
    \includegraphics[width=0.105\linewidth]{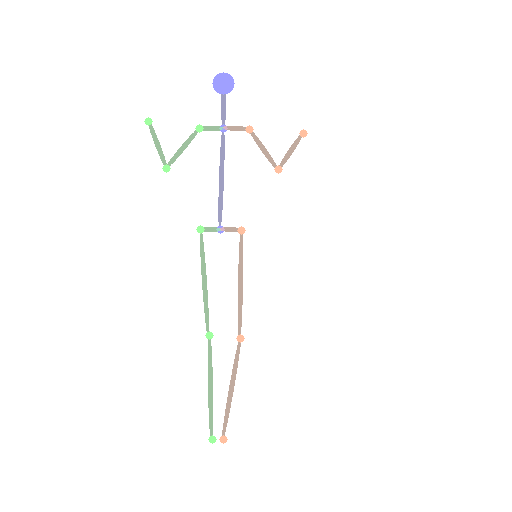} &
    \includegraphics[width=0.105\linewidth]{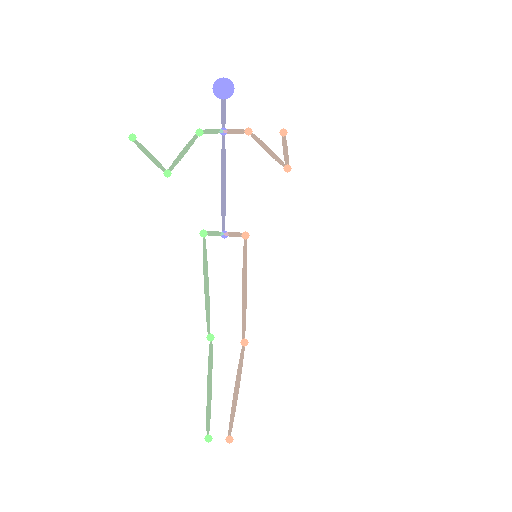} &
    \includegraphics[width=0.105\linewidth]{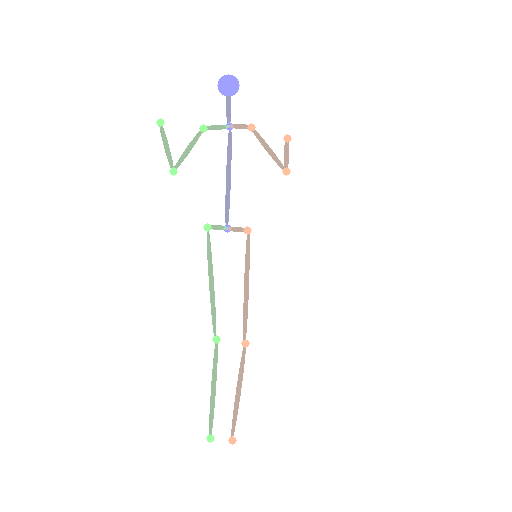} &
    \includegraphics[width=0.105\linewidth]{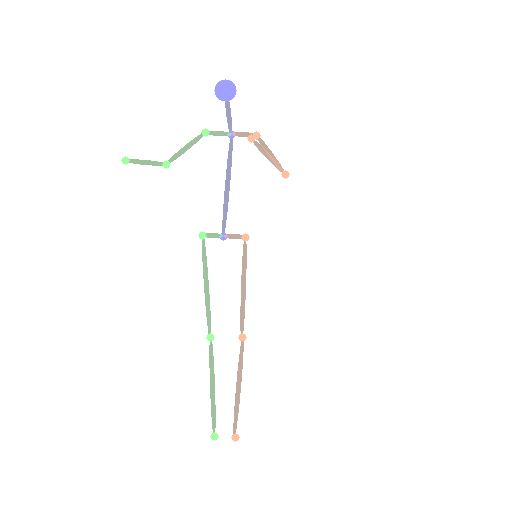} &
    \includegraphics[width=0.105\linewidth]{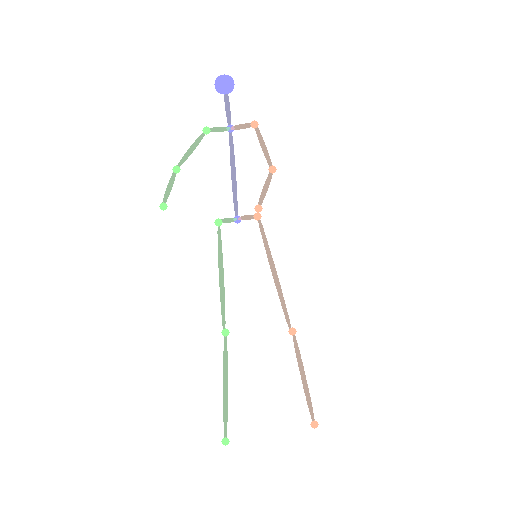} &
    \includegraphics[width=0.105\linewidth]{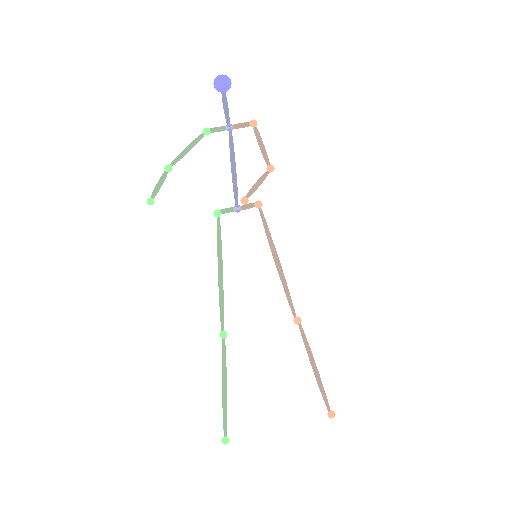} \\ \cline{2-9}
    \includegraphics[width=0.105\linewidth]{figures/application_retargeting/test.png} &
    \includegraphics[width=0.105\linewidth]{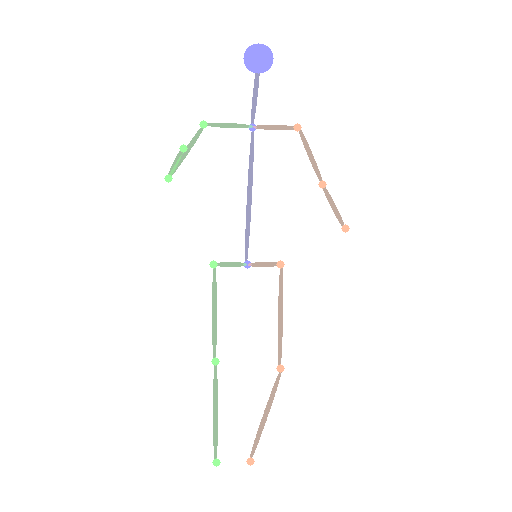} &
    \includegraphics[width=0.105\linewidth]{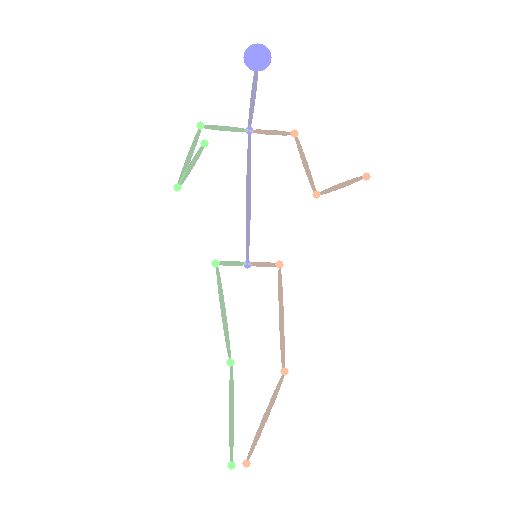} &
    \includegraphics[width=0.105\linewidth]{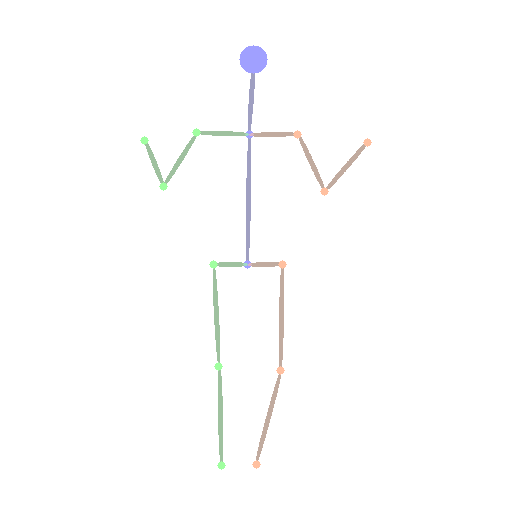} &
    \includegraphics[width=0.105\linewidth]{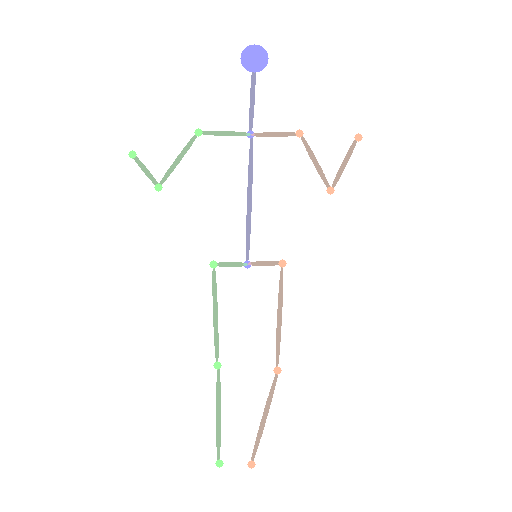} &
    \includegraphics[width=0.105\linewidth]{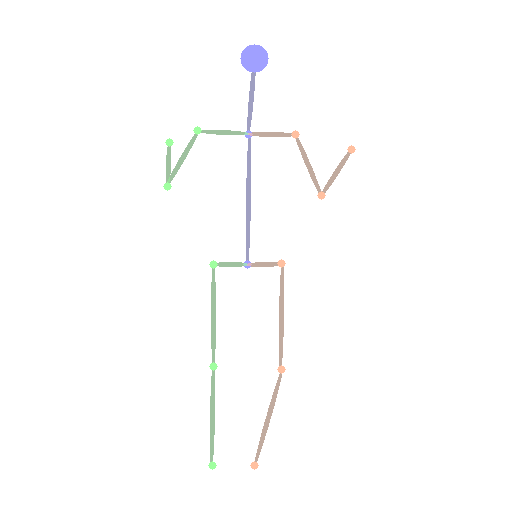} &
    \includegraphics[width=0.105\linewidth]{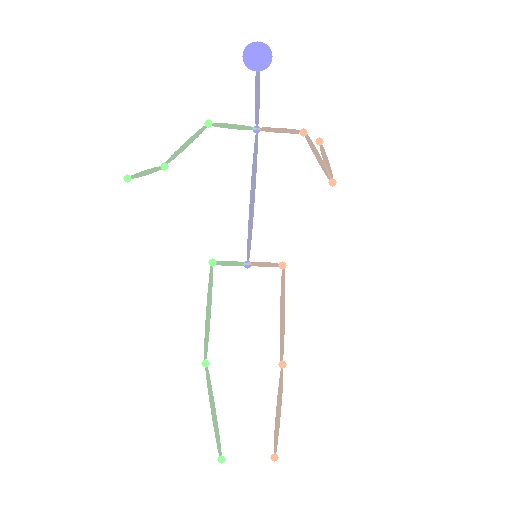} &
    \includegraphics[width=0.105\linewidth]{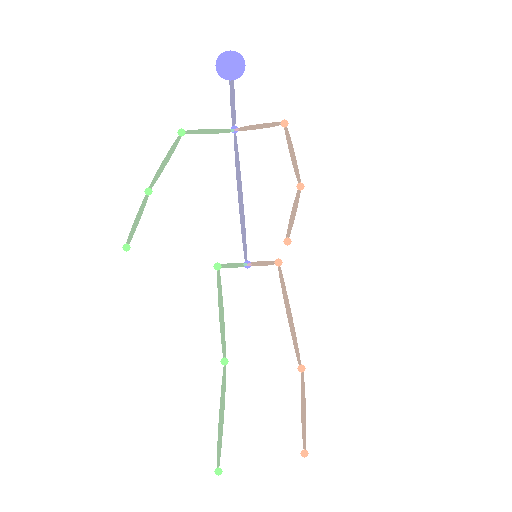} &
    \includegraphics[width=0.105\linewidth]{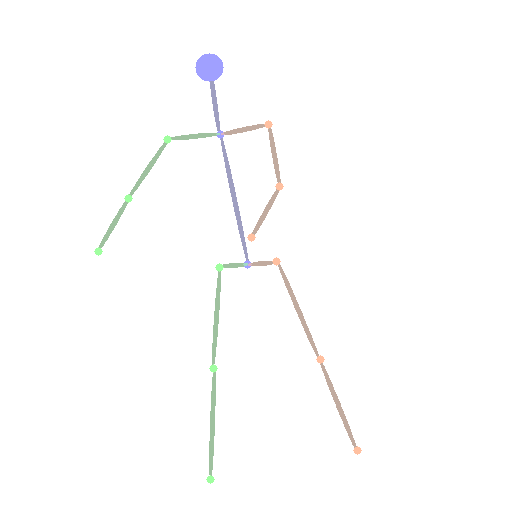} \\
    \includegraphics[width=0.105\linewidth]{figures/application_retargeting/input.jpg} &
    \includegraphics[width=0.105\linewidth]{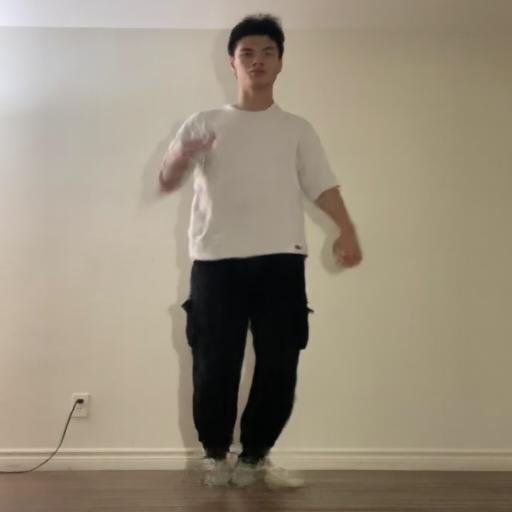} &
    \includegraphics[width=0.105\linewidth]{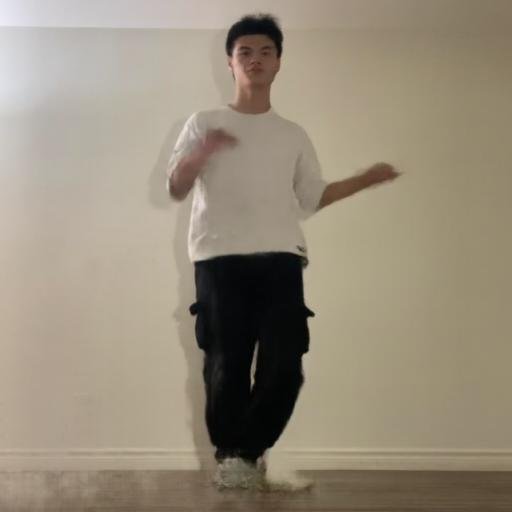} &
    \includegraphics[width=0.105\linewidth]{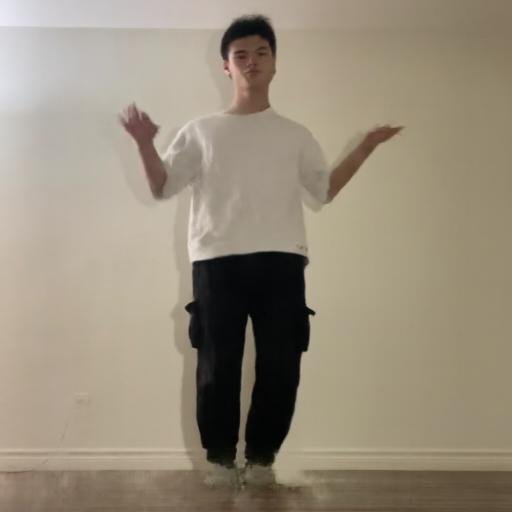} &
    \includegraphics[width=0.105\linewidth]{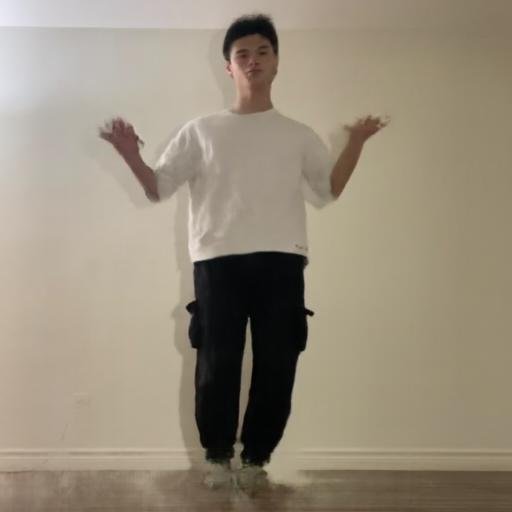} &
    \includegraphics[width=0.105\linewidth]{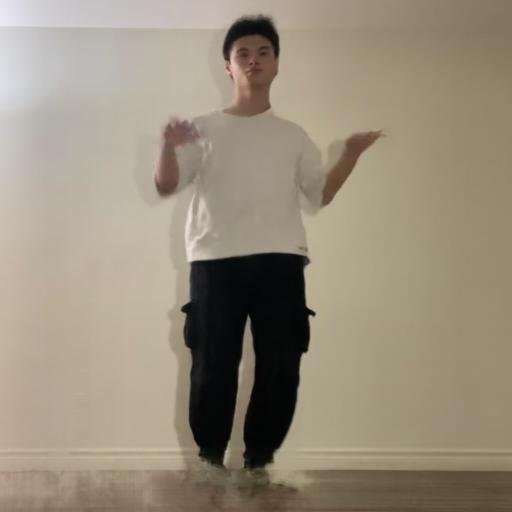} &
    \includegraphics[width=0.105\linewidth]{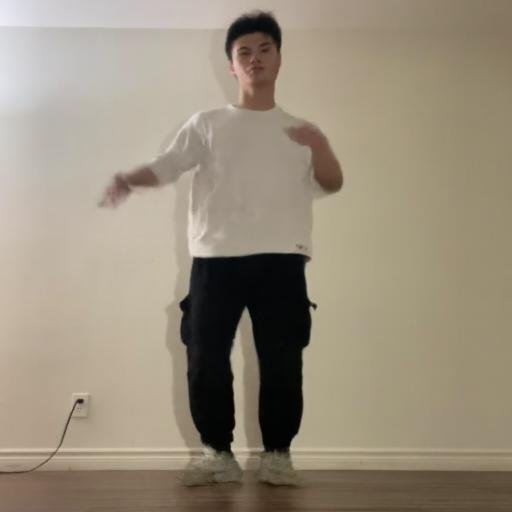} &
    \includegraphics[width=0.105\linewidth]{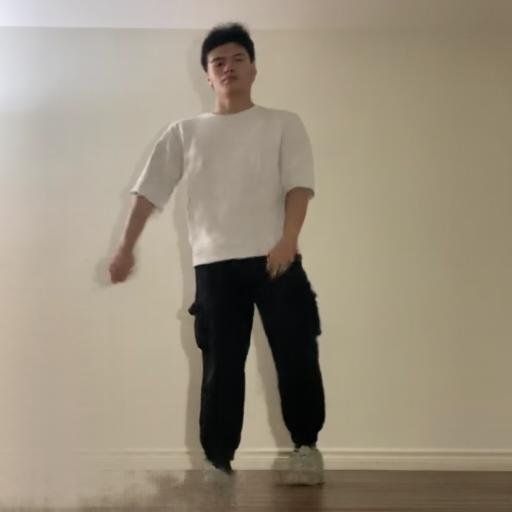} &
    \includegraphics[width=0.105\linewidth]{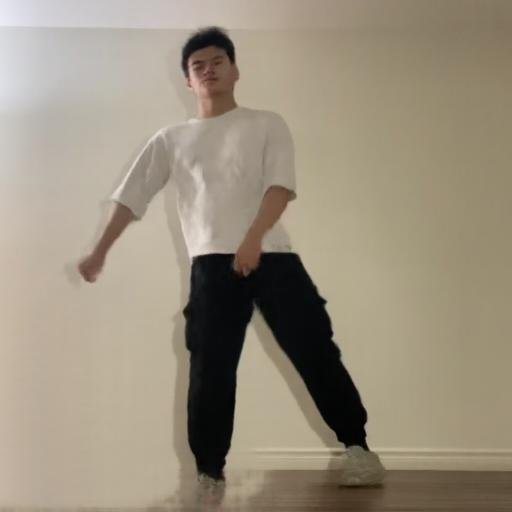} \\
    \multicolumn{9}{c}{(b)} \\
    \end{tabular}
    
    \end{figure*}

\end{document}